\documentclass[sigconf]{acmart}

\usepackage{enumitem}
\usepackage{xspace}
\usepackage{subfiles}
\usepackage{subcaption}
\usepackage{latexsym}
\usepackage{fdsymbol}
\usepackage{amsmath}
\usepackage{balance}
\usepackage{multirow}
\usepackage{pifont}
\usepackage{xcolor}
\usepackage{booktabs}
\usepackage{array}
\usepackage{graphicx}
\usepackage{tabularx}
\usepackage{tikz}
\usetikzlibrary{patterns}
\usepackage{amsmath}
\usepackage{nicefrac}
\usepackage{algorithm} 
\usepackage{algpseudocode} 
\usepackage{cancel}
\usepackage{type1cm}
\usepackage[compact]{titlesec}
\usepackage{placeins}
\usepackage{afterpage}

\AtBeginDocument{%
  \providecommand\BibTeX{{%
    \normalfont B\kern-0.5em{\scshape i\kern-0.25em b}\kern-0.8em\TeX}}}

\copyrightyear{2025}
\acmYear{2025}
\setcopyright{cc}
\setcctype{BY}
\acmConference[KDD '25]{Proceedings of the 31st ACM SIGKDD Conference on Knowledge Discovery and Data Mining V.1}{August 3--7, 2025}{Toronto, ON, Canada}
\acmBooktitle{Proceedings of the 31st ACM SIGKDD Conference on Knowledge Discovery and Data Mining V.1 (KDD '25), August 3--7, 2025, Toronto, ON, Canada}
\acmDOI{10.1145/3690624.3709163}
\acmISBN{979-8-4007-1245-6/25/08}

\theoremstyle{definition}
\newtheorem{defn}{Definition}
\newtheorem{problem}{Problem}
\newtheorem{property}{Property}

\newcommand{\rom}[1]{\uppercase\expandafter{\romannumeral#1}}

\newcommand{\random}{\textsc{Rand. Attack}\xspace}
\newcommand{\dice}{\textsc{DICE}\xspace}
\newcommand{\pgd}{\textsc{PGD Attack}\xspace}
    \newcommand{\structack}{\textsc{Structack}\xspace}
\newcommand{\metattack}{\textsc{Metattack}\xspace}

\newcommand{\svd}{\textsc{SVD}\xspace}
\newcommand{\cosine}{\textsc{Cosine Sim.}\xspace}

\newcommand{\gcn}{\textsc{GCN}\xspace}
\newcommand{\rgcn}{\textsc{RGCN}\xspace}
\newcommand{\mediangcn}{\textsc{MedianGCN}\xspace}
\newcommand{\gnnguard}{\textsc{GNNGuard}\xspace}
\newcommand{\metagc}{\textsc{MetaGC}\xspace}

\newcommand{\gcnsvd}{\textsc{GCNSVD}\xspace}
\newcommand{\degreelg}{\textsc{Degree Model}\xspace}
\newcommand{\lclg}{\textsc{Clustering Coeff. Model}\xspace}
\newcommand{\homophilylg}{\textsc{Homophily Model}\xspace}
\newcommand{\model}{\textsc{LEO}\xspace}
\newcommand{\lfo}{\textsc{LFO}\xspace}
\newcommand{\measure}{\textsc{HideNSeek}\xspace}
\newcommand{\cooccur}{\textsc{Co-occur}\xspace}

\newcommand{\cora}{\textsc{Cora}\xspace}
\newcommand{\citeseer}{\textsc{Citeseer}\xspace}
\newcommand{\coraml}{\textsc{Cora-ML}\xspace}
\newcommand{\lastfmasia}{\textsc{LastFMAsia}\xspace}
\newcommand{\squirrel}{\textsc{Squirrel}\xspace}
\newcommand{\chameleon}{\textsc{Chameleon}\xspace}

\definecolor{RedFig3}{RGB}{200, 0, 0}
\definecolor{BlueFig3}{RGB}{68, 114, 196}

\definecolor{Red}{RGB}{244, 124, 124}
\definecolor{Green}{RGB}{162, 222, 147}
\definecolor{Blue}{RGB}{112, 161, 211}
\definecolor{BlueMark}{RGB}{43,131,186}
\definecolor{GrayMark}{RGB}{99,99,99}
\definecolor{DarkBlue}{RGB}{4,90,141}
\definecolor{SkyBlue}{RGB}{151,186,217}
\definecolor{DarkBlueRec}{RGB}{169,199,224}
\definecolor{SkyBlueRec}{RGB}{222,228,240}
\definecolor{auburn}{rgb}{0.43, 0.21, 0.1}

\newcommand{\darkbluebox}{\raisebox{-1pt}{\tikz{\draw[preaction={fill=DarkBlueRec}, pattern=north east lines, pattern color=gray, draw=DarkBlue](0,0) rectangle (3.0mm,3.0mm);}}}

\newcommand{\skybluebox}{\raisebox{-1pt}{\tikz{\draw[draw=SkyBlue,fill=SkyBlueRec](0,0) rectangle (3.0mm,3.0mm);}}}

\definecolor{newRED}{RGB}{215,25,28}
\definecolor{newBLUE}{RGB}{5,113,176}

\newcolumntype{M}[1]{>{\centering\arraybackslash}m{#1}}

\newcommand\resubmit[1]{\textcolor{black}{#1}}

\newcommand\redstar[1]{\textcolor{redstar}{#1}}

\definecolor{redstar}{RGB}{196, 78, 82}
\definecolor{lucky}{RGB}{255, 97, 97}

\definecolor{hj}{rgb}{0.58, 0.44, 0.86}

\newcommand\kijung[1]{\textcolor{black}{#1}}

\usepackage{contour}
\usepackage[normalem]{ulem}
\contourlength{0.8pt}

\newcommand{\myuline}[1]{%
  \uline{\phantom{#1}}%
  \llap{\contour{white}{#1}}%
}
\newcommand{\smallsection}[1]{{\noindent {\textbf{\myuline{#1}}}}}

\makeatletter
\tikzset{
        hatch distance/.store in=\hatchdistance,
        hatch distance=5pt,
        hatch thickness/.store in=\hatchthickness,
        hatch thickness=5pt
        }
\pgfdeclarepatternformonly[\hatchdistance,\hatchthickness]{north east hatch}
    {\pgfqpoint{-1pt}{-1pt}}
    {\pgfqpoint{\hatchdistance}{\hatchdistance}}
    {\pgfpoint{\hatchdistance-1pt}{\hatchdistance-1pt}}%
    {
        \pgfsetcolor{\tikz@pattern@color}
        \pgfsetlinewidth{\hatchthickness}
        \pgfpathmoveto{\pgfqpoint{0pt}{0pt}}
        \pgfpathlineto{\pgfqpoint{\hatchdistance}{\hatchdistance}}
        \pgfusepath{stroke}
    }
\makeatother

\begin{document}

\emergencystretch 3em

\title[On Measuring Unnoticeability of Graph Adversarial Attacks]{On Measuring Unnoticeability of Graph Adversarial Attacks:\\Observations, New Measure, and Applications}

	\author{Hyeonsoo Jo}
        \authornote{Equal contribution.}
	\affiliation{%
    	\institution{KAIST}
            \city{Seoul}
            \country{Republic of Korea}
	}
	\email{hsjo@kaist.ac.kr}

        \author{Hyunjin Hwang}
        \authornotemark[1]
	\affiliation{%
		\institution{KAIST}
            \city{Seoul}
            \country{Republic of Korea}
	}
	\email{hyunjinhwang@kaist.ac.kr}

        \author{Fanchen Bu}
	\affiliation{%
		\institution{KAIST}
            \city{Daejeon}
            \country{Republic of Korea}
	}
	\email{boqvezen97@kaist.ac.kr}

        \author{Soo Yong Lee}
	\affiliation{%
		\institution{KAIST}
            \city{Seoul}
            \country{Republic of Korea}
	}
	\email{syleetolow@kaist.ac.kr}

        \author{Chanyoung Park}
	\affiliation{%
		\institution{KAIST}
            \city{Daejeon}
            \country{Republic of Korea}
	}
	\email{cy.park@kaist.ac.kr}
	
	\author{Kijung Shin}
    
	\affiliation{%
		\institution{KAIST}
            \city{Seoul}
            \country{Republic of Korea}
	}
	\email{kijungs@kaist.ac.kr}

\renewcommand{\shortauthors}{Hyeonsoo Jo et al.}


\begin{abstract}
  Adversarial attacks are allegedly \textit{unnoticeable}.
Prior studies have designed attack noticeability measures on graphs, 
primarily using statistical tests to compare the topology of original and (possibly) attacked graphs.
However, we observe two critical limitations in the existing measures.
First, because the measures rely on simple rules,
attackers can readily enhance their attacks to \textit{bypass} them, 
reducing their attack ``noticeability'' and, yet, maintaining their attack performance.
Second, because the measures naively leverage global statistics, such as degree distributions,
they may entirely \textit{overlook} attacks until severe perturbations occur, letting the attacks be almost ``totally unnoticeable.''

To address the limitations, we introduce \measure, a learnable measure for graph attack noticeability.
First, to mitigate the bypass problem, \measure \textit{learns} to distinguish the original and (potential) attack edges using a \textbf{\underline{l}}earnable \textbf{\underline{e}}dge sc\textbf{\underline{o}}rer (\model), which scores each edge on its likelihood of being an attack. 
Second, to mitigate the overlooking problem, \measure conducts \textit{imbalance-aware aggregation} of all the edge scores to obtain the final noticeability score.
Using six real-world graphs, we empirically demonstrate that \measure effectively alleviates the observed limitations, and \model (i.e., our learnable edge scorer) outperforms eleven competitors in distinguishing attack edges under five different attack methods.
For an additional application, we show that \model can boost the performance of robust GNNs by removing attack-like edges.

\end{abstract}



\maketitle
\newcommand\kddavailabilityurl{https://doi.org/10.5281/zenodo.14548034}

\ifdefempty{\kddavailabilityurl}{} {
\begingroup\small\noindent\raggedright\textbf{KDD Availability Link:}\\
The source code of this paper has been made publicly available at \url{\kddavailabilityurl}.
\endgroup
}

\section{Introduction}
\label{sec:intro}
Recently, graph neural networks (GNNs) have shown remarkable performance in various tasks, including node classification~\citep{kipf2016semi,hamilton2017inductive, velivckovic2018graph, gasteiger2019predict}, link prediction~\citep{barbieri2014follow, zhang2018link, ou2016asymmetric}, and graph classification~\citep{zhang2018end,lee2018graph}.
However, GNNs are susceptible to adversarial attacks, which is a common issue in deep learning approaches~\citep{jin2021adversarial, zugner2019adversarial, wu2019adversarial, xu2019topology}.

An adversarial attack generates \textit{adversarial examples} intentionally designed to induce incorrect predictions from an attacked model.
Adversarial examples are supposed to be imperceptible or \textit{unnoticeable}~\citep{goodfellow2014explaining,madry2018towards,kurakin2018adversarial,carlini2017towards}.
Otherwise, defenders (who are attacked) would easily detect the attacks to avoid potential harm.
In the image domain, adversarial examples should visually look like normal images.
However, in the graph domain, whether two graphs ``look similar'' may be vague and subjective.
Therefore, numerical \textit{noticeability measures} are necessary to evaluate attack noticeability.

\kijung{Studying graph (topological) adversarial attacks and measuring their noticeability has unique challenges, primarily due to the \textit{discrete} nature of graph changes, i.e., the minimal change is inserting/deleting an edge.}
There have been several attempts using a statistical test on graphs to define (un)noticeable attacks,
where an attack is considered ``unnoticeable'' if the difference after the attack is not statistically significant.
\kijung{Examples include} the likelihood-ratio (LR) test on degree distributions~\citep{zugner2018adversarial,zugner2019adversarial}
and the two-sample Kolmogorov-Smirnov (KS) test on degree distributions, clustering coefficient distributions, and node-level homophily scores~\citep{hussain2021structack,chen2021understanding}.

However, we posit that such existing ``noticeability measures''
\textit{unrealistically assume} naive defenders using a simple and rule-based measure.
Hence, we raise suspicion that the existing measures may be inadequate and do not well evaluate noticeability.
Intuitively, two graphs can significantly differ, even if they share the same topological statistics (e.g., degree distributions or clustering coefficient distributions).
Indeed, we observe the following critical limitations of the existing noticeability measures:

\textit{\textbf{L1) Readily bypassable}}: 
Attackers can readily enhance attacks to 
significantly reduce their noticeability (w.r.t. the existing measures) and, yet, largely maintain their attack performance.
Surprisingly, we observe that (1) the noticeability scores of the existing measures can be reduced by $64\%$ to $100\%$ and, yet, (2) the change in the attack performance is marginal.

\textit{\textbf{L2) Overlooking attacks}}: 
If the attack rate (i.e., the proportion of the attack edges over the original edges) is not large,
the attacks can be ``totally unnoticeable,'' with near-zero noticeability scores.
Specifically, we report that less than $5\%$ attack rate results in near-zero noticeability.
Surprisingly, even with a $10$-$15\%$ attack rate, the existing measures often fail to reject their null hypothesis, mistakenly suggesting that no attacks have occurred. 

\kijung{Motivated by the above observations, we propose \measure, a learnable measure for graph attack noticeability.
It leverages a GNN-based \textbf{\myuline{l}}earnable \textbf{\myuline{e}}dge sc\textbf{\myuline{o}}rer (\model) 
to score each edge on its likelihood of being an attack.
If attack edges have lower scores (i.e., they are well distinguished by \model), they are considered more noticeable.
To obtain the final noticeability score, \measure conducts imbalance-aware aggregation of the scores.
In summary, \measure 
(1) alleviates the ``bypassable'' problem by learning to distinguish the original and (potential) attack edges rather than relying on simple rules, and 
(2) alleviates the ``overlooking'' problem by imbalance-aware aggregation.}


We demonstrate the effectiveness of \model and \measure via extensive experiments on six real-world graphs.
First, \model outperforms eleven competitors in detecting attack edges in 23 out of 28 cases, under five different attack methods.
Second, \measure is robust to the bypassable problem, being $0.38\times$ to $5.75\times$ less bypassable than the second-best measure.
Third, \measure mitigates the overlooking problem, returning considerable noticeability scores even with a low attack rate.
Lastly, as a further application, we demonstrate that removing attack-like edges using \model enhances the node classification performance of GNNs. Even for robust GNNs \cite{zhu2019robust,chen2021understanding2}, it  
yields up to 12.4\% better accuracy than the best competitor.


Our contributions are summarized as follows:
\begin{itemize}[leftmargin=*,topsep=0pt]
    \item \textbf{Observations} (Sec.~\ref{sec:limit}): 
    We observe two critical limitations of the existing noticeability measures: they are readily bypassable; 
    they overlook attacks until severe perturbations occur.
    \item \textbf{New measure} (Sec.~\ref{sec:method}): 
    To address the limitations, we propose \measure, a noticeability measure  that leverages 
    (1) a \textbf{\myuline{l}}earnable \textbf{\myuline{e}}dge sc\textbf{\myuline{o}}rer (\model) and (2) imbalance-aware aggregation.    
    \item \textbf{Empirical evaluation} (Sec.~\ref{sec:exp}): 
    In six real-world graphs, we empirically demonstrate the effectiveness of \measure and \model.
    Compared to eleven baselines, \model achieves the overall best performance in detecting attack edges. 
    Most importantly, \measure achieves significantly smaller bypass rates and significantly higher sensitivity to small attack rates, compared to all the existing measures, alleviating the observed limitations.
    \item \textbf{Further applications} (Sec.~\ref{sec:exp:q3}):
    We further apply \model to filter out attack-like edges in graphs to improve GNN robustness,
    and we find \model to be more effective than the baselines.
\end{itemize}

\titlespacing{\section}{0pt}{\parskip}{-\parskip}
\titlespacing{\section}{0pt}{\parskip}{\parskip}
\titlespacing{\subsection}{0pt}{1mm}{1mm}
\titlespacing{\subsubsection}{0pt}{1mm}{1mm}

\section{Preliminaries and Related Work}
\label{sec:prelim}
\smallsection{Graphs.}
A graph $G = (A, X)$ with $n$ nodes (WLOG, assume the nodes are $V = [n] = \{1, 2, \ldots, n\}$) is defined by 
(1) its graph topology, i.e., an adjacency matrix $A \in \{0, 1\}^{n \times n}$, where each edge $(i, j)$ exists if and only if $A_{ij} = 1$, and
(2) its node features $X \in {\mathbb{R}}^{n \times k}$, where each node $i$ has a $k$-dimensional node feature $X_i \in \mathbb{R}^k$.

\smallsection{Graph statistics.}
The set of \textit{neighbors} $N_i = \{j \in [n]: A_{ij} = 1\}$ of a node $i$ consists of the nodes adjacent to $i$.
The \textit{degree} of $i$, $d_i = \vert N_i \vert$, is the number of neighbors of $i$.
The \textit{clustering coefficient} of $i$, $c_i = (\sum_{(j, k) \in {\binom{N_i}{2}}} A_{jk})/{\binom{d_i}{2}}$, measures the connectivity among the neighbors of $i$.
The \textit{node homophily}~\citep{chen2021understanding} of $i$, 
$h_i = \operatorname{CosSim} (X_i, \sum_{j\in N_i}{\frac{1}{\sqrt{d_j}\sqrt{d_i}}X_j})$, measures the similarity between $i$ and its neighbors, where $\operatorname{CosSim}$ is the cosine similarity.

\smallsection{Graph adversarial attacks.}
For attackers, a graph adversarial attack~\citep{sun2022adversarial} is a constrained optimization problem:\footnote{
Researchers consider different variants of graph adversarial attacks, and we refer the readers to a recent survey~\citep{sun2022adversarial} for more details.}

\vspace{-1mm}
\begin{problem}[Graph adversarial attacks]\label{prob:graph_adv_atk}
    \begin{itemize}[leftmargin=*,topsep=0pt]
        \item \textbf{Given:} A graph $G = (A, X)$, 
        a target model $f_{\text{tgt}}$,
        an attack budget $\Delta$,
        a noticeability constraint $\delta$, and an noticeability measure $U$;
        \item \textbf{Find:} A perturbed graph (i.e., an attack) $\hat{G} = (\hat{A}, \hat{X})$;
        \item \textbf{to Minimize:} The performance of the target model $f_{\text{tgt}}$ on $\hat{G}$,\footnote{Depending on how much knowledge about the target model the attacker has, the performance can be different forms (including estimation) obtained by the attacker.} 
        denoted by $\mathcal{P}(\hat{G}; f_{\text{tgt}})$;
        \item \textbf{with Constraints:} Limited perturbations $\Vert \hat{A} - A \Vert + \Vert \hat{X} - X \Vert \leq \Delta$ \kijung{(or $\gamma \Vert A \Vert_1$; see below)}
        and limited noticeability $U(\hat{G}|G) \leq \delta$.
    \end{itemize}    
\end{problem}

\noindent For topological attacks (i.e., $\hat{X} = X$), the perturbation budget can be rewritten as $\Delta = \gamma \Vert A \Vert_1$ for a \textit{attack rate} $\gamma \in (0, 1)$.

An attack has high \textit{attack performance} if the performance of the target model $\mathcal{P}(\hat{G}; f_{\text{tgt}})$ is low.
\kijung{Adding malicious edges or deleting crucial edges has been introduced for graph adversarial attacks on the node classification task.\footnote{We focus on node classification considered by most graph adversarial attack methods~\cite{waniek2018hiding,hussain2021structack,dai2018adversarial,sun2022adversarial}.}
For example, \dice~\citep{waniek2018hiding} connects nodes with different labels and removes the edges between same-labeled nodes. 
\structack~\cite{hussain2021structack} uses node centrality (e.g., Pagerank) or similarity (e.g., Katz) metrics, selecting nodes with the lowest centrality and connecting the node pairs with the lowest similarity. 
Gradient-based attacks~\citep{zugner2018adversarial,xu2019topology,wu2019adversarial, dai2018adversarial} have also been developed, including \pgd~\citep{xu2019topology}  and \metattack~\citep{zugner2019adversarial}, which are based on projected gradient descent and meta-learning, respectively.
In addition, graph attacks through node injection \cite{chen2021understanding,tao2021single,sun2020adversarial} and node feature perturbation \cite{zugner2018adversarial, zugner2019adversarial, ma2020towards, xu2024attacks, takahashi2019indirect} have been explored.}

\kijung{\textbf{This work focuses on topological attacks} (i.e., $\hat{X} = X$) 
due to the relative scarcity of feature attack methods and especially the limited study of their noticeability, which is our focus.
However, note that our proposed measure, \measure, is extended for the noticeability of node feature attacks, as discussed in the Appendix~\ref{appendix:feature}.}

\kijung{Besides the development of attack methods, generating unnoticeable attacks is another major challenge.
In this context, we discuss perturbation budgets and noticeability measures below.
}

\begin{figure*}[t]
    \begin{subfigure}{0.22\linewidth}
        \centering
        \includegraphics[width=\textwidth]{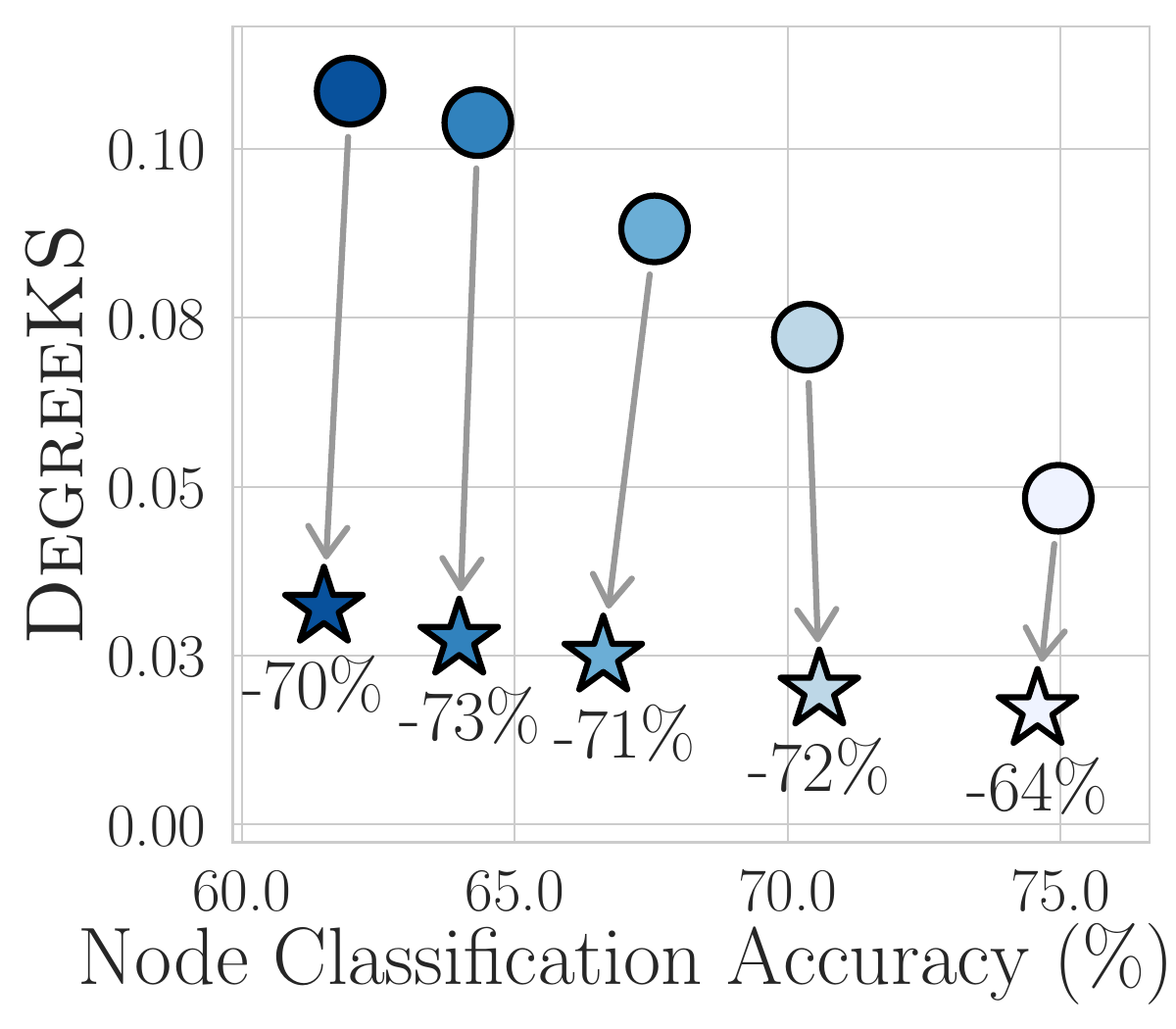}
        \caption{\textsc{DegreeKS}}
        \label{fig:crown:degree_bypass}
    \end{subfigure}
    \hfill
    \begin{subfigure}{0.22\linewidth}
        \centering
        \includegraphics[width=\linewidth]{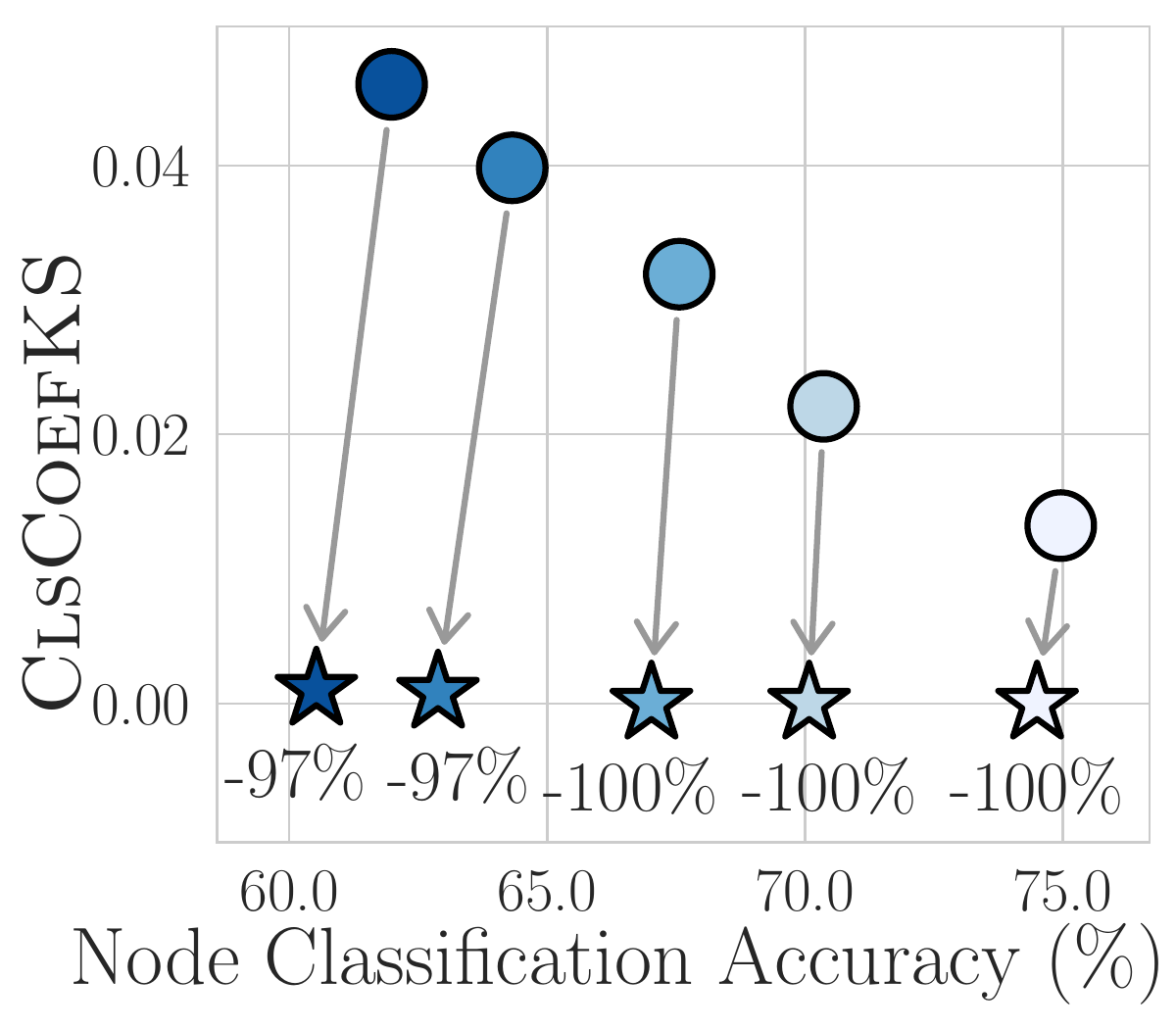}
        \caption{\textsc{ClsCoefeKS}}
        \label{fig:crown:lc_bypass}
    \end{subfigure} 
    \hfill
    \begin{subfigure}{0.22\linewidth}
        \centering
        \includegraphics[width=\linewidth]{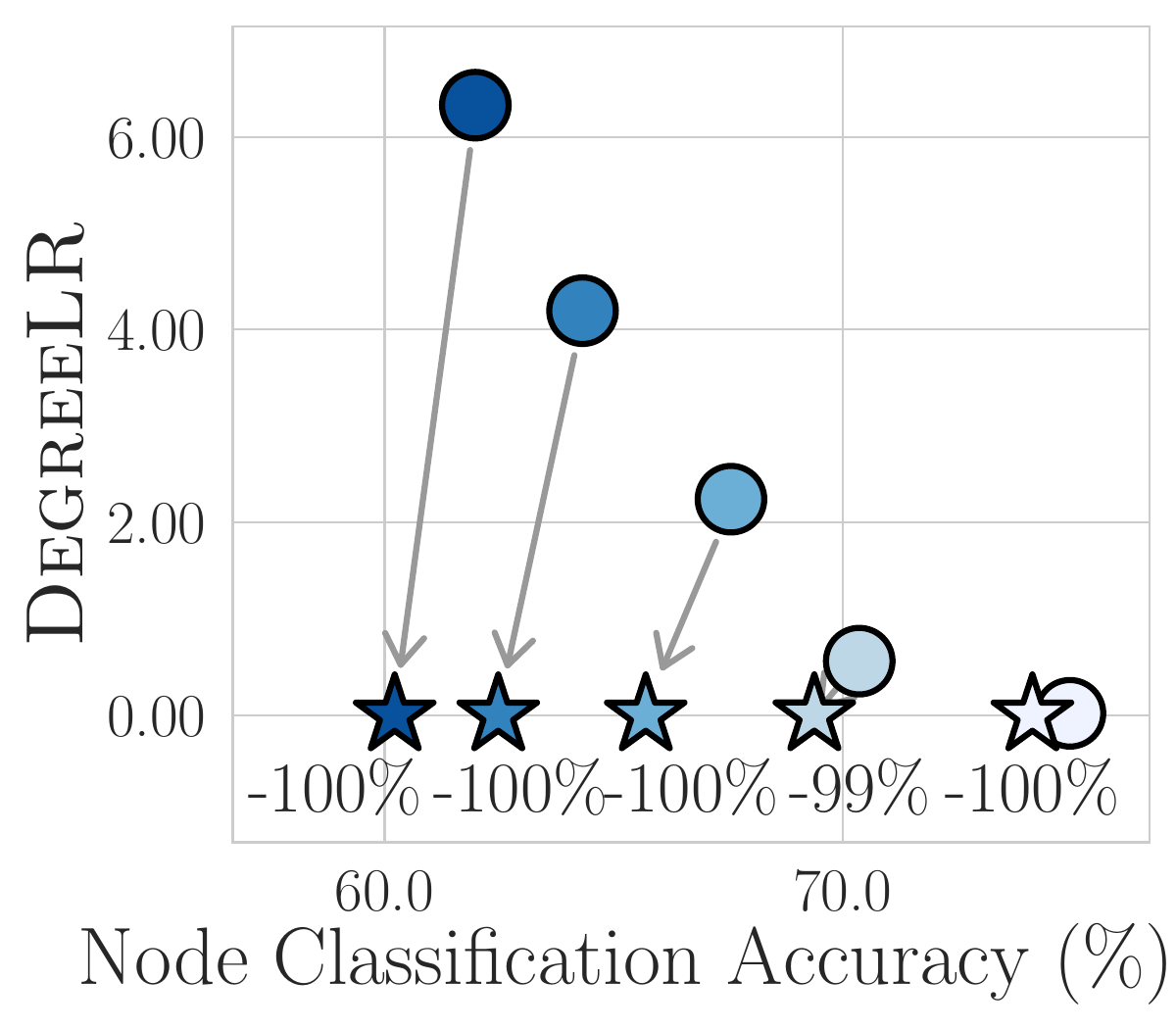}
        \caption{\textsc{DegreeLR}}
        \label{fig:crown:llr_bypass}
    \end{subfigure} 
    \hfill
    \begin{subfigure}{0.22\linewidth}
        \centering
        \includegraphics[width=\linewidth]{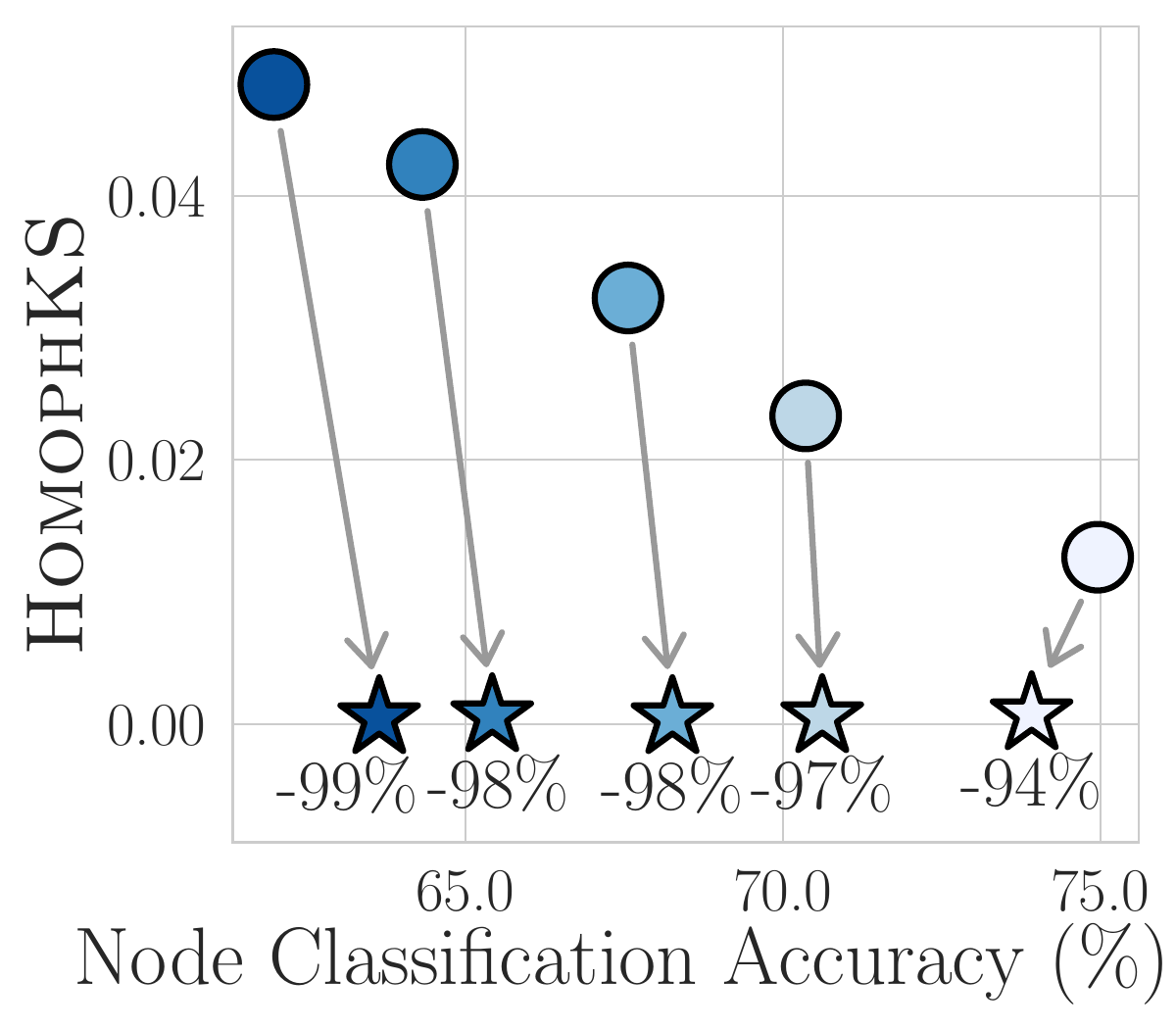}
        \caption{\textsc{HomophKS}}
        \label{fig:crown:homophily_bypass}
    \end{subfigure} 
    \includegraphics[width=0.074\linewidth]{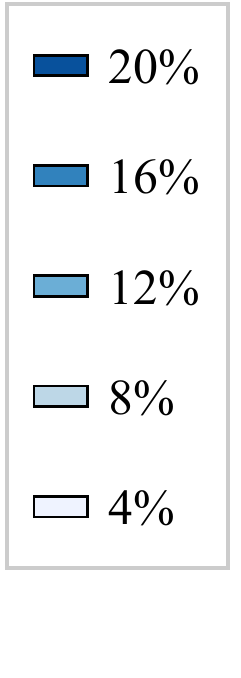}
    \caption{
    The existing noticeability measures are \textbf{\textit{bypassable}}. 
    The y-axis stands for noticeability (the attack is more noticeable when it is higher)
    and the x-axis stands for node classification accuracy (the attack is more effective when it is lower).
    Circle markers ($\medcirc$) represent original attacks generated through \pgd on \cora, 
    while star markers ($\medwhitestar$) indicate adaptive attacks aiming to minimize the corresponding noticeability. 
    The color filled in each marker indicates the attack rate.
    \kijung{Note that adaptive attacks significantly reduce the noticeability measures while largely maintaining the attack performance.}
    }
    \label{fig:crown:bypass}
\end{figure*}

\smallsection{Noticeability measures.}
The noticeability measure $U(\hat{G} | G)$ in Problem~\ref{prob:graph_adv_atk} numerically evaluates the difference between $\hat{G}$ and $G$, and gives a higher value when they are more dissimilar.
Specifically, 
statistical tests have been used to compare the distributions of graph statistics $S$ of $G$ and $\hat{G}$.
The null hypothesis is that the two distributions are the same.
\citet{hussain2021structack} and \citet{chen2021understanding} use the (two-sample) Kolmogorov-Smirnov (KS) test~\citep{massey1951kolmogorov,smirnov1939estimate} on the graph statistics.
The (two-sample) Kolmogorov-Smirnov (KS) test computes the KS D-statistic to measure the difference between $G$ and $\hat{G}$ w.r.t. $S$.
The null hypothesis is rejected when the D-statistic is high (i.e., $G$ and $\hat{G}$ are significantly different w.r.t. the tested graph statistics).
\citet{hussain2021structack} adopts degree distributions and local clustering coefficient distributions, and \citet{chen2021understanding} examine on the node homophily distributions.
\citet{zugner2018adversarial, zugner2019adversarial} use the likelihood-ratio (LR) test~\citep{king1998unifying} on degree distributions.
The LR test compares two distributions by 
(1) fitting the first distribution to a prior model, and
(2) computing the likelihood of the second distribution under the fit model.
The null hypothesis is rejected when the likelihood is low.
Instead of statistical tests,
\citet{gosch2023revisiting} measured noticeability by examining the changes in the predicted labels of each node caused by graph attacks, using synthetic graphs (e.g., CSBM) where an optimal classifier is defined. 
However, their analysis was confined to synthetic graphs where an optimal classifier is known, and extending the analysis for practical usage on real-world graphs is non-trivial.

\section{Observations: Limitations of Existing ``Noticeability Measures''}
\label{sec:limit}
\smallsection{Intuition.}
What makes a good noticeability measure?
\textit{Attackers} use noticeability measures for self-evaluation.
That is, attackers assume that defenders (i.e., who are attacked) use a specific measure and check whether their attacks are unnoticeable under that measure.
Hence, attackers should consider measures that defenders would realistically use.
For \textit{defenders}, on the contrary, a good measure should 
(1) be a strong constraint to enforce the attackers to 
\kijung{sacrifice 
attack performance to reduce noticeability} and 
(2) have high sensitivity, detecting even a small number of attack edges.

\begin{figure*}[t!]
    \begin{subfigure}{0.245\linewidth}
        \centering
        \includegraphics[width=\textwidth]{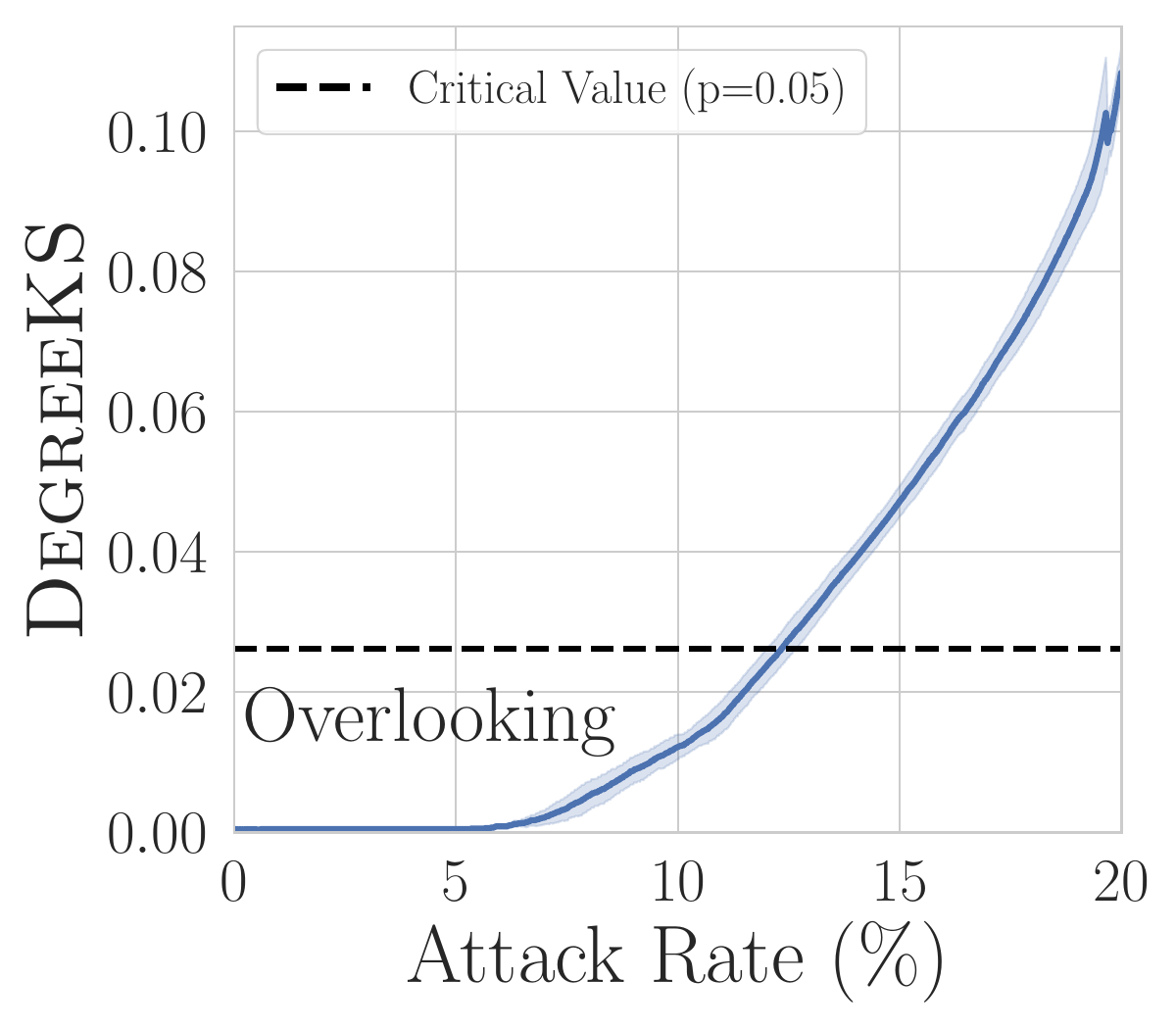}
        \caption{\textsc{DegreeKS}}
        \label{fig:crown:degree_overlooking}
    \end{subfigure}
    \hfill
    \begin{subfigure}{0.245\linewidth}
        \centering
        \includegraphics[width=\linewidth]{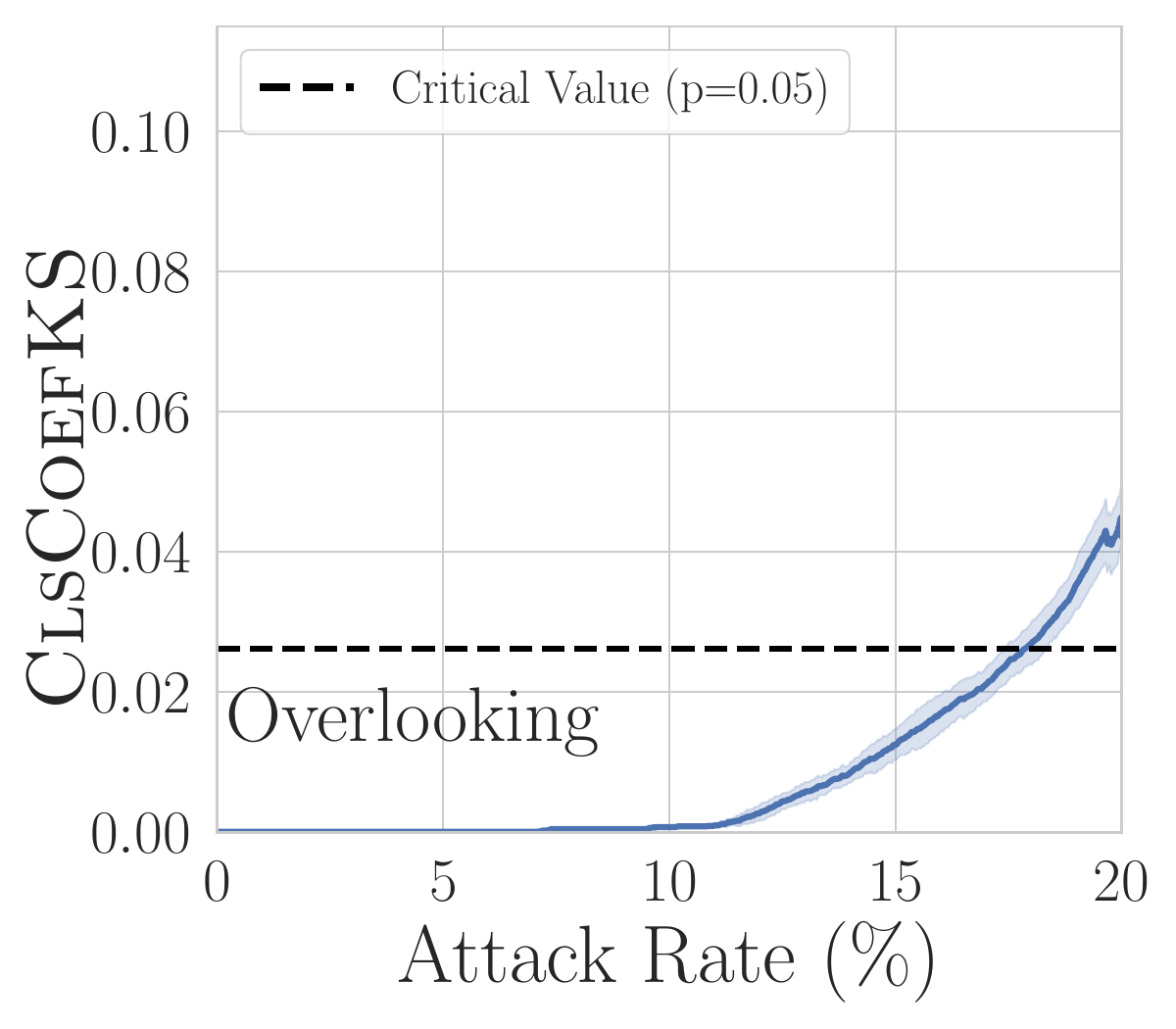}
        \caption{\textsc{ClsCoefKS}}
        \label{fig:crown:lc_overlooking}
    \end{subfigure} 
    \hfill
    \begin{subfigure}{0.245\linewidth}
        \centering
        \includegraphics[width=\linewidth]{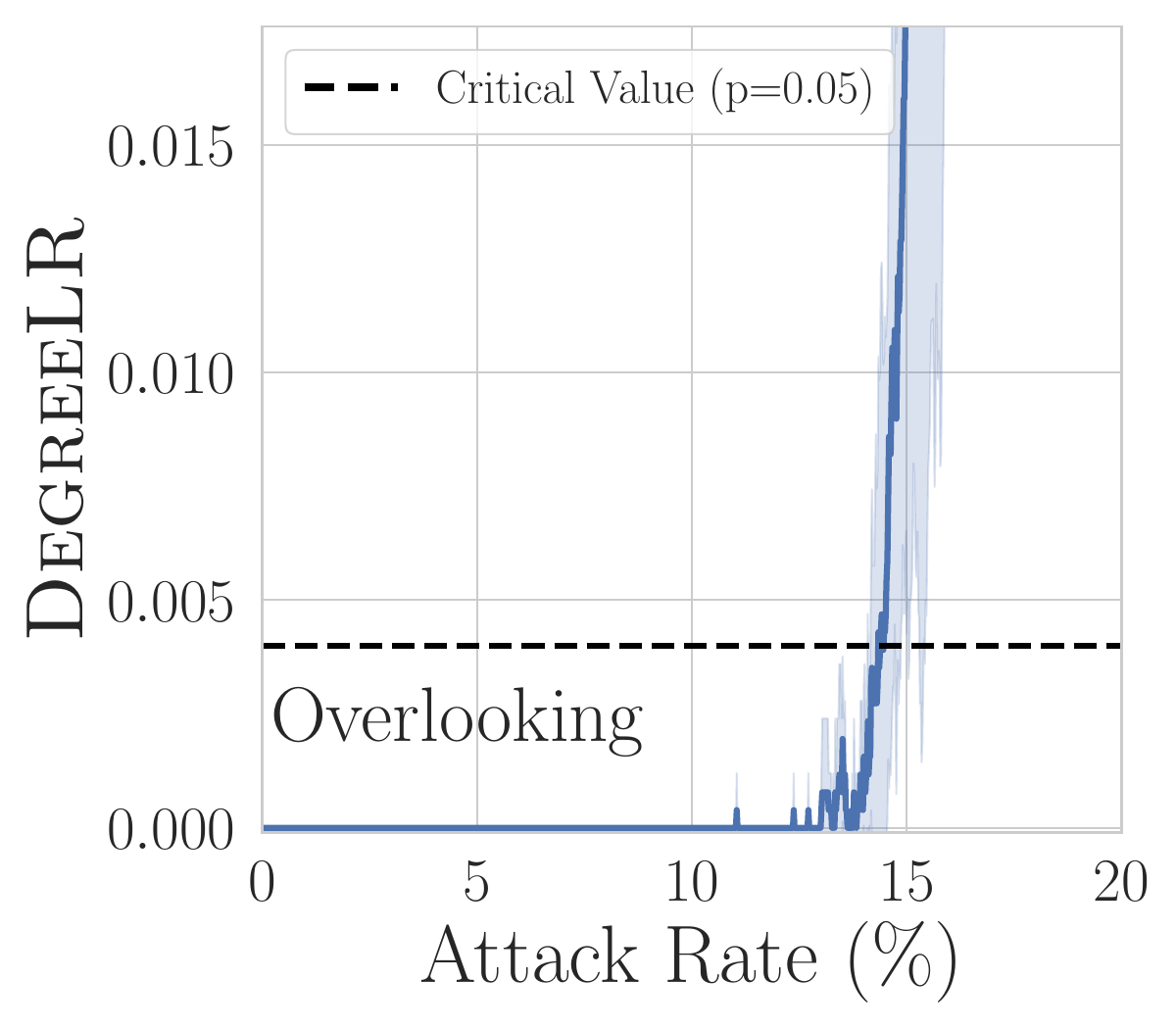}
        \caption{\textsc{DegreeLR}}
        \label{fig:crown:llr_overlooking}
    \end{subfigure} 
    \hfill
    \begin{subfigure}{0.245\linewidth}
        \centering
        \includegraphics[width=\linewidth]{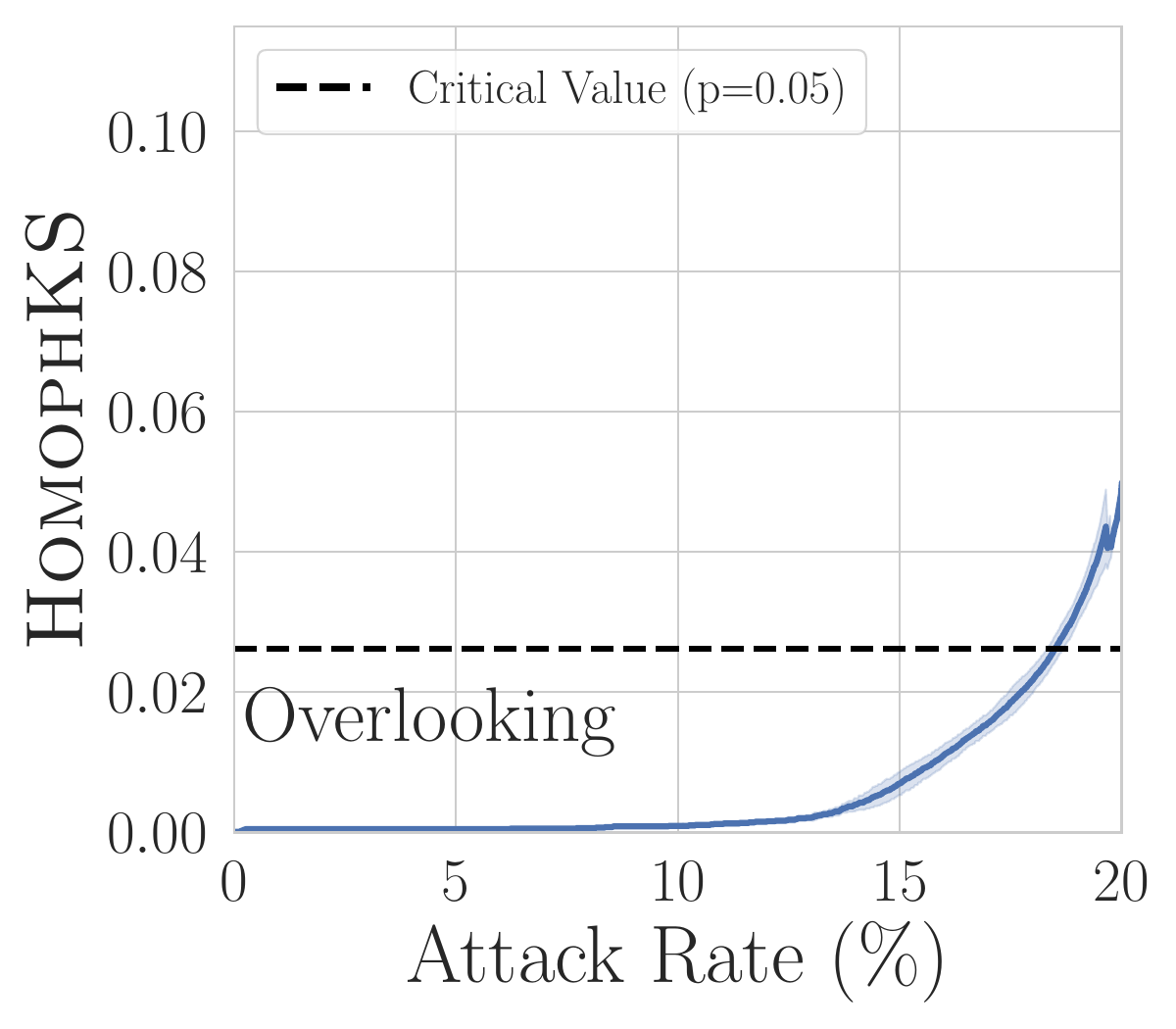}
        \caption{\textsc{HomophKS}}
        \label{fig:crown:homophily_overlooking}
    \end{subfigure} 
    \caption{
    The existing noticeability measures are \textbf{\textit{overlooking}}.    
    The y-axis stands for noticeability scores
    (the attack is more noticeable when it is higher)
    and the x-axis stands for the attack rate.
    Each black dashed line indicates a threshold where the attack is considered noticeable with statistical significance (with $p = 0.05$). 
    \kijung{Note that the noticeability scores remain near-zero until the attack rate $\gamma$ reaches a sufficiently high level (around 5-12\%).
    }}
    \label{fig:crown:overlooking}
\end{figure*}

\smallsection{Overview.}
We raise suspicion that the existing noticeability measure $U$'s inadequately imply \textit{unrealistic defenders}
who warrant attack detection with some graph statistics similarity.
Accordingly, we observe two critical limitations in the existing measure $U$'s:
\begin{enumerate}[leftmargin=2.5em,topsep=0pt]
    \item[\textbf{L1)}] \textbf{Bypassable (Fig.~\ref{fig:crown:bypass})}:
    Attackers can readily, yet significantly, reduce the existing noticeability measures while largely maintaining the attack performance.
    \item[\textbf{L2)}] \textbf{Overlooking (Fig.~\ref{fig:crown:overlooking})}: 
    Until severe perturbations to graph topology occur,
    an attack is overlooked by the existing noticeability measures (i.e., near-zero noticeability).  
\end{enumerate}
\noindent
Details of our observations are provided in the rest of this section.

\smallsection{General settings.}
Throughout this section, we use four existing measure $U$'s:
(1; \textsc{DegreeKS}) KS test on degree distribution~\cite{hussain2021structack},
(2; \textsc{ClsCoefKS}) KS test on clustering coefficient distribution~\cite{hussain2021structack},
(3; \textsc{DegreeLR}) LR test on degree distributions~\cite{zugner2018adversarial,zugner2019adversarial}, and
(4; \textsc{HomophKS}) KS test on node homophily distributions~\cite{chen2021understanding}.
For attacks,
as suggested by~\citet{mujkanovic2022defenses},\footnote{\citet{mujkanovic2022defenses} argued that using adaptive attacks in the evaluation of a defense method (and, accordingly, noticeability in our case) should be the `gold standard.'} we design an adaptive attack (refer to Def.~\ref{defn:adaptive} for details) and use it throughout our analysis.
\begin{defn}[Adaptive Attack]
\label{defn:adaptive}
Given a candidate budget $\Delta_{C}$ ($\geq \Delta$) and an attack method $K$, in addition to the inputs of Problem~\ref{prob:graph_adv_atk}
(i.e., a graph $G$, an attack budget $\Delta$, a noticeability constraint $\delta$, and an noticeability measure $U$),
the adaptive attack $K_a$ (built upon $K$)
(1) obtains $\Delta_{C}$ candidate attack edges (i.e., by using $\Delta_{C}$ as the budget instead of $\Delta$) using $K$ without considering noticeability (i.e.,  $\delta=\infty$),
and (2) among them, greedily adds ${\Delta}$ attack edges one by one to $G$ to minimize $U$, resulting in the attacked graph $\hat{G}_{U}$.
\end{defn}

\subsection{Limitation 1: Bypassable}
\label{sec:limit:bypassable}
\smallsection{Hypothesis.}
We question whether the existing measures are readily \textit{bypassable}.
Intuitively, since they are non-adaptive and rule-based, 
attackers can bypass them with adaptive attacks that intentionally minimize the corresponding noticeability scores.
We define the bypassable problem as follows:
\begin{defn}[Bypassability]
Given an \kijung{unattacked} graph $G$, an original attack graph $\hat{G}$, an adaptive attack graph $\hat{G}_U$, and a target model $f_{\text{tgt}}$,
a noticeability measure $U$ is bypassable if
(1) the noticeability score is significantly reduced (i.e., $U(\hat{G}_U | G) \ll U(\hat{G} | G)$) and 
(2) the attack performance is largely maintained (i.e., the target model performance $\mathcal{P}(\cdot~; f_{\text{tgt}})$ satisfies that $\mathcal{P}(\hat{G}_U; f_{\text{tgt}}) \approx \mathcal{P}(\hat{G}; f_{\text{tgt}})$).
\end{defn}

\smallsection{Experimental settings.}
We empirically test the bypassable problem of the existing measure $U$'s.
Original attack graphs ($\hat{G}$'s) are generated using $K = \text{\pgd}$~\cite{xu2019topology} on the \cora dataset, with an attack rate $\gamma \in \{4\%, 8\%, 12\%, 16\%, 20\%\}$ (recall that $\Delta = \gamma \Vert A \Vert_1$; see Problem~\ref{prob:graph_adv_atk}).
For each measure $U$ and each attack rate $\gamma$, we generate an adaptive attack graph $\hat{G}_U$ (see Def.~\ref{defn:adaptive}) with $\Delta_C = 4\Delta$.

\smallsection{Observation.}
We observe that the existing measure $U$'s are readily bypassable.
For each measure $U$, Fig.~\ref{fig:crown:bypass} shows the attack performance and noticeability of each original attack (i.e., $\mathcal{P}(\hat{G}; f_{\text{tgt}})$ and $U(\hat{G} | G)$; denoted by $\medcirc$'s) and 
those of the corresponding adaptive attack (i.e., $\mathcal{P}(\hat{G}_U; f_{\text{tgt}})$ and $U(\hat{G}_U | G)$; denoted by $\medwhitestar$'s).
Note that $\hat{G}$ (i.e., the $\medcirc$'s) is the same for all the measures while $\hat{G}_U$ (i.e., each $\medwhitestar$) is distinct for each measure $U$.
{The noticeability scores of $\hat{G}_U$ are 64\% to 100\% smaller than those of $\hat{G}$ in all measure $U$'s.}
For three out of four existing measures, the decrease in the noticeability score is at least 97\%.
\kijung{However, surprisingly,} the attack performance decreases by only at most 2 percentage points. 
Surprisingly, for \textsc{ClsCoefKS} (see Fig.~\ref{fig:crown:bypass}(b)) and \textsc{DegreeLR} (see Fig.~\ref{fig:crown:bypass}(c)), the attack performances even increase.
Based on the limitation, we propose the following desirable property of a noticeability measure $U$:
\begin{property}[Non-bypassability]\label{prop:nonbypassable}
    {A desirable noticeability measure $U$ should enforce a trade-off between noticeability and attack performance for adaptive} attacks.
    An attack either
    (1) fails to significantly reduce its noticeability, i.e., $U(\hat{G}_U | G) \not \ll U(\hat{G} | G)$, or
        (2) fails to maintain its attack performance, i.e., $\mathcal{P}(\hat{G}_U; f_{\text{tgt}}) \gg \mathcal{P}(\hat{G}; f_{\text{tgt}})$.
\end{property}

\subsection{Limitation 2: Overlooking}
\label{sec:limit:overlooking}
\smallsection{Hypothesis.}
We further question whether the existing measures \textit{overlook} attacks until severe perturbations to graph topology occur.
Intuitively, we suspect they suffer from the overlooking problem because they naively leverage \textit{global} graph statistics, which may require substantial perturbation to reach statistical significance.
We define the overlooking problem as follows:
\begin{defn}[Overlookingness]
Given an unattacked graph $G$, an adaptive attack graph $\hat{G}_U$, and an attack rate $\gamma$,
the noticeability measure $U$ is overlooking if
{the attack has near-zero noticeability score (i.e., $U(\hat{G}_U | G) \approx 0$),
when $\gamma$ is considerably high (i.e., $\gamma \gg 0$).}
\end{defn}

\smallsection{Experimental settings.}
We empirically test the overlooking problem of the existing measure $U$'s.
First, we use $K = \pgd$ on the \cora dataset with candidate budget $\Delta_C = \gamma_C\Vert A \Vert_1$ with $\gamma_C = 20\%$, to generate attack edge candidates.
As described in Def.~\ref{defn:adaptive}, for each measure $U$, to generate adaptive attacks, we greedily add attack edges from the candidates, to minimize the noticeability score.
To test the hypothesis, we observe the minimum attack rate $\gamma$ (recall that $\Delta = \gamma \Vert A \Vert_1$; see Problem~\ref{prob:graph_adv_atk}) necessary for the {adaptive attack} to be noticeable.

\smallsection{Observation}.
We observe that the existing measure $U$'s suffer from the overlooking problem.
For each measure $U$, Fig.~\ref{fig:crown:overlooking} visualizes the change in its noticeability scores as the attack rate $\gamma$ increases from $0$ to $\gamma_C = 20\%$.
An adaptive attack has a near-zero noticeability score until its attack rate $\gamma$ is up to 5-12\%.
Furthermore, the minimum $\gamma$ required to show statistical significance (with $p = 0.05$) is higher than 10\% for each existing measure $U$.
Therefore, we propose the second desirable property of a noticeability measure $U$:
\begin{property}[Sensitivity]\label{prop:sensitive}
\vspace{-2mm}
    Even for adaptive attacks with a low attack rate $\gamma$,
    a desirable noticeability measure $U$ should return considerable noticeability, i.e., $U(\hat{G}_U | G) \gg 0$.
\end{property}

\smallsection{Distinction between bypassable and overlooking issues.} %
A measure $U$ can be bypassable yet not overlooking, or vice versa.
Bypassable refers to the measure $U$'s failure in enforcing a trade-off between noticeability and attack performance, regardless of the attack rate $\gamma$. 
Overlooking, on the other hand, depends on $\gamma$. 
Specifically, it means that an attack may have near-zero noticeability if the attack rate $\gamma$ is low.
Below is an example.
Let $p \gg q \gg 0$. A measure $U$ is bypassable but not overlooking if, with a low attack rate $\gamma$, an adaptive attack can reduce the noticeability score w.r.t. $U$ significantly from $p$ to $q$ while maintaining attack performance, where the \textit{relative} drop in noticeability is huge ($q \ll p$), but the final \textit{absolute} noticeability is still high ($q \gg 0$).

\section{Proposed Noticeability Measure}
\label{sec:method}
In this section, we propose a novel noticeability measure, \measure, that significantly alleviates the observed limitations.

\begin{figure}[t!]
     \centering
         \centering         
         \includegraphics[width=\linewidth]{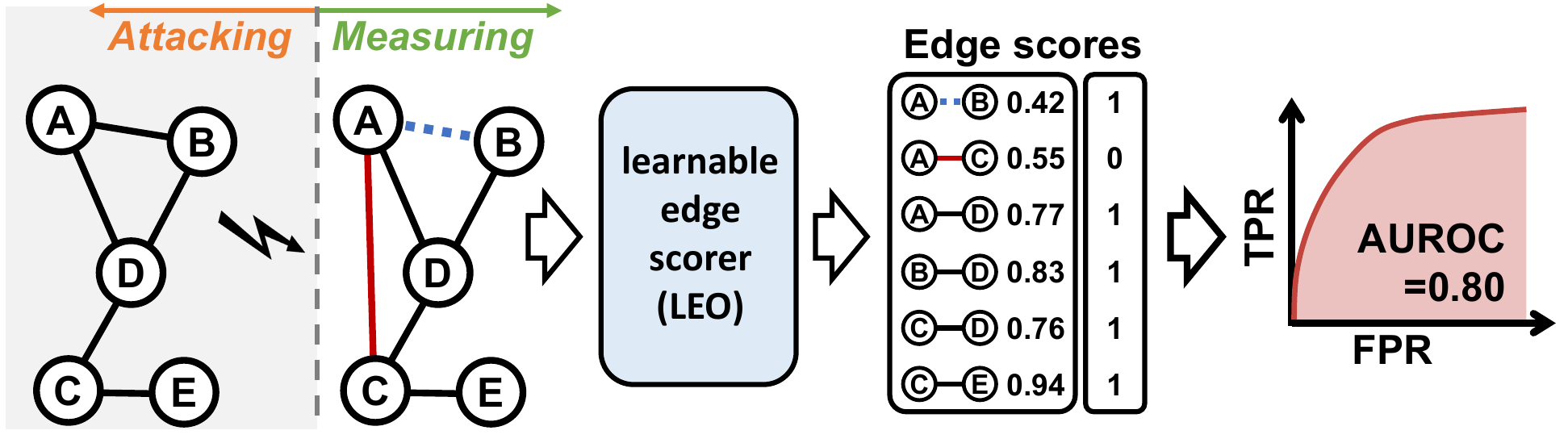} 
         \caption{(Left) \kijung{Attack phase where some attack edges can be added (\textcolor{RedFig3}{red line}) and some original edges can be deleted} (\textcolor{BlueFig3}{blue dashed line}).
         (Right) The procedure of computing \measure.
         Using the edge scores from \model (see Fig.~\ref{fig:model} for its details) as (soft) predictions, the noticeability score w.r.t. \measure is computed as the AUROC score.
         The attack is considered more noticeable if the AUROC is higher.}
         \label{fig:measure}
\end{figure}

\subsection{\measure}\label{sec:method:measure}

\smallsection{Outline.}
\kijung{To prevent \measure from being bypassed}, we use a learnable edge scorer (\model; see Sec.~\ref{sec:method:training}).
\model output scores indicating how attack-like each edge is.
Unlike the existing measures, \measure is robust to bypassing by learning to distinguish between original and (potential) attack edges, instead of relying on simple rules.
To address the issue of overlooking, we aggregate the scores in an imbalance-aware manner to obtain the final noticeability score.
By using imbalance-aware aggregation, \measure can notice attacks even when the attack rate is very small.

\smallsection{Detailed definition.}
Given \kijung{an unattacked graph $G$,} an attacked graph $\hat{G}$,  and a trained \model on $\hat{G}$ \kijung{(see Sec.~\ref{sec:method:training})}, 
\measure is defined as the Area Under the Receiver Operating Characteristic Curve (AUROC) of the edge scores output by \model as (soft) predictions.
Specifically, let \kijung{$E_0 = \{(u,v) | A_{uv}=1 \text{ or } \hat{A}_{uv} = 1\}$} be the union of edge sets in $G$ and  $\hat{G}$.
Then, we
(1) sort the pairs in $E_0$ w.r.t. their edge scores in descending order,
(2) add the pairs one by one and records the true-positive and false-positive rates, \kijung{forming a curve, and
(3) compute the area under the curve} (see Appendix~\ref{appendix:auroc} for more details).
Since AUROC is based on \textit{rankings} (instead of absolute values) and uses \textit{normalized} values (true-positive and false-positive rates), it is robust to class imbalance, e.g., with a low attack rate.
The more accurately \model distinguishes the original edges in $G$ and the attack edges (in $\hat{G}$ but not in $G$), the higher the AUROC value is, and the more noticeable the attack is considered to be.
This process is illustrated in Fig.~\ref{fig:measure} and detailed in Algorithm~\ref{algorithm:hidenseek} in Appendix~\ref{appendix:auroc}.
\begin{figure}[t!]
     \centering
         \centering
         \includegraphics[width=0.95\linewidth]{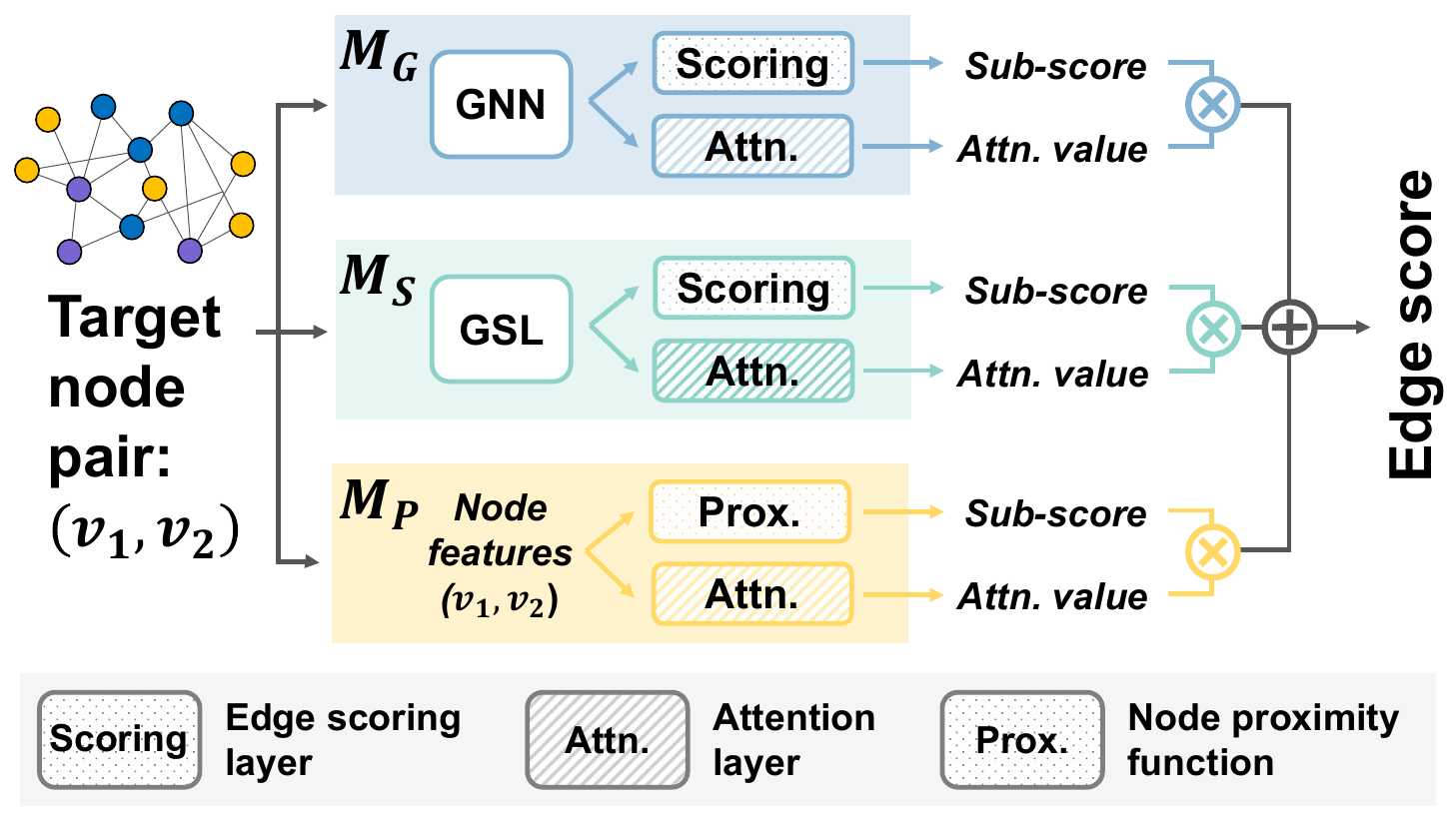} 
         \caption{Structure of \model. \model uses an ensemble model with three modules, 
         a vanilla GNN module ($M_G$), 
         a GSL-based GNN module ($M_S$), and 
         a node proximity module ($M_P$).
         For each node pair, each module learns or computes a sub-score and an attention value,
         and the final score of the pair is the sum of its sub-scores weighted by their attention values.}
        \label{fig:model}
\end{figure}

\subsection{LEO: \underline{L}earnable \underline{E}dge Sc\underline{o}rer}\label{sec:method:training}
As discussed, we use \model to obtain edge scores, which are then aggregated to produce the final \measure scores.
Below, we present how we design \model and how it is trained in a self-supervised manner \textit{without knowing the original unattacked graph}.

\smallsection{Structure.}
The input of \model is an attacked graph $\hat{G}$, and the output is an edge score $f((u, v); \hat{G})$ for each node pair $(u, v)$ indicating how likely $(u, v)$ is an edge in the original graph.
Since \model is trained on the attacked graph, we use an ensemble model with three modules to enhance its robustness.
Each ensemble module uses information from a different perspective.
The first module $M_G$ is a vanilla Graph Neural Network (GNN) module that extracts intrinsic information by leveraging graph topology and node features.
To obtain node embeddings, any GNN architecture, such as Graph Convolutional Networks (GCNs), can be used.
The second module $M_S$ is a graph structure-learning-based (GSL-based) GNN module.
Unlike $M_G$, which directly uses attacked graphs with structural noise, 
$M_S$ uses structural learning to infer latent relationships, building another graph $G_S$ and conducting message passing on $G_S$.
The third module $M_P$ computes scores based on the proximity (e.g., cosine similarity) between each node pair.
The structure of \model is illustrated in Fig.~\ref{fig:model}.

\smallsection{Edge scoring.}
For each node pair $(u, v)$, each module computes a sub-score and its corresponding attention value,
and the final edge score $f((u, v); \hat{G})$ is the sum of the sub-scores weighted by their attention values.
For $M_G$ or $M_S$, the sub-score is computed through an edge scoring layer that takes the node embeddings of the two endpoints as inputs and calculates the sub-score through a bilinear layer, followed by a sigmoid function.
For $M_P$, the sub-score is the proximity value between the two endpoints.
For each module, an attention weight layer 
(1) takes two node embeddings/features as inputs,
(2) concatenates the element-wise mean and element-wise max vectors of them (to ensure permutation-invariance), and
(3) applies a 2-layer MLP to calculate the attention values.
The attention values are normalized using the softmax function.

\smallsection{Training.}
Since \model lacks knowledge of which edges are attacks, we use self-supervised learning for its training.
Specifically, we assign positive pseudo-labels to all edges in the attacked graph $\hat{G}$ and negative pseudo-labels to randomly selected non-edge node pairs.
That is, the positive samples are $T_p=\{(u,v)|\hat{A}_{uv}=1\}$,
and the negative samples $T_n$ are randomly sampled from $\{(u,v)|\hat{A}_{uv}=0\}$, such that $|T_n|=|T_p|$.
Note that $T_p$ contains attack edges, and they may mislead the training of \model.
To alleviate this problem, we adaptively filter out positive pairs with low edge scores in each training epoch.
Specifically, we obtain the filtered positive sample set $T_p(k)$ with the top $k\%$ pairs w.r.t. edge scores 
$f((u,v); \hat{G})$'s.\footnote{\kijung{We do not filter negative pairs since the proportion of fake non-edges (edges originally in $G$ but deleted in $\hat{G}$) is typically low due to the large total number of non-edges.}}
Then, we define the loss function as 
$\sum_{(u,v)\in T_p(k)}{\mathcal{L}(f((u,v);\hat{G}), 1)} + \sum_{(u,v)\in T_n}{\mathcal{L}(f((u,v);\hat{G}), 0)}$,
where $\mathcal{L}$ is the cross-entropy loss.
We additionally use the cross-entropy loss based on sub-scores from each module to stabilize the training of each module.
See Online Appendix \textsc{E.1}~\cite{code} for more details.

\subsection{Extension to Node-Feature Attacks}
While we have focused on topological attacks, \measure can be applied to node-feature attacks. 
In Appendix~\ref{appendix:feature}, we introduce a \textbf{\myuline{l}}earnable \textbf{\myuline{f}}eature sc\textbf{\myuline{o}}rer   (\textsc{LFO}), an extension of LEO designed for node-feature attacks, 
and use it to measure the noticeability score on \textsc{RWCS}~\cite{ma2020towards} and \metattack~\cite{zugner2019adversarial}. 
See Appendix~\ref{appendix:feature} for details.

\section{Empirical Justification}
\label{sec:exp}
\begin{table*}
\centering
\setlength{\tabcolsep}{10pt}
\caption{\label{tab:exp:q1}
(Q1) \underline{\smash{Justification of \model:}} 
\model ranks first under all attack methods in terms of average rank (AR). 
O.O.T.: out of time ($>$ 3 hours). 
O.O.M.: out of (GPU) memory. 
N.A.: not applicable (specifically, the maximum number of attacks allowed by \structack is exceeded). For each setting (each column), the best and second-best results are in \setlength{\fboxsep}{1pt}\colorbox{Red!40}{red} and \setlength{\fboxsep}{1pt}\colorbox{Blue!40}{blue}, respectively.
}
\resizebox{!}{0.48\textheight}{
    \begin{tabular}{c||c||c|c|c|c|c|c||c}
    \hline
    & \multirow{2}{*}{\textbf{Method}} & \multicolumn{6}{c||}{\textbf{Datasets}} & \multirow{2}{*}{\textbf{AR}} \\
    \cline{3-8}
    & & \cora & \citeseer & \coraml & \lastfmasia & \chameleon & \squirrel & \\
    \hline
    \hline
    \multirow{12}{*}{\rotatebox[origin=c]{90}{\random}}
    & \svd & 0.694$\pm$0.011 & 0.636$\pm$0.011 & 0.794$\pm$0.005 & 0.835$\pm$0.002 & 0.937$\pm$0.002 & {\setlength{\fboxsep}{1pt}\colorbox{Blue!40}{0.953$\pm$0.001}} & 5.8 \\
    & \cosine & 0.802$\pm$0.007 & {\setlength{\fboxsep}{1pt}\colorbox{Blue!40}{0.893$\pm$0.006}} & 0.795$\pm$0.005 & 0.835$\pm$0.002 & 0.561$\pm$0.007 & 0.487$\pm$0.001 & 6.2 \\
    \cline{2-9}
    & \degreelg & 0.672$\pm$0.009 & 0.623$\pm$0.012 & 0.759$\pm$0.007 & 0.810$\pm$0.003 & 0.845$\pm$0.003 & 0.892$\pm$0.001 & 8.5 \\
    & \lclg & 0.583$\pm$0.020 & 0.609$\pm$0.011 & 0.613$\pm$0.011 & 0.683$\pm$0.006 & 0.775$\pm$0.003 & 0.851$\pm$0.003 & 10.5 \\
    & \homophilylg & 0.639$\pm$0.011 & 0.638$\pm$0.007 & 0.673$\pm$0.009 & 0.728$\pm$0.006 & 0.694$\pm$0.006 & 0.633$\pm$0.007 & 9.8 \\
    \cline{2-9}
    & \gcn & 0.790$\pm$0.006 & 0.720$\pm$0.123 & {\setlength{\fboxsep}{1pt}\colorbox{Blue!40}{0.826$\pm$0.019}} & 0.911$\pm$0.003 & 0.939$\pm$0.009 & 0.941$\pm$0.002 & {\setlength{\fboxsep}{1pt}\colorbox{Blue!40}{3.8}} \\
    & \gcnsvd & 0.675$\pm$0.018 & 0.722$\pm$0.023 & 0.750$\pm$0.008 & 0.919$\pm$0.002 & 0.907$\pm$0.027 & 0.923$\pm$0.003 & 5.7 \\
    & \rgcn & {\setlength{\fboxsep}{1pt}\colorbox{Blue!40}{0.824$\pm$0.011}} & 0.710$\pm$0.165 & 0.784$\pm$0.154 & 0.916$\pm$0.002 & 0.548$\pm$0.063 & 0.596$\pm$0.050 & 6.8 \\
    & \mediangcn & 0.700$\pm$0.138 & 0.719$\pm$0.132 & 0.736$\pm$0.012 & O.O.T & 0.618$\pm$0.170 & O.O.T & 9.3 \\
    & \gnnguard & 0.760$\pm$0.050 & 0.729$\pm$0.025 & 0.812$\pm$0.004 & {\setlength{\fboxsep}{1pt}\colorbox{Blue!40}{0.922$\pm$0.002}} & 0.884$\pm$0.032 & 0.921$\pm$0.015 & 4.2 \\
    & \metagc & 0.728$\pm$0.007 & 0.704$\pm$0.017 & 0.710$\pm$0.035 & 0.830$\pm$0.007 & {\setlength{\fboxsep}{1pt}\colorbox{Blue!40}{0.970$\pm$0.001}} & 0.949$\pm$0.008 & 6.2 \\
    \cline{2-9}
    & \model(Proposed) & {\setlength{\fboxsep}{1pt}\colorbox{Red!40}{0.894$\pm$0.011}} & {\setlength{\fboxsep}{1pt}\colorbox{Red!40}{0.926$\pm$0.006}} & {\setlength{\fboxsep}{1pt}\colorbox{Red!40}{0.914$\pm$0.004}} & {\setlength{\fboxsep}{1pt}\colorbox{Red!40}{0.931$\pm$0.000}} & {\setlength{\fboxsep}{1pt}\colorbox{Red!40}{0.973$\pm$0.003}} & {\setlength{\fboxsep}{1pt}\colorbox{Red!40}{0.976$\pm$0.001}} & {\setlength{\fboxsep}{1pt}\colorbox{Red!40}{\textbf{1.0}}} \\
    \hline
    \hline
    \multirow{12}{*}{\rotatebox[origin=c]{90}{\dice}}
    & \svd & 0.671$\pm$0.010 & 0.619$\pm$0.019 & 0.776$\pm$0.007 & 0.827$\pm$0.002 & 0.910$\pm$0.005 & {\setlength{\fboxsep}{1pt}\colorbox{Blue!40}{0.940$\pm$0.001}} & 6.8 \\
    & \cosine & 0.809$\pm$0.011 & {\setlength{\fboxsep}{1pt}\colorbox{Blue!40}{0.896$\pm$0.009}} & 0.805$\pm$0.008 & 0.842$\pm$0.005 & 0.562$\pm$0.008 & 0.480$\pm$0.003 & 6.3 \\
    \cline{2-9}
    & \degreelg & 0.663$\pm$0.007 & 0.622$\pm$0.014 & 0.762$\pm$0.007 & 0.811$\pm$0.004 & 0.844$\pm$0.008 & 0.886$\pm$0.002 & 8.3 \\
    & \lclg & 0.587$\pm$0.014 & 0.615$\pm$0.005 & 0.614$\pm$0.010 & 0.684$\pm$0.005 & 0.776$\pm$0.012 & 0.843$\pm$0.003 & 10.5 \\
    & \homophilylg & 0.645$\pm$0.014 & 0.649$\pm$0.017 & 0.686$\pm$0.007 & 0.732$\pm$0.010 & 0.693$\pm$0.017 & 0.632$\pm$0.010 & 9.5 \\
    \cline{2-9}
    & \gcn & 0.779$\pm$0.021 & 0.675$\pm$0.163 & 0.835$\pm$0.019 & 0.917$\pm$0.005 & {\setlength{\fboxsep}{1pt}\colorbox{Blue!40}{0.944$\pm$0.009}} & 0.938$\pm$0.002 & {\setlength{\fboxsep}{1pt}\colorbox{Blue!40}{4.2}} \\
    & \gcnsvd & 0.677$\pm$0.008 & 0.645$\pm$0.136 & 0.762$\pm$0.008 & {\setlength{\fboxsep}{1pt}\colorbox{Blue!40}{0.932$\pm$0.002}} & 0.929$\pm$0.009 & 0.922$\pm$0.002 & 5.7 \\
    & \rgcn & {\setlength{\fboxsep}{1pt}\colorbox{Blue!40}{0.834$\pm$0.021}} & 0.725$\pm$0.165 & {\setlength{\fboxsep}{1pt}\colorbox{Blue!40}{0.847$\pm$0.058}} & 0.918$\pm$0.003 & 0.543$\pm$0.049 & 0.552$\pm$0.019 & 5.8 \\
    & \mediangcn & 0.799$\pm$0.030 & 0.788$\pm$0.014 & 0.758$\pm$0.017 & O.O.T & 0.620$\pm$0.102 & O.O.T & 8.5 \\
    & \gnnguard & 0.714$\pm$0.059 & 0.800$\pm$0.022 & 0.828$\pm$0.007 & 0.930$\pm$0.005 & 0.917$\pm$0.048 & 0.904$\pm$0.002 & 4.7 \\
    & \metagc & 0.726$\pm$0.028 & 0.715$\pm$0.032 & 0.674$\pm$0.045 & 0.814$\pm$0.010 & 0.923$\pm$0.003 & {\setlength{\fboxsep}{1pt}\colorbox{Red!40}{0.947$\pm$0.008}} & 6.0 \\
    \cline{2-9}
    & \model(Proposed) & {\setlength{\fboxsep}{1pt}\colorbox{Red!40}{0.899$\pm$0.011}} & {\setlength{\fboxsep}{1pt}\colorbox{Red!40}{0.942$\pm$0.008}} & {\setlength{\fboxsep}{1pt}\colorbox{Red!40}{0.932$\pm$0.003}} & {\setlength{\fboxsep}{1pt}\colorbox{Red!40}{0.941$\pm$0.003}} & {\setlength{\fboxsep}{1pt}\colorbox{Red!40}{0.960$\pm$0.003}} & 0.938$\pm$0.003 & {\setlength{\fboxsep}{1pt}\colorbox{Red!40}{\textbf{1.3}}} \\
    \hline
    \hline
    \multirow{12}{*}{\rotatebox[origin=c]{90}{\pgd}}
    & \svd & 0.799$\pm$0.008 & 0.725$\pm$0.005 & {\setlength{\fboxsep}{1pt}\colorbox{Blue!40}{0.881$\pm$0.003}} & 0.308$\pm$0.040 & {\setlength{\fboxsep}{1pt}\colorbox{Blue!40}{0.913$\pm$0.010}} & 0.943$\pm$0.006 & 4.5 \\
    & \cosine & {\setlength{\fboxsep}{1pt}\colorbox{Blue!40}{0.809$\pm$0.005}} & {\setlength{\fboxsep}{1pt}\colorbox{Blue!40}{0.882$\pm$0.006}} & 0.812$\pm$0.008 & 0.842$\pm$0.009 & 0.586$\pm$0.002 & 0.524$\pm$0.004 & 5.7 \\
    \cline{2-9}
    & \degreelg & 0.738$\pm$0.009 & 0.702$\pm$0.011 & 0.866$\pm$0.003 & 0.252$\pm$0.035 & {\setlength{\fboxsep}{1pt}\colorbox{Red!40}{0.943$\pm$0.003}} & {\setlength{\fboxsep}{1pt}\colorbox{Blue!40}{0.951$\pm$0.006}} & 5.5 \\
    & \lclg & 0.630$\pm$0.065 & 0.707$\pm$0.015 & 0.703$\pm$0.034 & 0.697$\pm$0.019 & 0.886$\pm$0.016 & {\setlength{\fboxsep}{1pt}\colorbox{Red!40}{0.955$\pm$0.004}} & 7.2 \\
    & \homophilylg & 0.731$\pm$0.014 & 0.719$\pm$0.009 & 0.808$\pm$0.014 & {\setlength{\fboxsep}{1pt}\colorbox{Red!40}{0.896$\pm$0.005}} & 0.829$\pm$0.008 & 0.759$\pm$0.014 & 6.8 \\
    \cline{2-9}
    & \gcn & 0.644$\pm$0.012 & 0.754$\pm$0.016 & 0.762$\pm$0.020 & 0.560$\pm$0.033 & 0.899$\pm$0.017 & 0.900$\pm$0.003 & 7.0 \\
    & \gcnsvd & 0.757$\pm$0.018 & 0.777$\pm$0.014 & 0.853$\pm$0.009 & 0.527$\pm$0.027 & 0.906$\pm$0.012 & 0.936$\pm$0.005 & {\setlength{\fboxsep}{1pt}\colorbox{Blue!40}{4.3}} \\
    & \rgcn & 0.586$\pm$0.042 & 0.738$\pm$0.118 & 0.702$\pm$0.111 & 0.454$\pm$0.033 & 0.536$\pm$0.062 & 0.601$\pm$0.061 & 10.0 \\
    & \mediangcn & 0.649$\pm$0.063 & 0.716$\pm$0.094 & 0.718$\pm$0.035 & O.O.T & 0.868$\pm$0.064 & O.O.T & 10.2 \\
    & \gnnguard & 0.738$\pm$0.008 & 0.759$\pm$0.008 & 0.785$\pm$0.013 & 0.428$\pm$0.014 & 0.907$\pm$0.009 & 0.911$\pm$0.010 & 5.8 \\
    & \metagc & 0.713$\pm$0.018 & 0.694$\pm$0.022 & 0.759$\pm$0.028 & 0.602$\pm$0.034 & 0.875$\pm$0.010 & 0.925$\pm$0.007 & 8.0 \\
    \cline{2-9}
    & \model(Proposed) & {\setlength{\fboxsep}{1pt}\colorbox{Red!40}{0.874$\pm$0.004}} & {\setlength{\fboxsep}{1pt}\colorbox{Red!40}{0.931$\pm$0.008}} & {\setlength{\fboxsep}{1pt}\colorbox{Red!40}{0.930$\pm$0.011}} & {\setlength{\fboxsep}{1pt}\colorbox{Blue!40}{0.853$\pm$0.008}} & 0.880$\pm$0.012 & 0.926$\pm$0.019 & {\setlength{\fboxsep}{1pt}\colorbox{Red!40}{\textbf{2.8}}} \\
    \hline
    \hline
    \multirow{12}{*}{\rotatebox[origin=c]{90}{\structack}}
    & \svd & 0.823$\pm$0.000 & 0.764$\pm$0.000 & 0.879$\pm$0.000 & 0.889$\pm$0.000 & 0.948$\pm$0.000 & N.A & 4.8 \\
    & \cosine & 0.842$\pm$0.000 & {\setlength{\fboxsep}{1pt}\colorbox{Blue!40}{0.886$\pm$0.000}} & 0.823$\pm$0.000 & 0.854$\pm$0.000 & 0.565$\pm$0.000 & N.A & 6.2 \\
    \cline{2-9}
    & \degreelg & {\setlength{\fboxsep}{1pt}\colorbox{Blue!40}{0.880$\pm$0.004}} & 0.837$\pm$0.004 & {\setlength{\fboxsep}{1pt}\colorbox{Blue!40}{0.924$\pm$0.002}} & 0.928$\pm$0.001 & 0.852$\pm$0.001 & N.A & {\setlength{\fboxsep}{1pt}\colorbox{Blue!40}{3.4}} \\
    & \lclg & 0.590$\pm$0.007 & 0.702$\pm$0.004 & 0.609$\pm$0.006 & 0.659$\pm$0.002 & 0.736$\pm$0.004 & N.A & 10.2 \\
    & \homophilylg & 0.730$\pm$0.007 & 0.744$\pm$0.006 & 0.786$\pm$0.004 & 0.787$\pm$0.003 & 0.705$\pm$0.005 & N.A & 9.0 \\
    \cline{2-9}
    & \gcn & 0.819$\pm$0.023 & 0.680$\pm$0.165 & 0.894$\pm$0.014 & 0.868$\pm$0.006 & 0.942$\pm$0.003 & N.A & 6.2 \\
    & \gcnsvd & 0.819$\pm$0.018 & 0.799$\pm$0.011 & 0.881$\pm$0.007 & {\setlength{\fboxsep}{1pt}\colorbox{Blue!40}{0.929$\pm$0.002}} & 0.910$\pm$0.019 & N.A & 4.2 \\
    & \rgcn & 0.800$\pm$0.011 & 0.526$\pm$0.057 & 0.656$\pm$0.186 & 0.887$\pm$0.002 & 0.544$\pm$0.022 & N.A & 9.4 \\
    & \mediangcn & 0.754$\pm$0.091 & 0.685$\pm$0.117 & 0.606$\pm$0.040 & O.O.T & 0.500$\pm$0.000 & N.A & 11.0 \\
    & \gnnguard & 0.858$\pm$0.007 & 0.785$\pm$0.005 & 0.866$\pm$0.013 & 0.893$\pm$0.002 & 0.888$\pm$0.032 & N.A & 4.8 \\
    & \metagc & 0.750$\pm$0.007 & 0.724$\pm$0.027 & 0.742$\pm$0.033 & 0.852$\pm$0.008 & {\setlength{\fboxsep}{1pt}\colorbox{Red!40}{0.968$\pm$0.002}} & N.A & 7.4 \\
    \cline{2-9}
    & \model(Proposed) & {\setlength{\fboxsep}{1pt}\colorbox{Red!40}{0.941$\pm$0.004}} & {\setlength{\fboxsep}{1pt}\colorbox{Red!40}{0.950$\pm$0.006}} & {\setlength{\fboxsep}{1pt}\colorbox{Red!40}{0.965$\pm$0.002}} & {\setlength{\fboxsep}{1pt}\colorbox{Red!40}{0.939$\pm$0.002}} & {\setlength{\fboxsep}{1pt}\colorbox{Red!40}{0.968$\pm$0.003}} & N.A & {\setlength{\fboxsep}{1pt}\colorbox{Red!40}{\textbf{1.0}}} \\
    \hline
    \hline
    \multirow{12}{*}{\rotatebox[origin=c]{90}{\metattack}}
    & \svd & 0.691$\pm$0.011 & 0.589$\pm$0.015 & 0.726$\pm$0.014 & O.O.M & 0.940$\pm$0.005 & {\setlength{\fboxsep}{1pt}\colorbox{Red!40}{0.972$\pm$0.002}} & 5.8 \\
    & \cosine & {\setlength{\fboxsep}{1pt}\colorbox{Blue!40}{0.840$\pm$0.004}} & {\setlength{\fboxsep}{1pt}\colorbox{Blue!40}{0.886$\pm$0.005}} & {\setlength{\fboxsep}{1pt}\colorbox{Blue!40}{0.791$\pm$0.011}} & O.O.M & 0.554$\pm$0.005 & 0.454$\pm$0.003 & 5.8 \\
    \cline{2-9}
    & \degreelg & 0.549$\pm$0.006 & 0.461$\pm$0.011 & 0.646$\pm$0.007 & O.O.M & 0.950$\pm$0.002 & 0.969$\pm$0.001 & 8.2 \\
    & \lclg & 0.644$\pm$0.075 & 0.708$\pm$0.010 & 0.781$\pm$0.052 & O.O.M & 0.926$\pm$0.002 & {\setlength{\fboxsep}{1pt}\colorbox{Blue!40}{0.970$\pm$0.001}} & {\setlength{\fboxsep}{1pt}\colorbox{Blue!40}{5.4}} \\
    & \homophilylg & 0.682$\pm$0.021 & 0.649$\pm$0.012 & 0.763$\pm$0.011 & O.O.M & 0.765$\pm$0.022 & 0.622$\pm$0.011 & 7.8 \\
    \cline{2-9}
    & \gcn & 0.696$\pm$0.025 & 0.617$\pm$0.067 & 0.613$\pm$0.016 & O.O.M & 0.939$\pm$0.005 & 0.901$\pm$0.009 & 8.2 \\
    & \gcnsvd & 0.673$\pm$0.017 & 0.696$\pm$0.019 & 0.703$\pm$0.012 & O.O.M & 0.955$\pm$0.006 & 0.921$\pm$0.022 & 6.0 \\
    & \rgcn & 0.773$\pm$0.034 & 0.551$\pm$0.099 & 0.635$\pm$0.102 & O.O.M & 0.579$\pm$0.029 & 0.582$\pm$0.034 & 9.0 \\
    & \mediangcn & 0.744$\pm$0.145 & 0.666$\pm$0.099 & 0.723$\pm$0.031 & O.O.M & 0.675$\pm$0.162 & O.O.T & 8.0 \\
    & \gnnguard & 0.762$\pm$0.018 & 0.685$\pm$0.018 & 0.561$\pm$0.025 & O.O.M & 0.940$\pm$0.004 & 0.919$\pm$0.003 & 6.6 \\
    & \metagc & 0.695$\pm$0.034 & 0.680$\pm$0.025 & 0.664$\pm$0.049 & O.O.M & {\setlength{\fboxsep}{1pt}\colorbox{Blue!40}{0.958$\pm$0.004}} & 0.932$\pm$0.011 & 5.6 \\
    \cline{2-9}
    & \model(Proposed) & {\setlength{\fboxsep}{1pt}\colorbox{Red!40}{0.894$\pm$0.003}} & {\setlength{\fboxsep}{1pt}\colorbox{Red!40}{0.893$\pm$0.010}} & {\setlength{\fboxsep}{1pt}\colorbox{Red!40}{0.856$\pm$0.016}} & O.O.M & {\setlength{\fboxsep}{1pt}\colorbox{Red!40}{0.963$\pm$0.004}} & {\setlength{\fboxsep}{1pt}\colorbox{Blue!40}{0.970$\pm$0.003}} & {\setlength{\fboxsep}{1pt}\colorbox{Red!40}{\textbf{1.2}}} \\
    \hline
    \hline
    \end{tabular}
}
\end{table*}

In this section, we evaluate \measure to answer the Q1-Q3:
\begin{enumerate}[leftmargin=2.5em,topsep=0pt]
    \item[\textbf{Q1.}] \textbf{Justification of \model}: Does \model accurately distinguish attack edges, compared to baselines?
    \item[\textbf{Q2.}] \textbf{Effectiveness of \measure}: Does \measure overcome limitations suffered by existing measures?
    \item[\textbf{Q3.}] \textbf{Additional Application}: Can abnormal edge scores obtained from \model enhance the performance of various GNNs?
\end{enumerate}

\begin{table}[h!]
	\begin{center}
		\caption{\label{tab:dataset} Summary of the real-world datasets}
		\small
		\scalebox{0.9}{
			\begin{tabular}{r|r|r|r|r|r}
				\toprule 
				\textbf{Name} & \textbf{\# Nodes} & \textbf{\# Edges} & \textbf{\# Attributes} & \textbf{\# Classes} & \textbf{Homphily}\\
				\midrule
				Cora & 2,708 & 5,278& 1,433 & 7 & 0.81 \\
				Cora-ML & 2,995 & 8,158& 2,879 & 7 & 0.79 \\
			    Citeseer & 3,379 & 4,552& 3,703 & 6 & 0.74  \\
			    LastFMAsia & 7,624& 27,806 & 128 & 18 & 0.87  \\
				Chameleon & 890 & 8,854&  2,325 & 5 & 0.24  \\
                Squirrel & 2,223 & 46,998&  2,089 & 5 & 0.21  \\
				\bottomrule
			\end{tabular}
		}
	\end{center}
\end{table}

\begingroup
\newcolumntype{C}{>{\centering\arraybackslash}m{0.17\linewidth}}
\begin{table*}
    \caption{ \label{tab:exp:q2_pgd}(Q2 \& Q3) \underline{\smash{Effectiveness of \measure:}} 
    \measure is less bypassable and more sensitive than all the existing measures. 
    For each measure, 
    the \textcolor{SkyBlue}{\textbf{sky blue curves}} \kijung{from original attacks} and \textcolor{DarkBlue}{\textbf{dark blue curves}} from \kijung{adaptive attacks} show the trade-offs between the noticeability scores and the attack performance as we add attack edges.
    \kijung{The gaps between the two curves tend to be smallest for \measure, indicating that it is least bypassable.}
    Red stars (\redstar{$\medstar$}) indicate the point of a low attack rate $\gamma = 2\%$, \kijung{where in most cases, noticeability (i.e., $y$ values) is significant only w.r.t. \measure, indicating that \measure is sensitive.}
    }        
    \centering    
    \begin{tabular}{l||C|C|C|C||C}    
        \hline
        & \textsc{DegreeKS} & \textsc{ClsCoefKS} & \textsc{DegreeLR} & \textsc{HomophKS} & \textbf{\measure} \\
        \hline\hline
        \rotatebox[origin=c]{90}{\cora} & 
        \includegraphics[width=\linewidth]{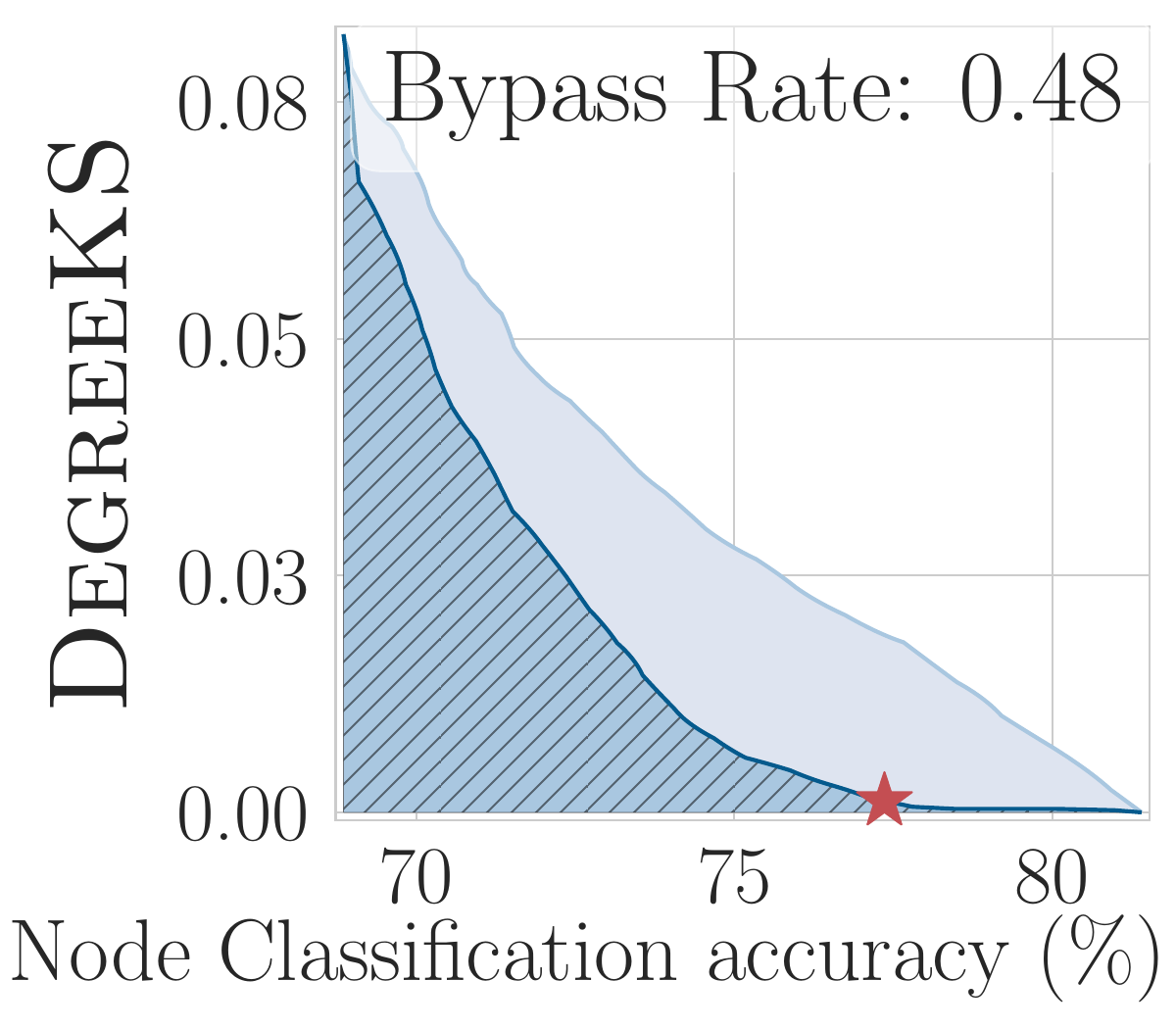} & 
        \includegraphics[width=\linewidth]{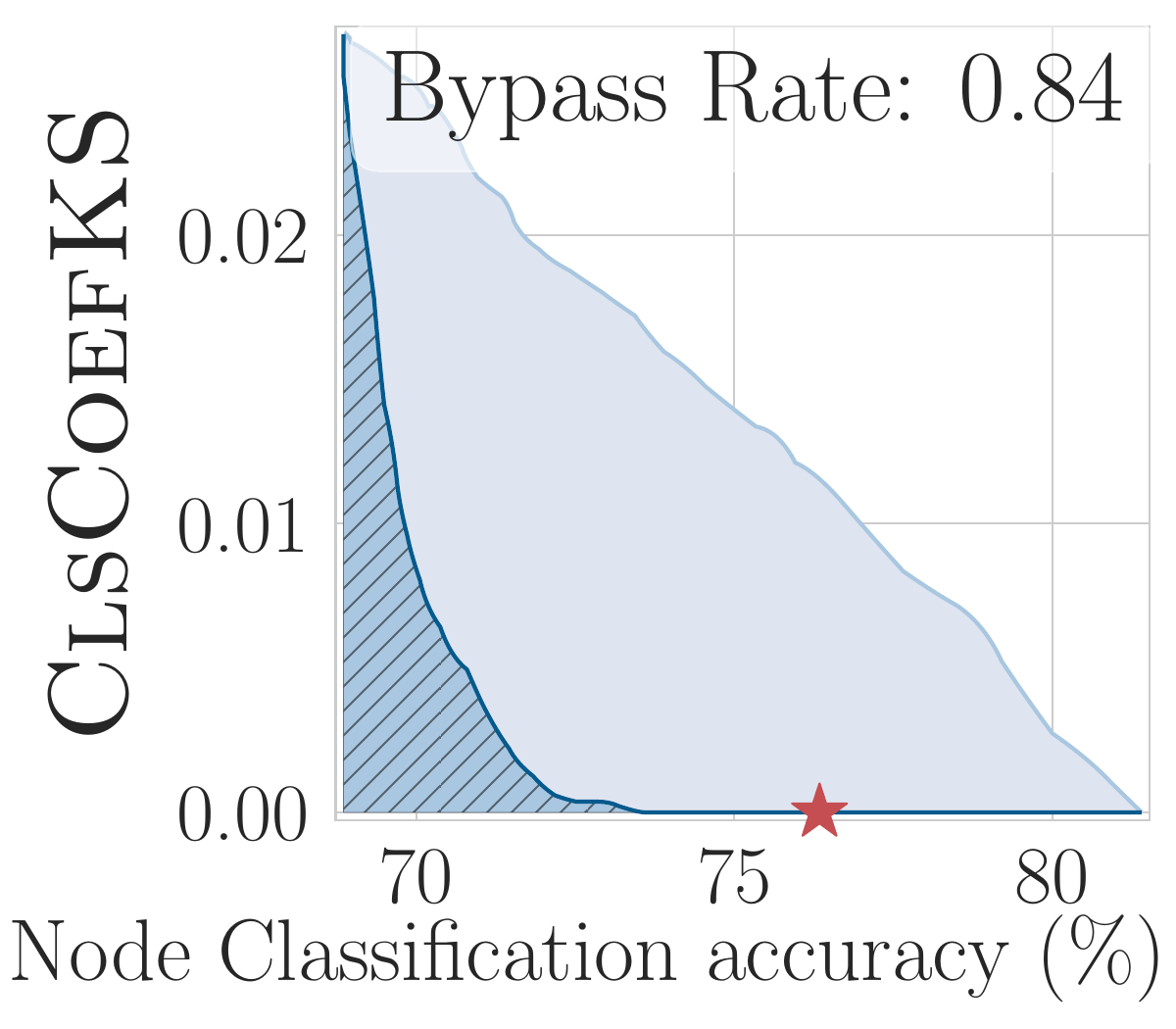} & 
        \includegraphics[width=\linewidth]{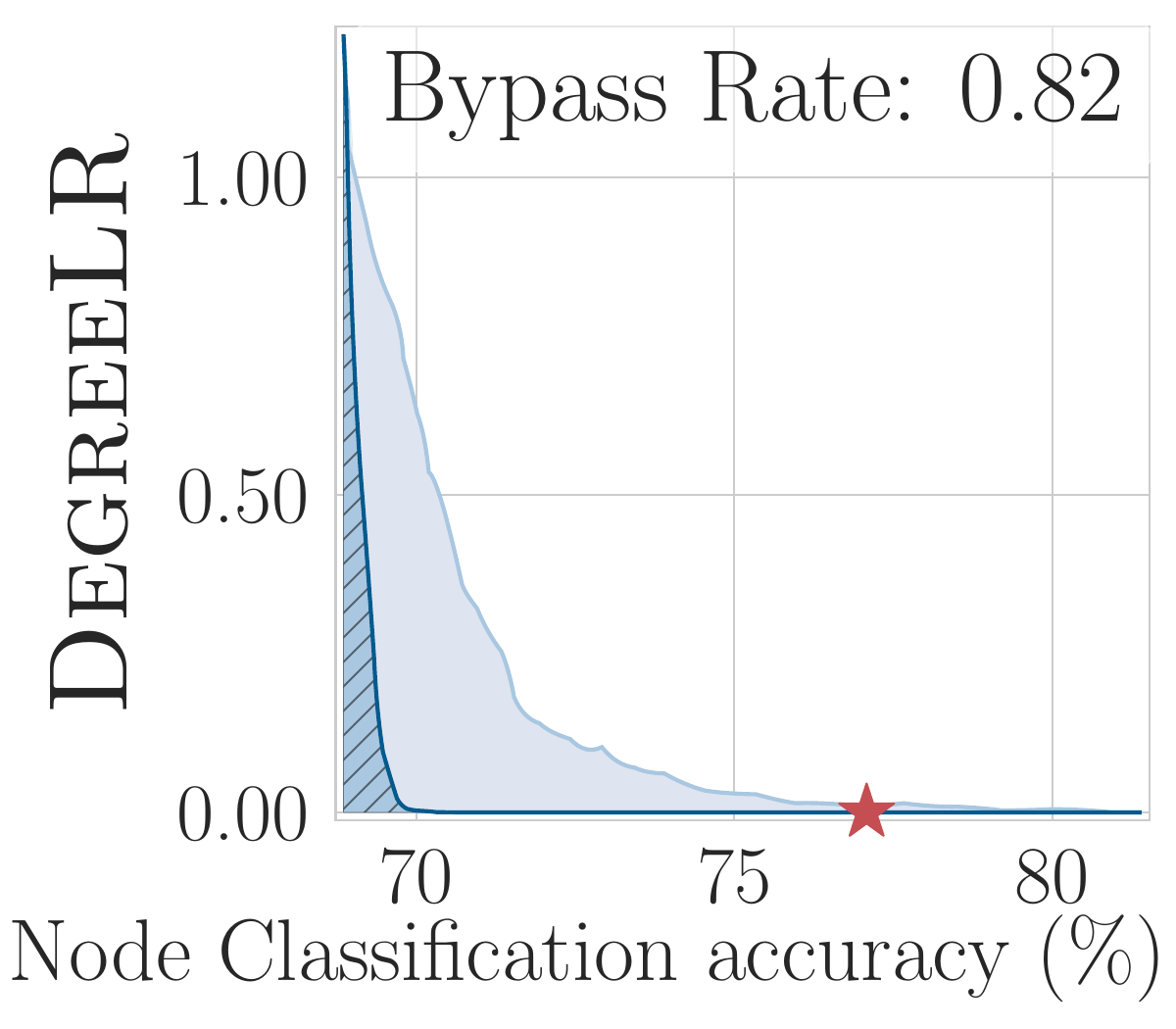} & 
        \includegraphics[width=\linewidth]{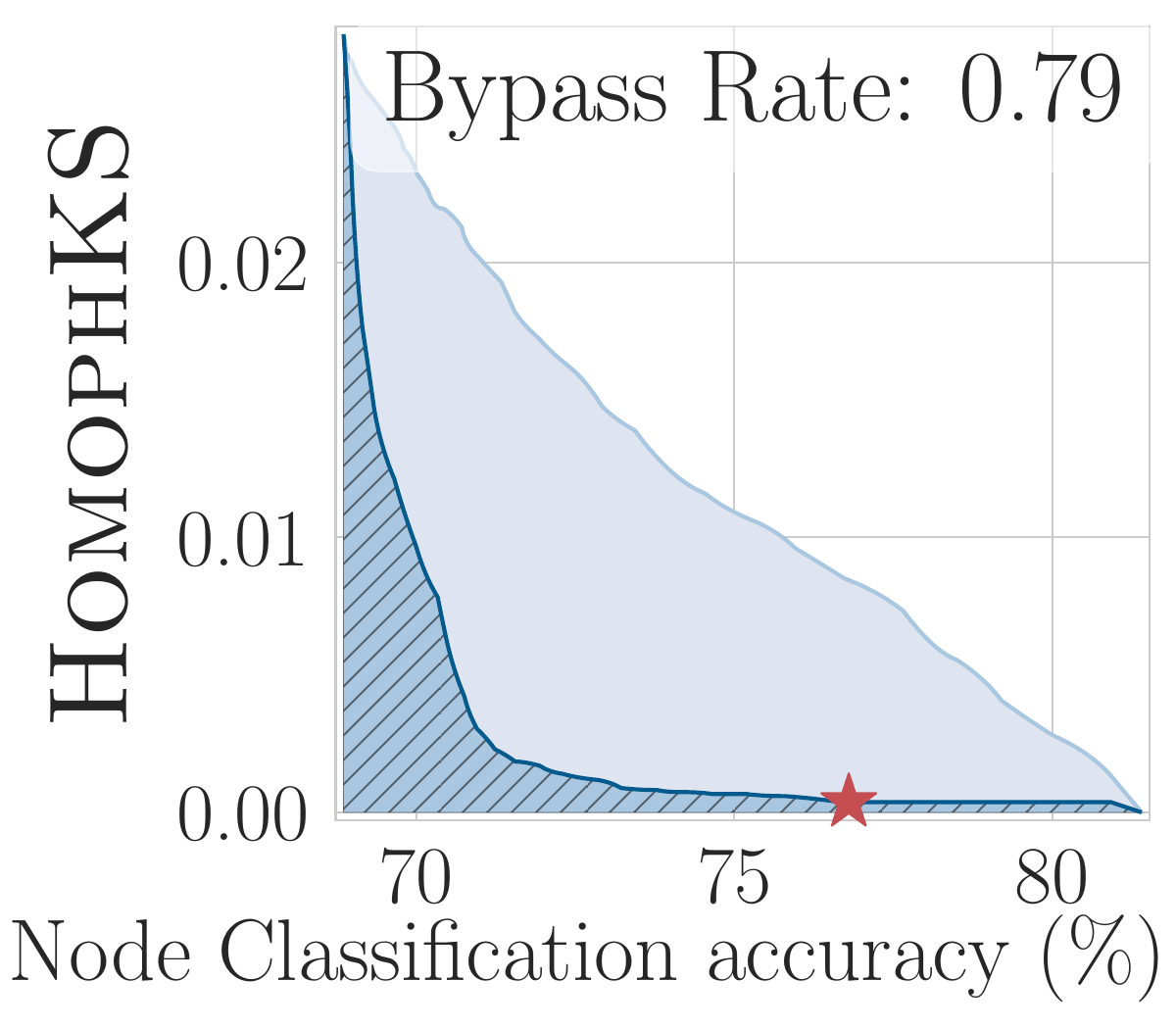} &
        \includegraphics[width=\linewidth]{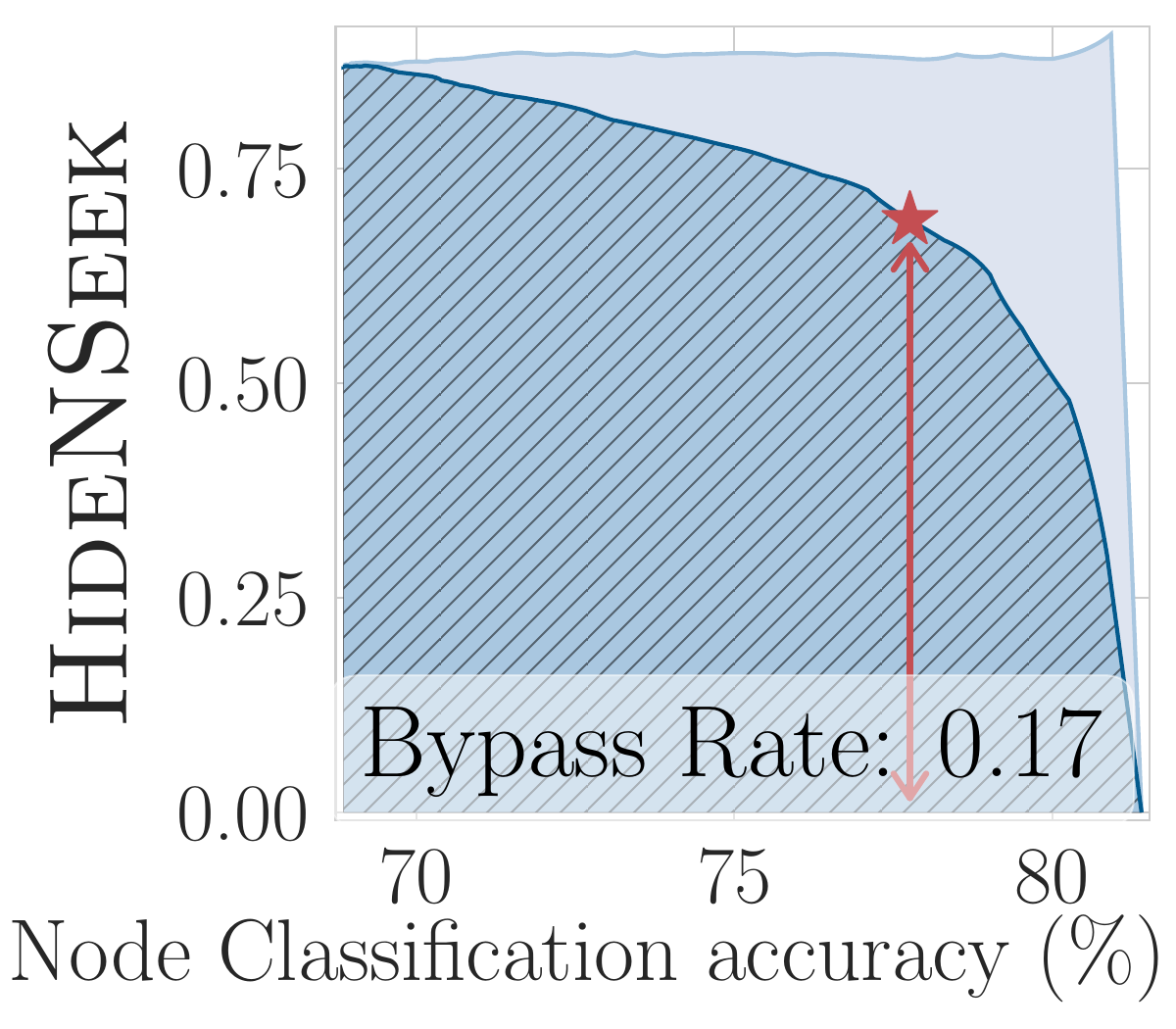} \\
        \hline
        \rotatebox[origin=c]{90}{\citeseer} & 
        \includegraphics[width=\linewidth]{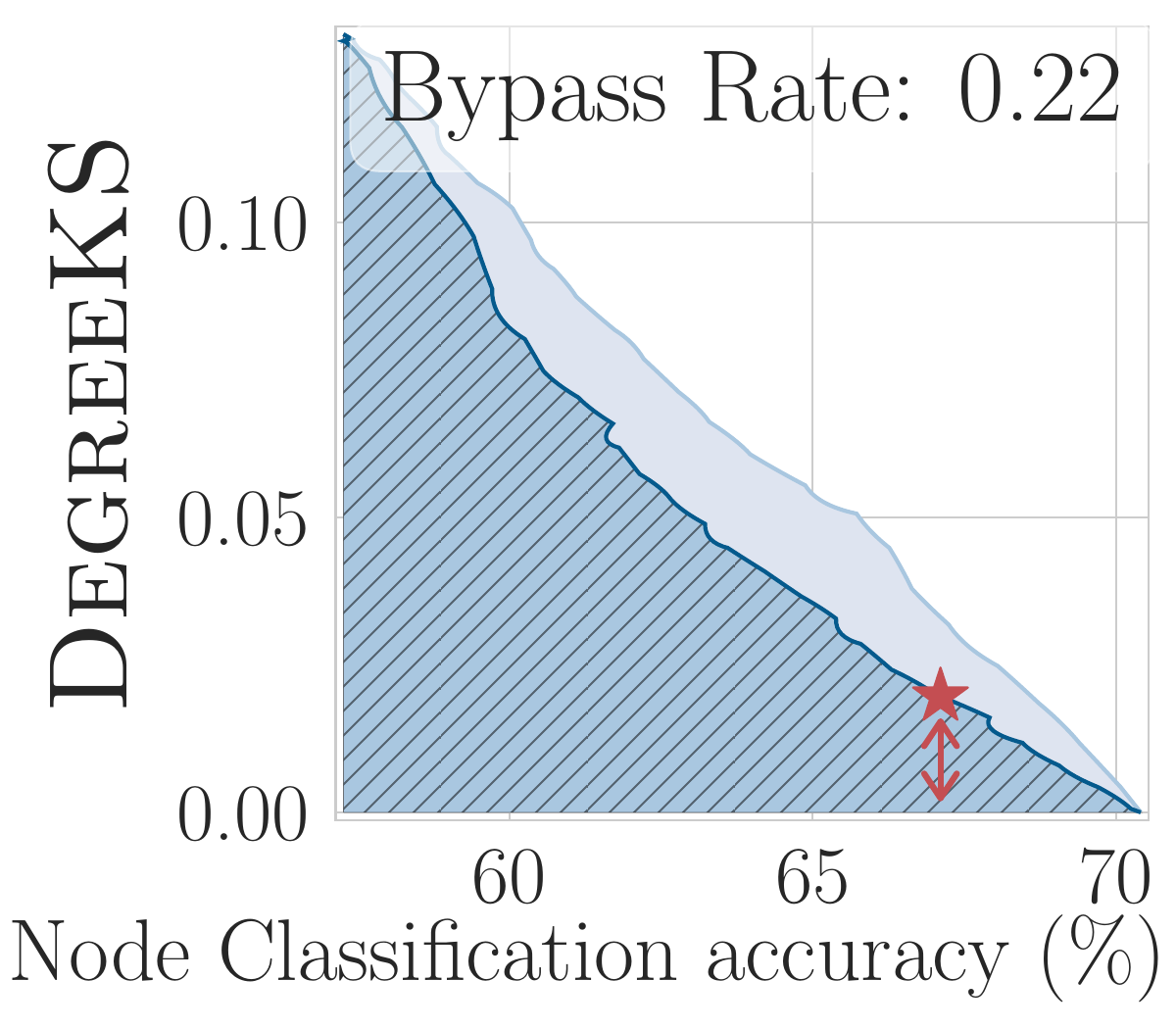} & 
        \includegraphics[width=\linewidth]{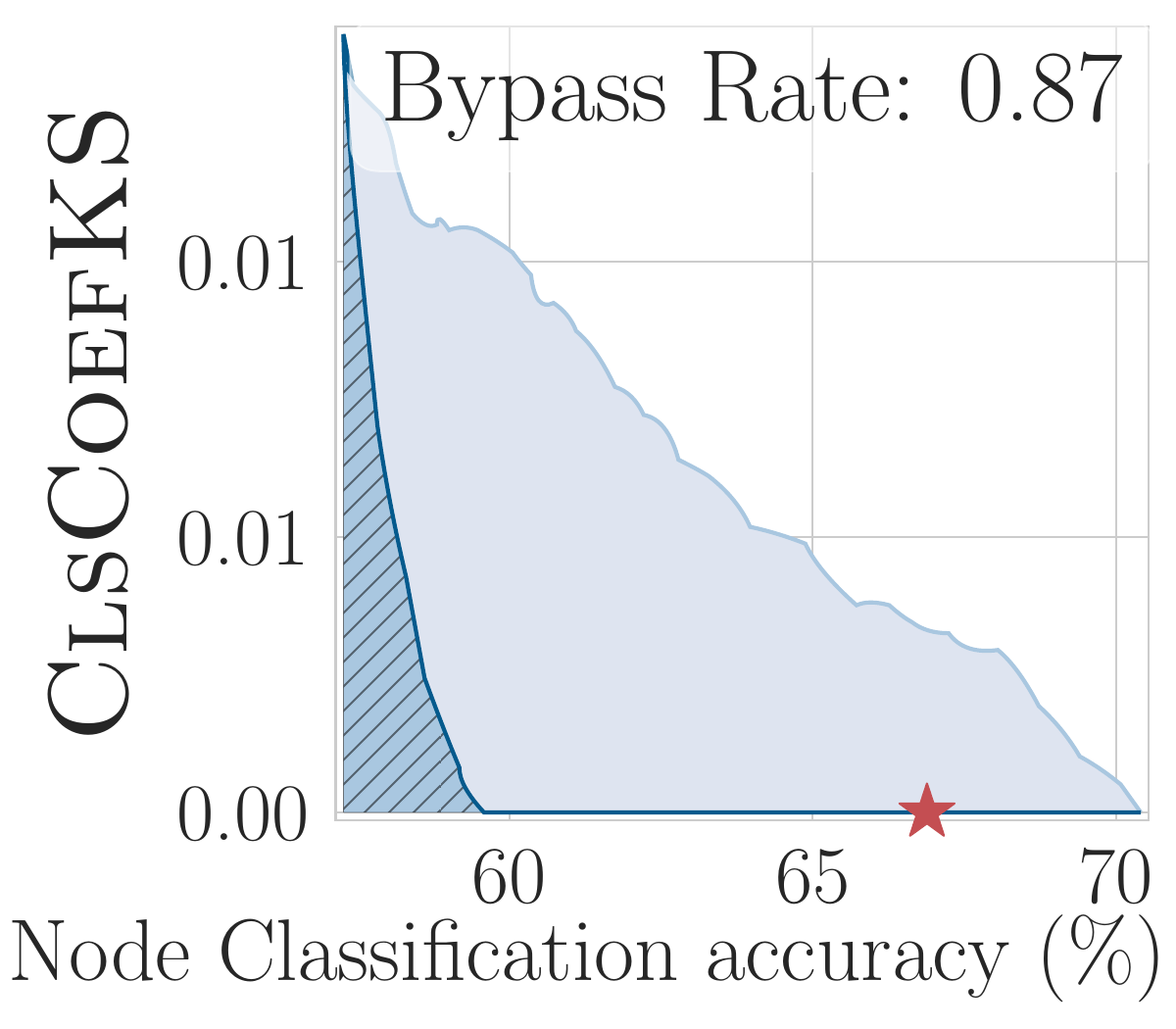} & 
        \includegraphics[width=\linewidth]{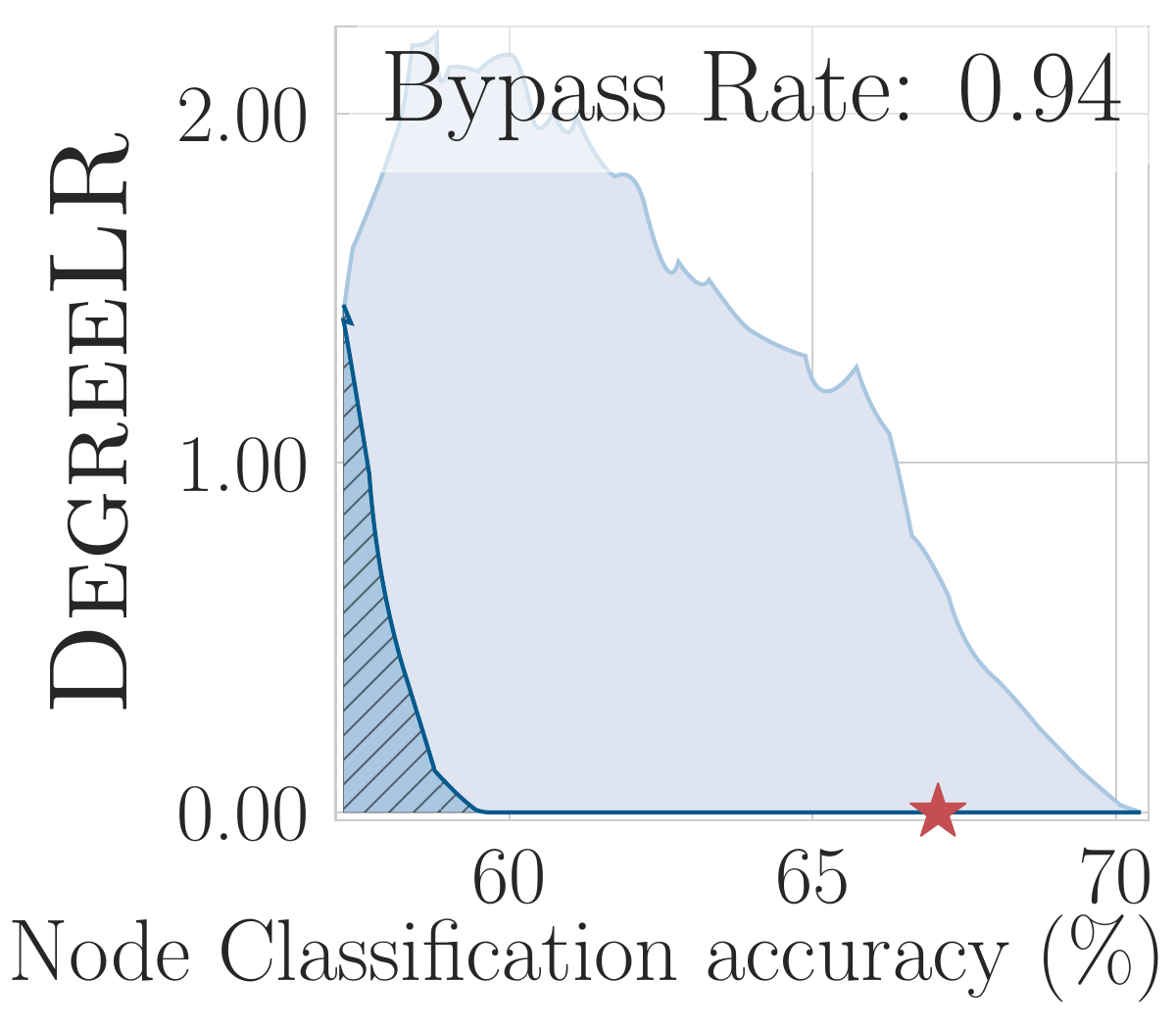} & 
        \includegraphics[width=\linewidth]{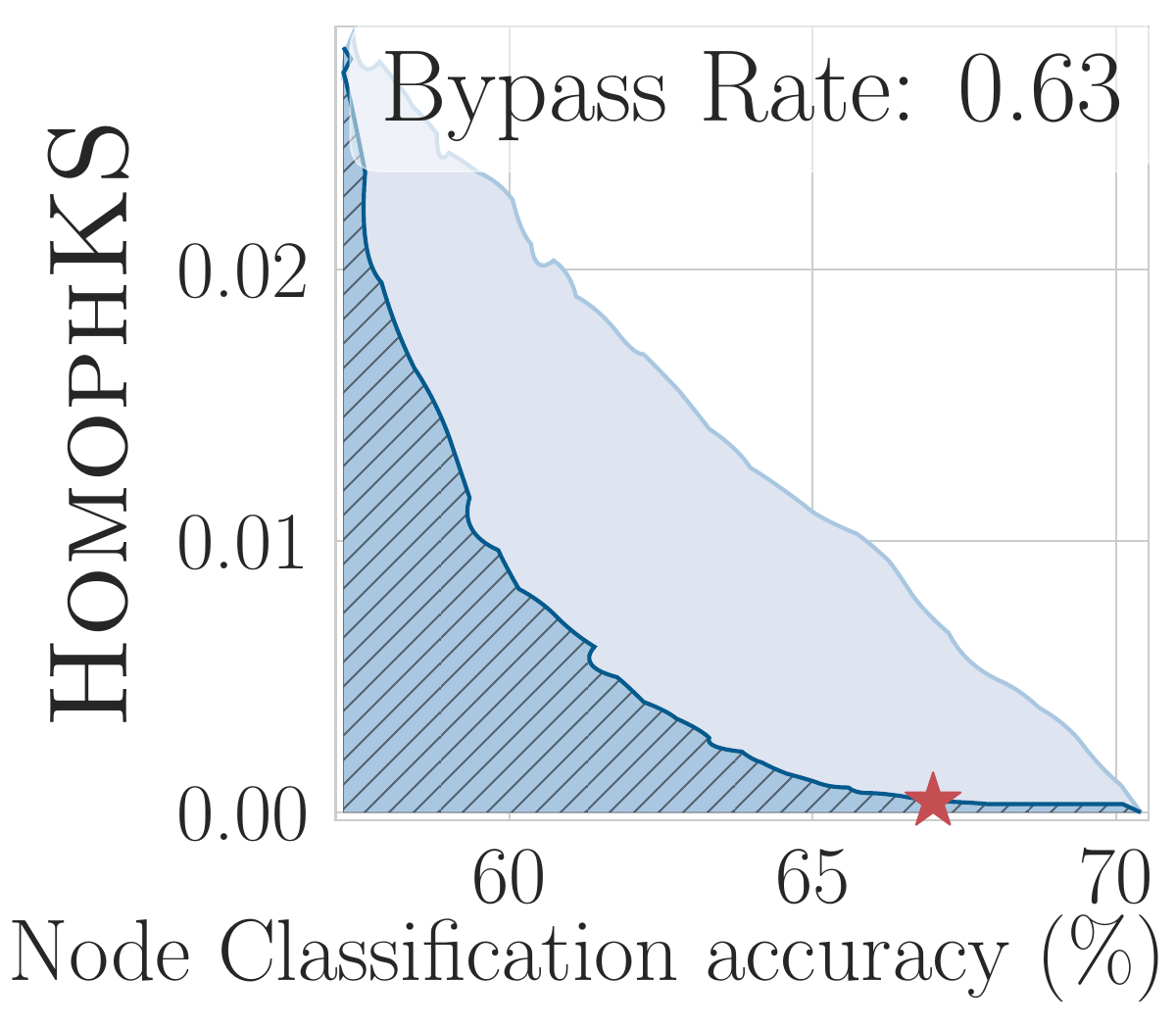} &
        \includegraphics[width=\linewidth]{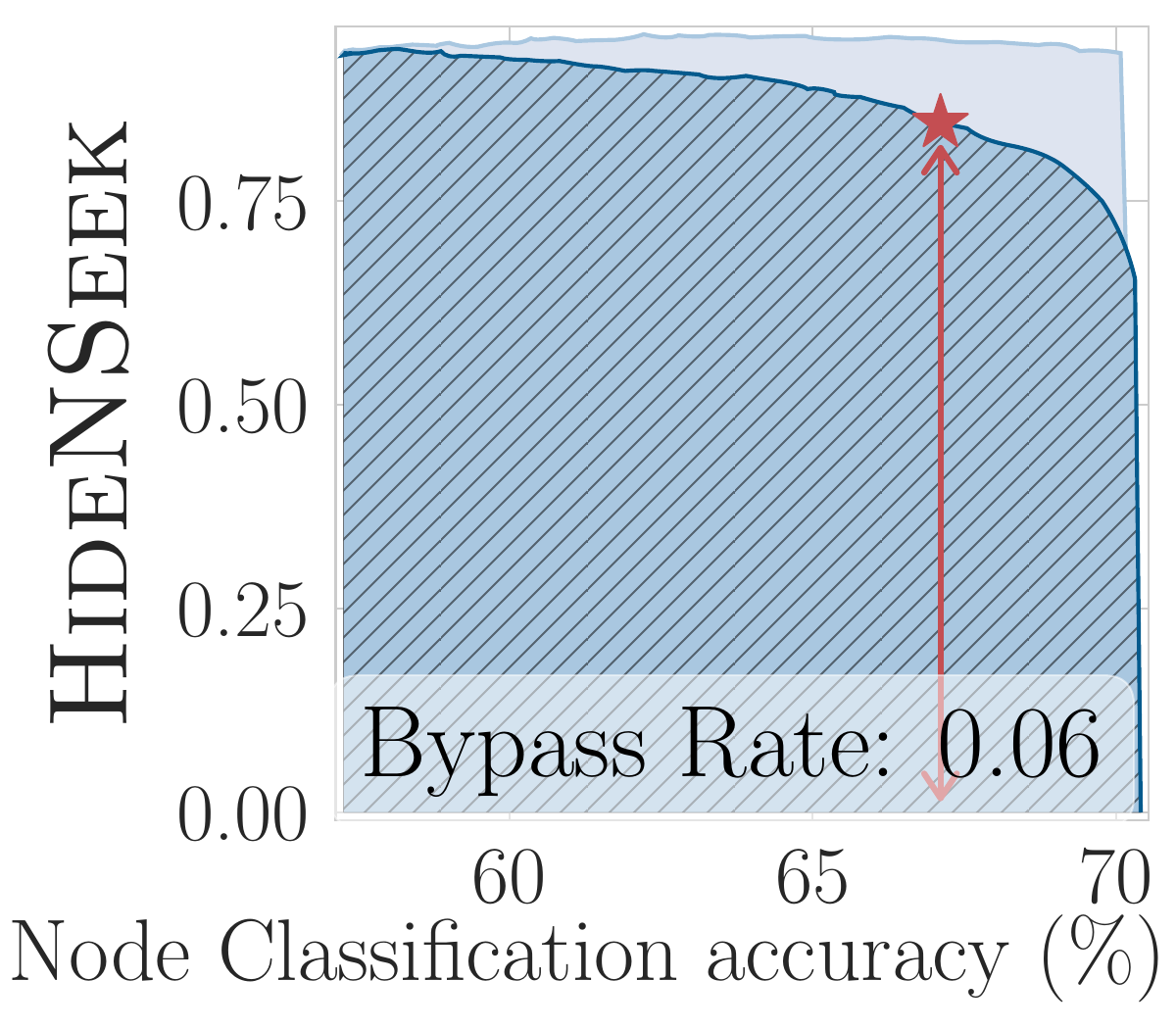} \\
        \hline
        \rotatebox[origin=c]{90}{\coraml} & 
        \includegraphics[width=\linewidth]{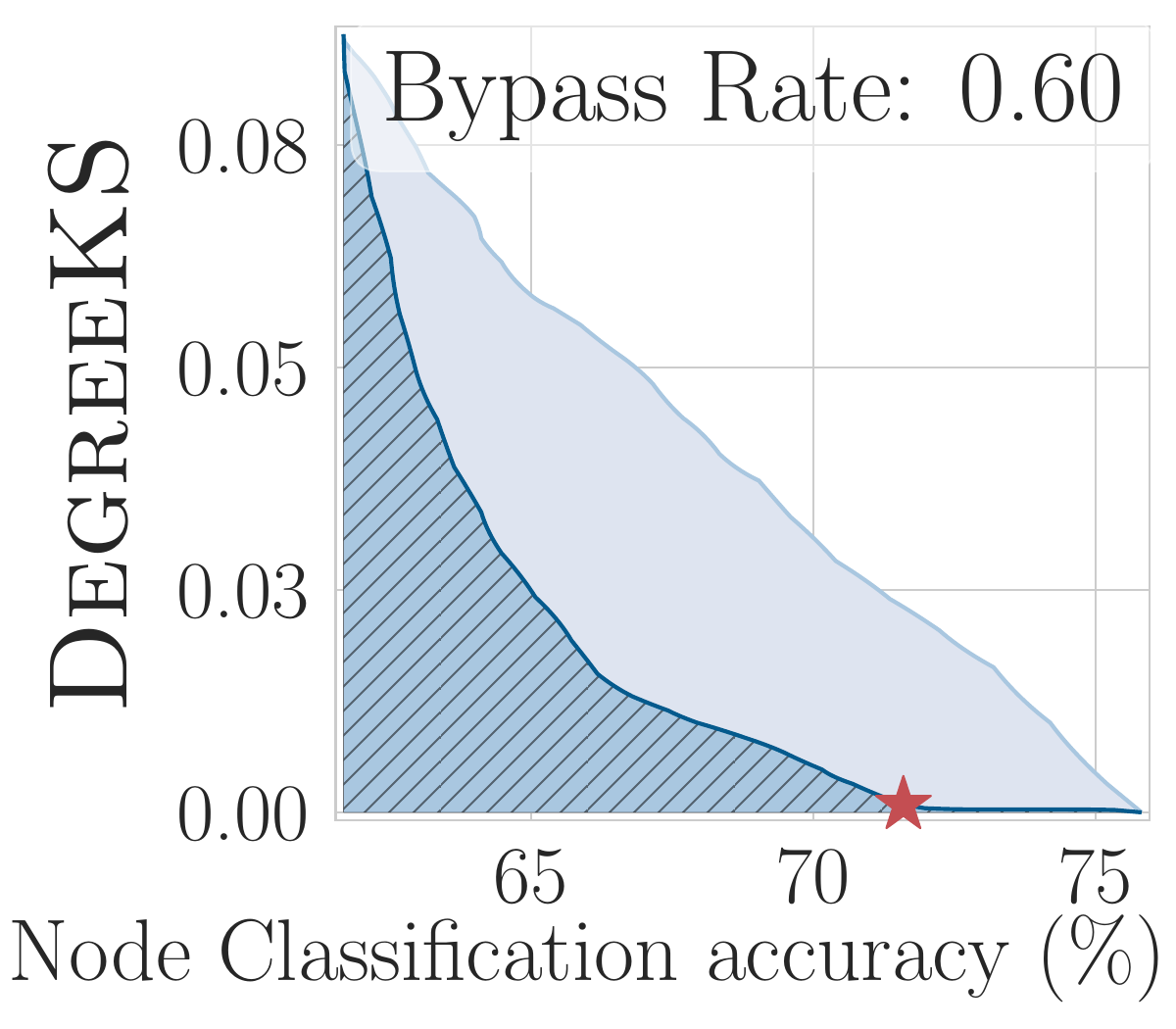} & 
        \includegraphics[width=\linewidth]{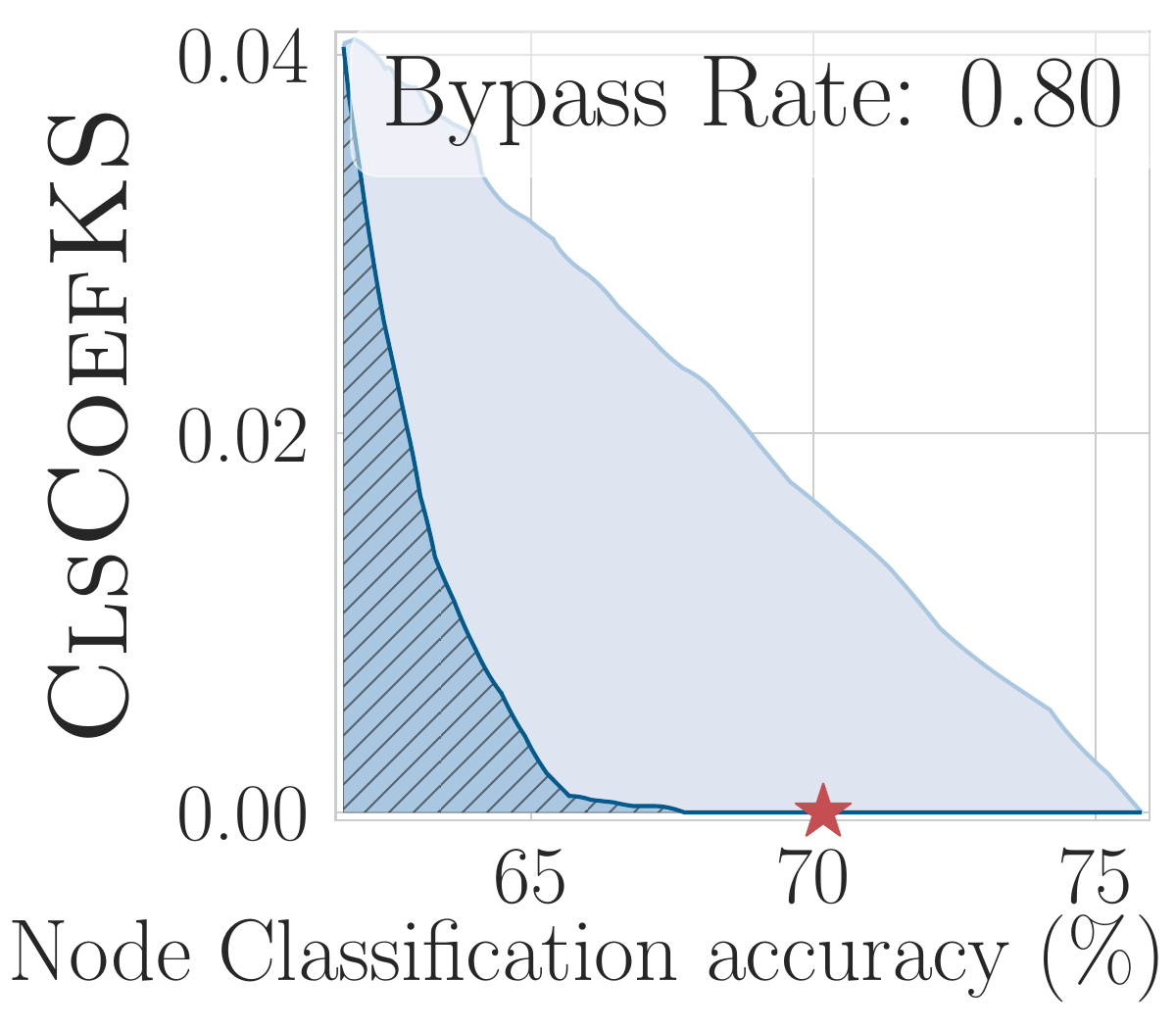} & 
        \includegraphics[width=\linewidth]{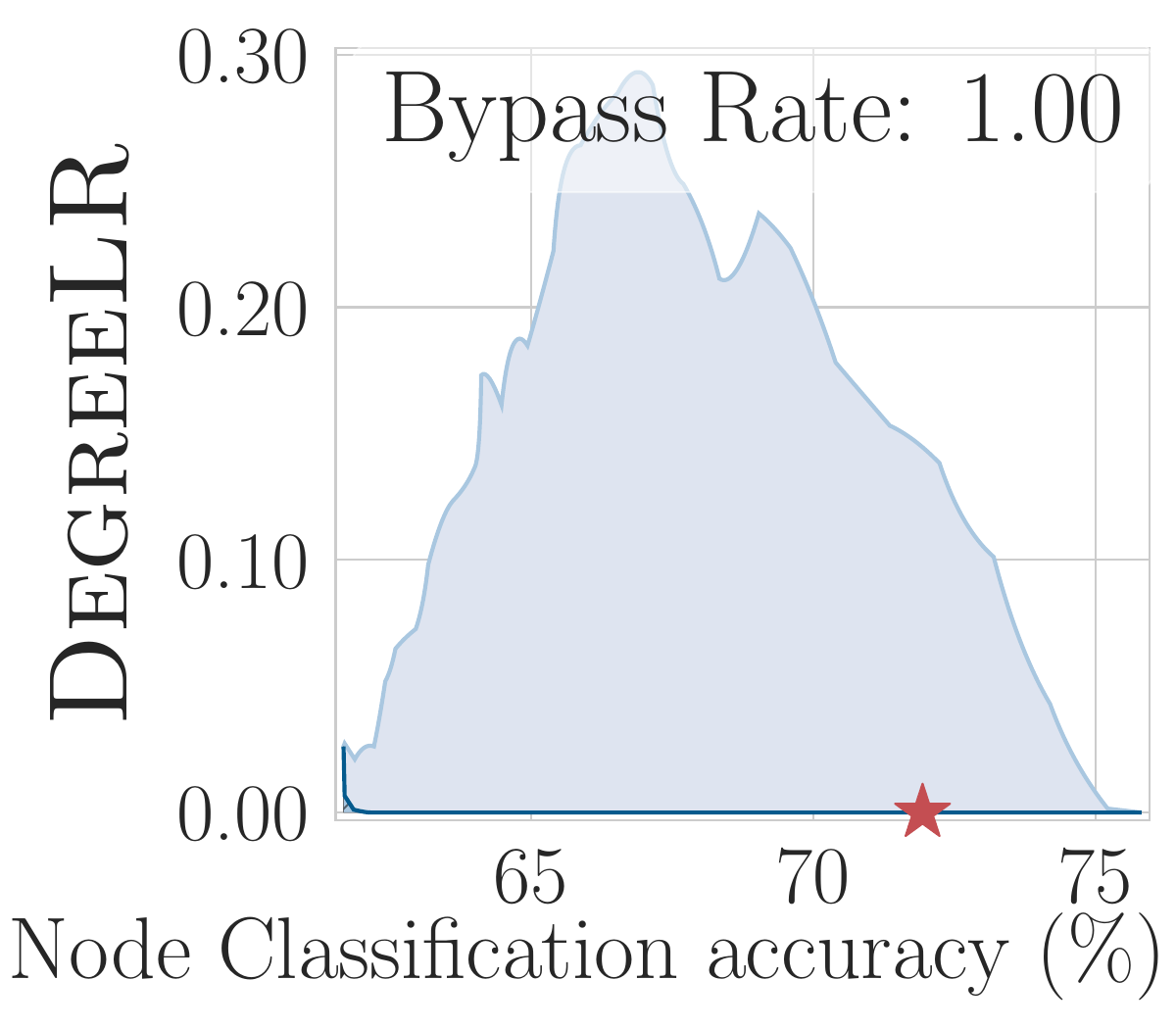} & 
        \includegraphics[width=\linewidth]{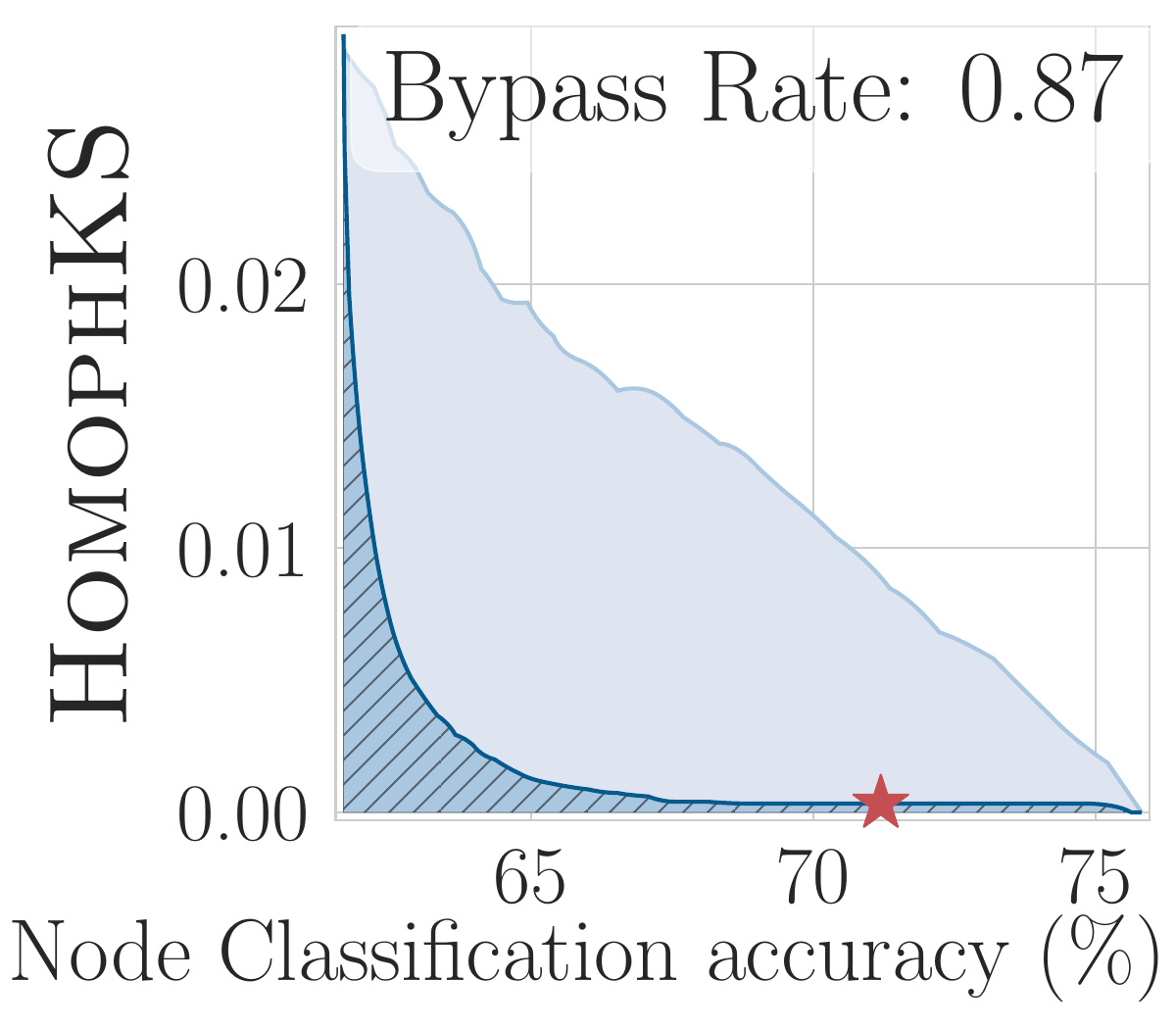} &
        \includegraphics[width=\linewidth]{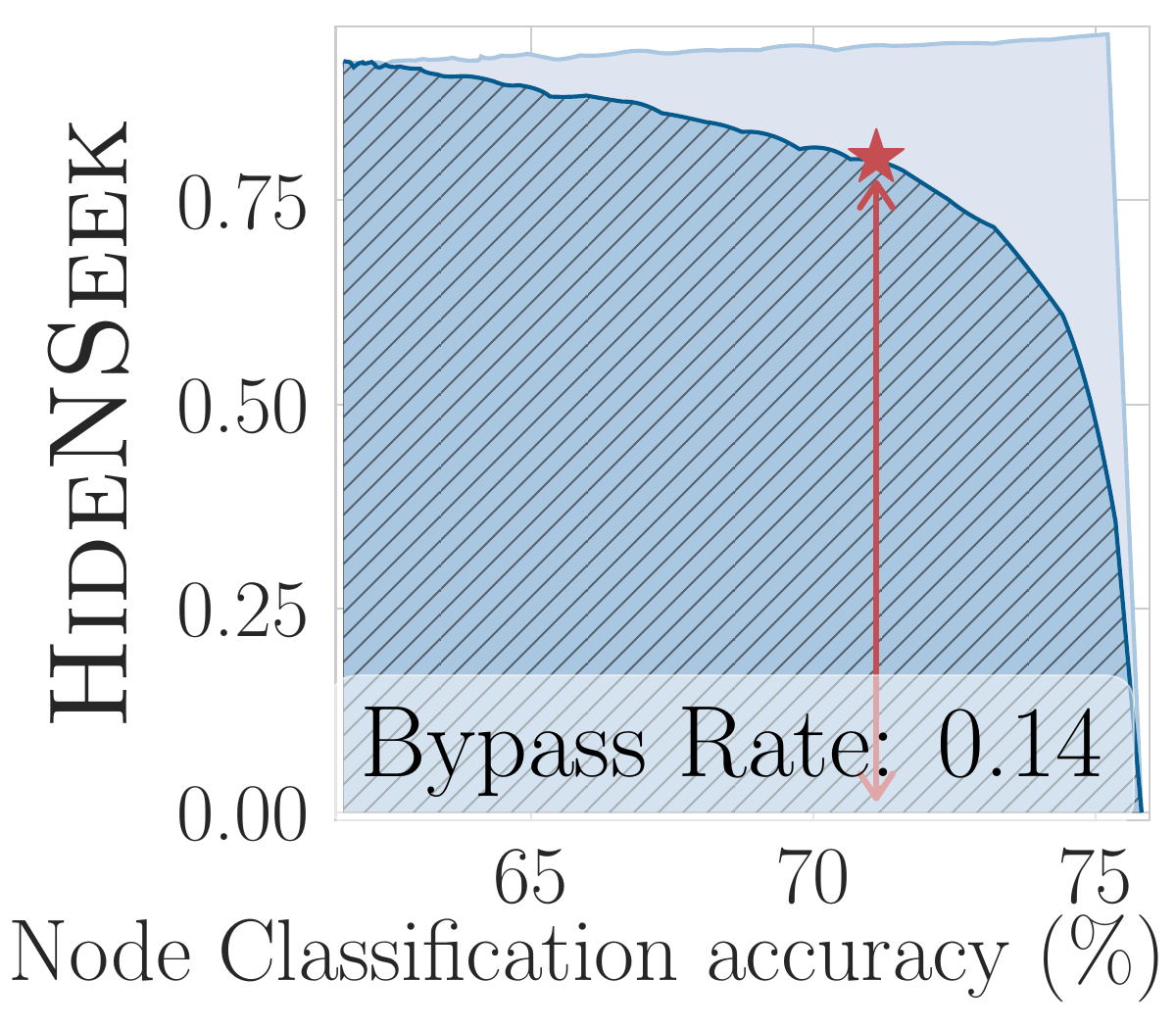} \\
        \hline
        \rotatebox[origin=c]{90}{\lastfmasia} & 
        \includegraphics[width=\linewidth]{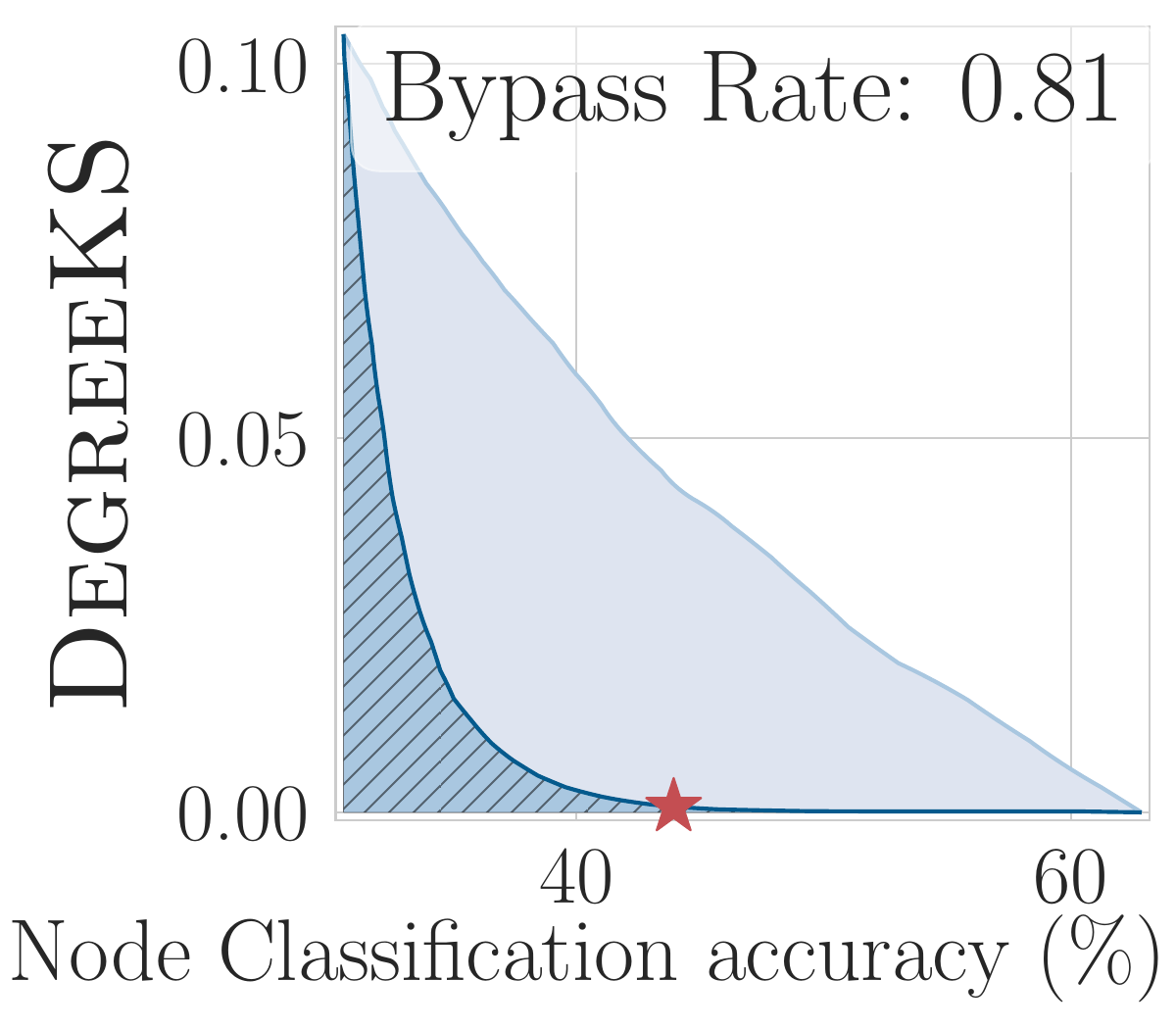} & 
        \includegraphics[width=\linewidth]{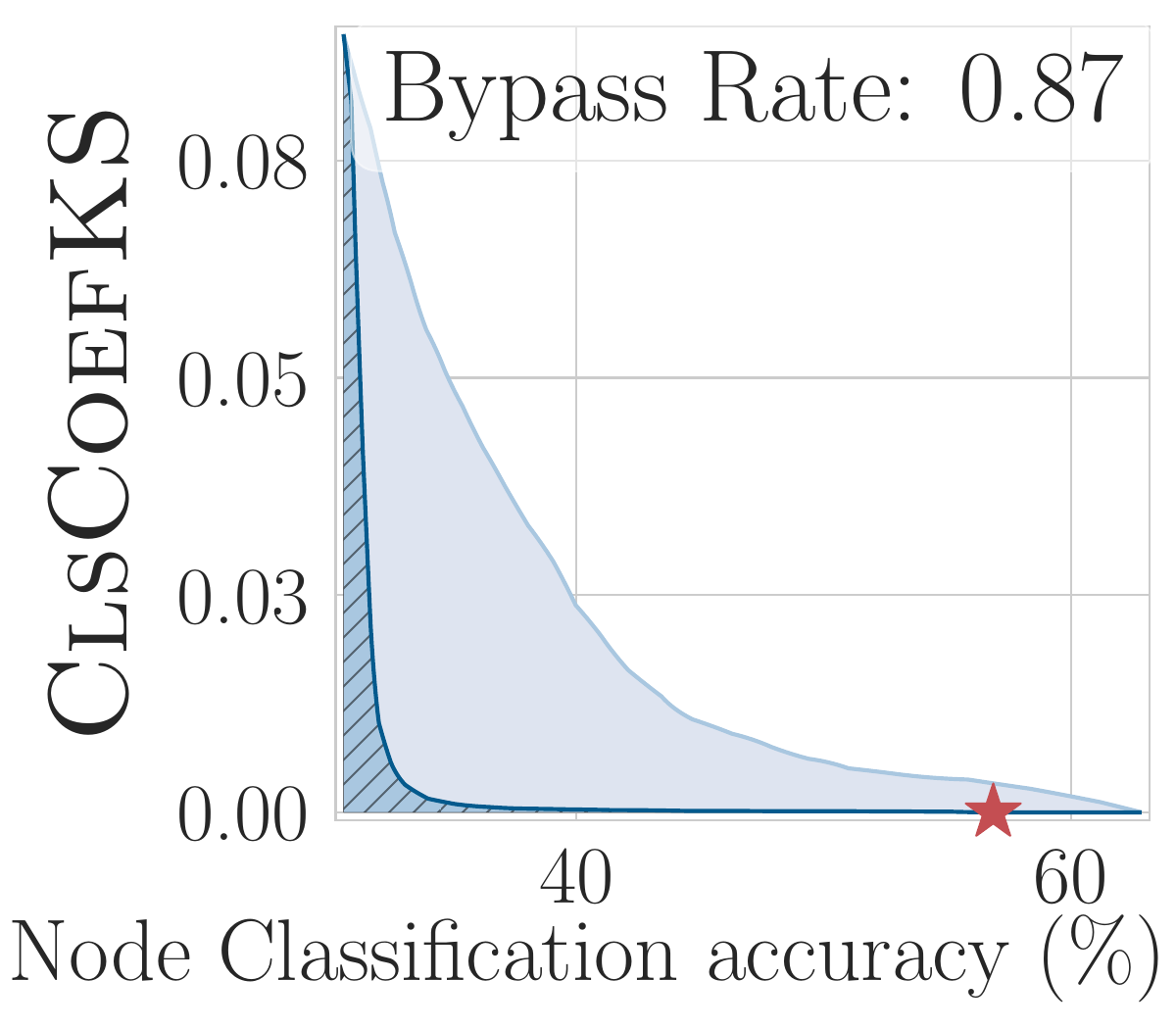} & 
        \includegraphics[width=\linewidth]{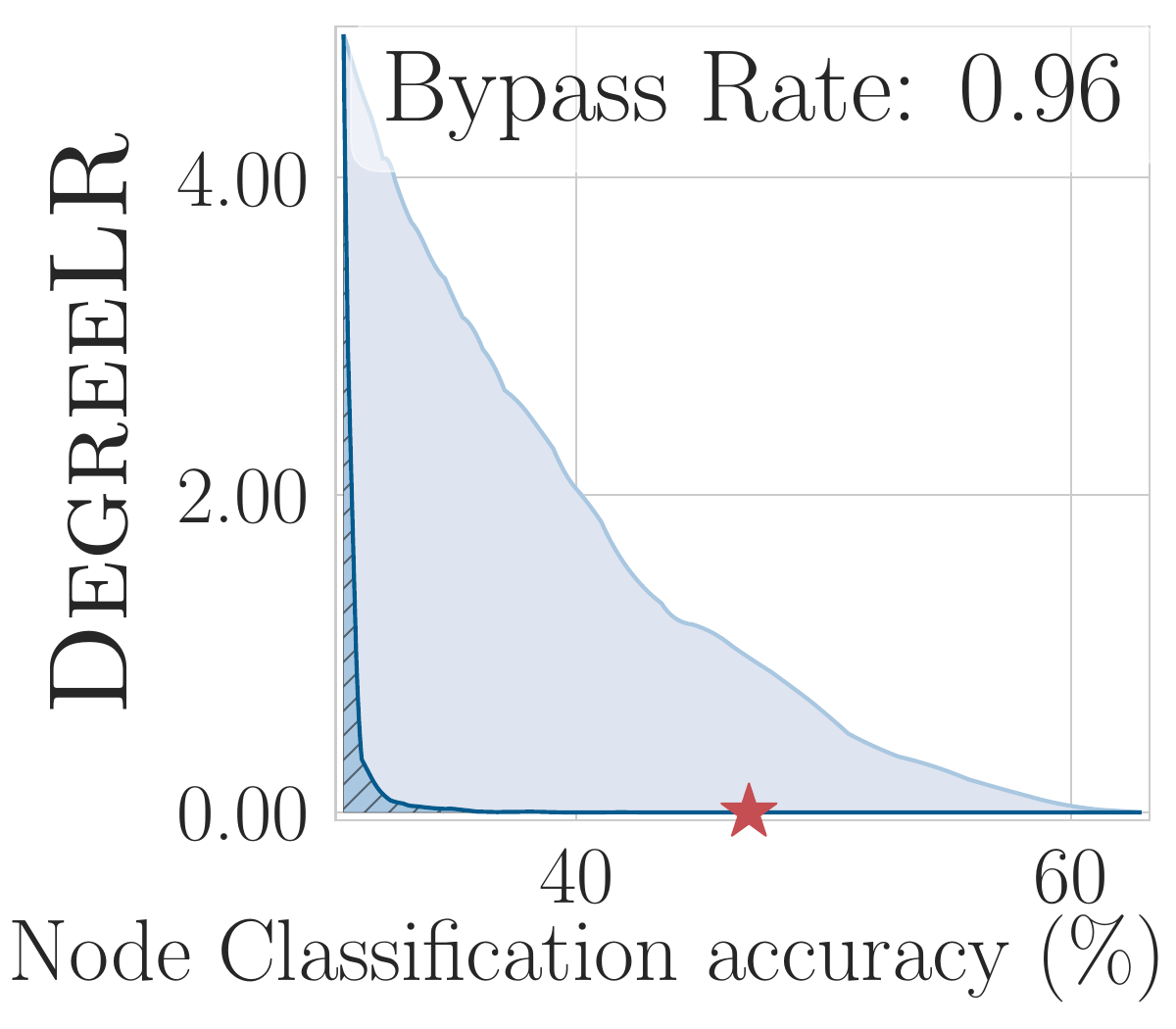} & 
        \includegraphics[width=\linewidth]{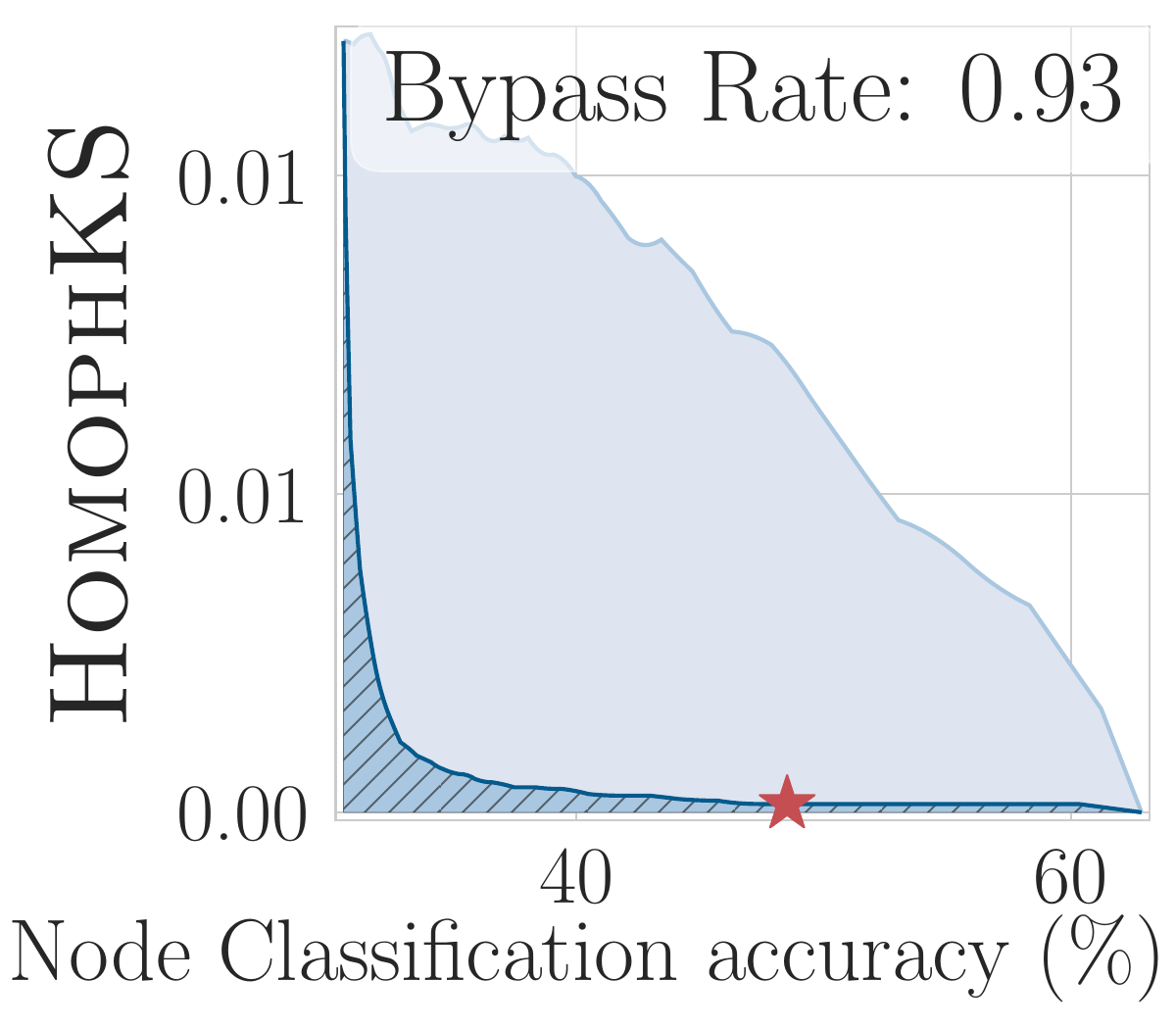} &
        \includegraphics[width=\linewidth]{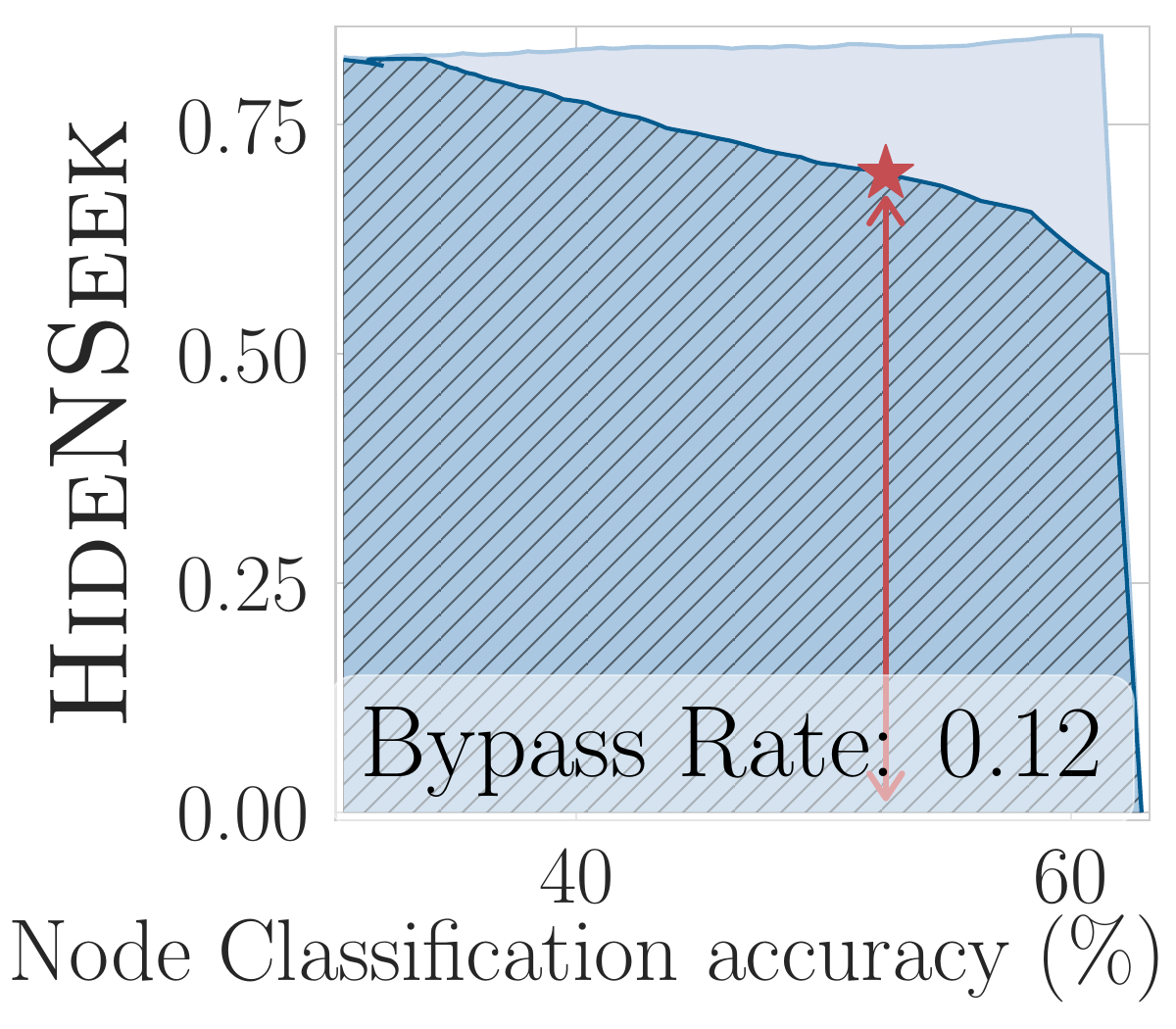} \\
        \hline
        \rotatebox[origin=c]{90}{\chameleon} & 
        \includegraphics[width=\linewidth]{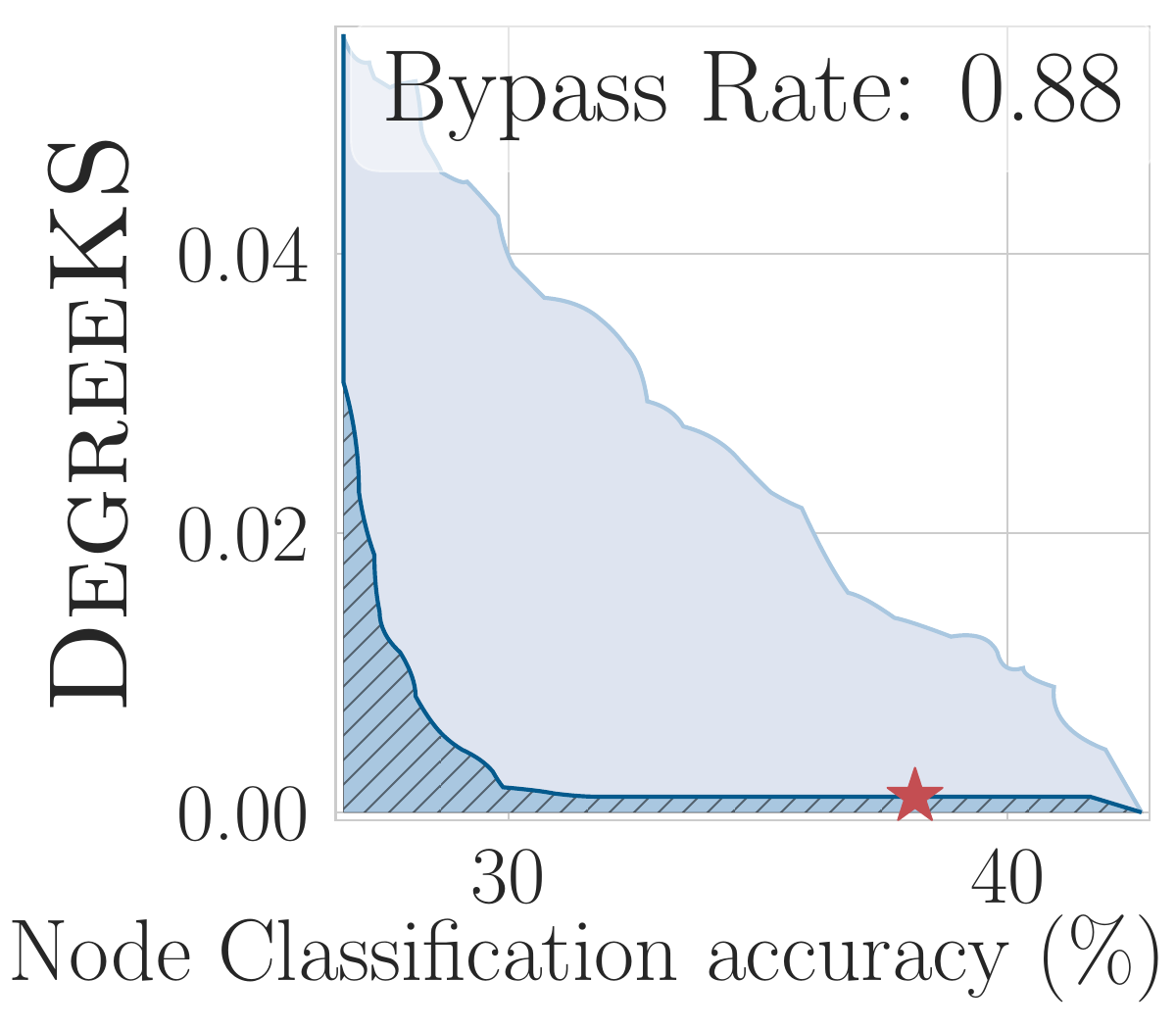} & 
        \includegraphics[width=\linewidth]{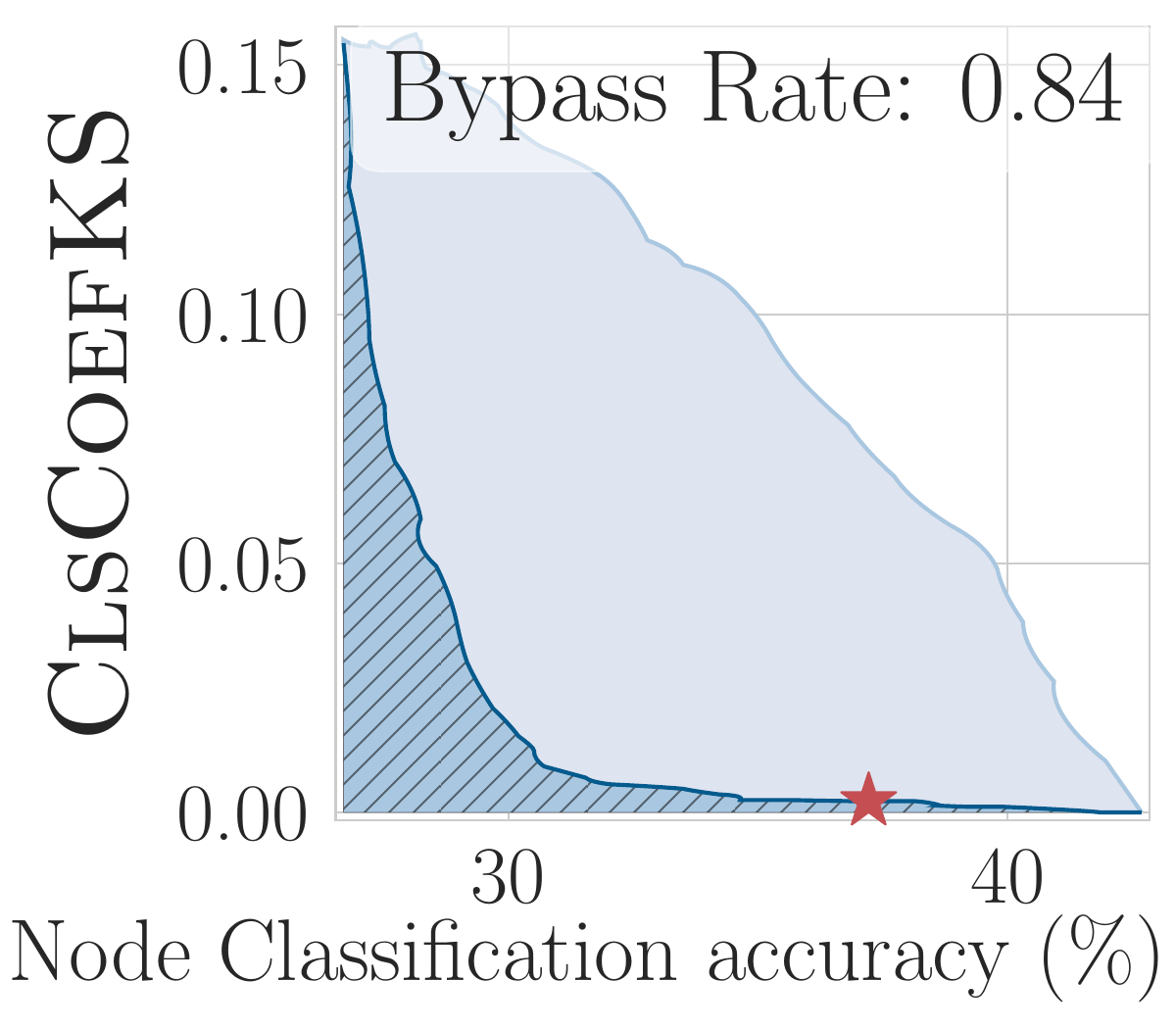} & 
        \includegraphics[width=\linewidth]{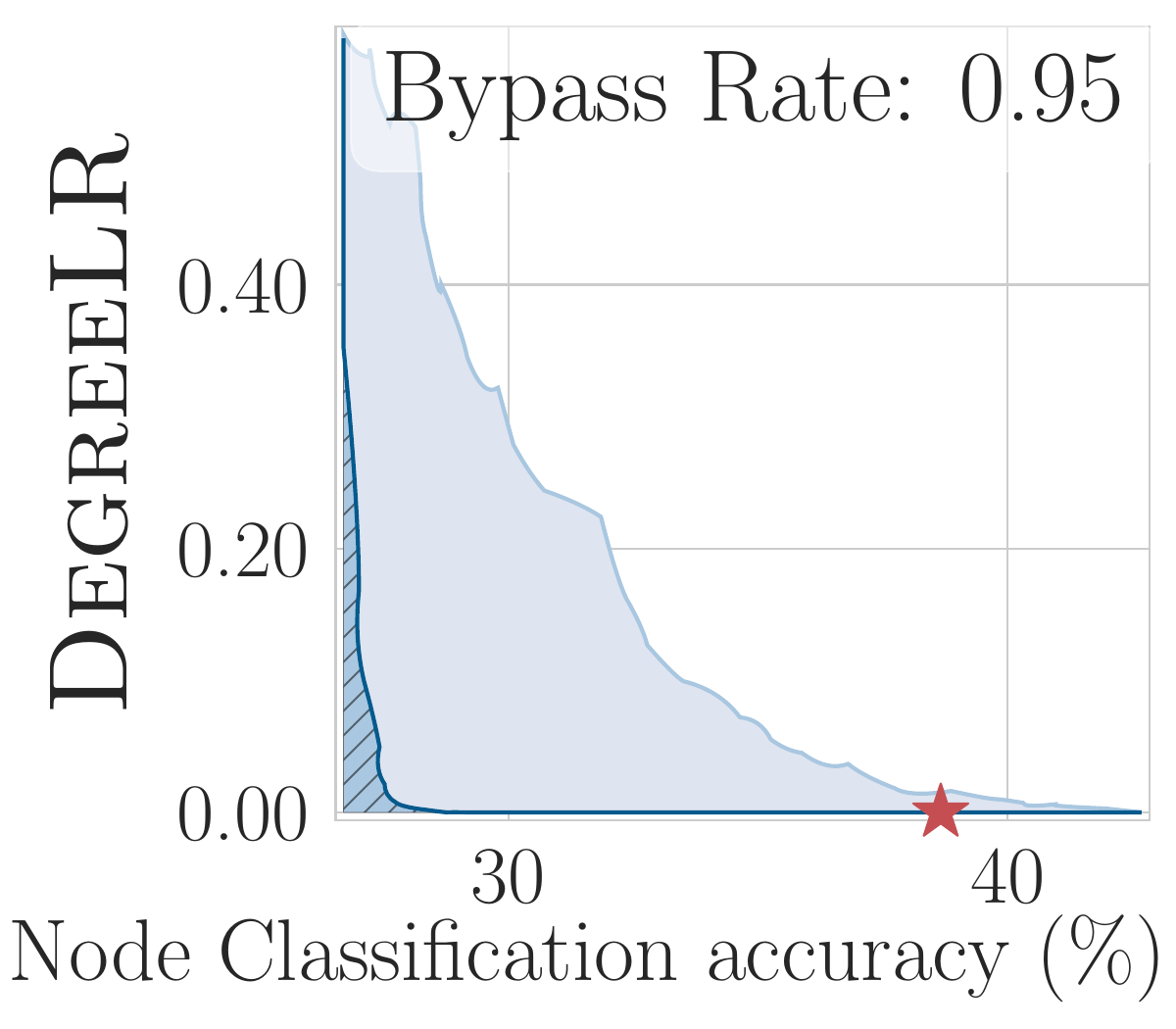} & 
        \includegraphics[width=\linewidth]{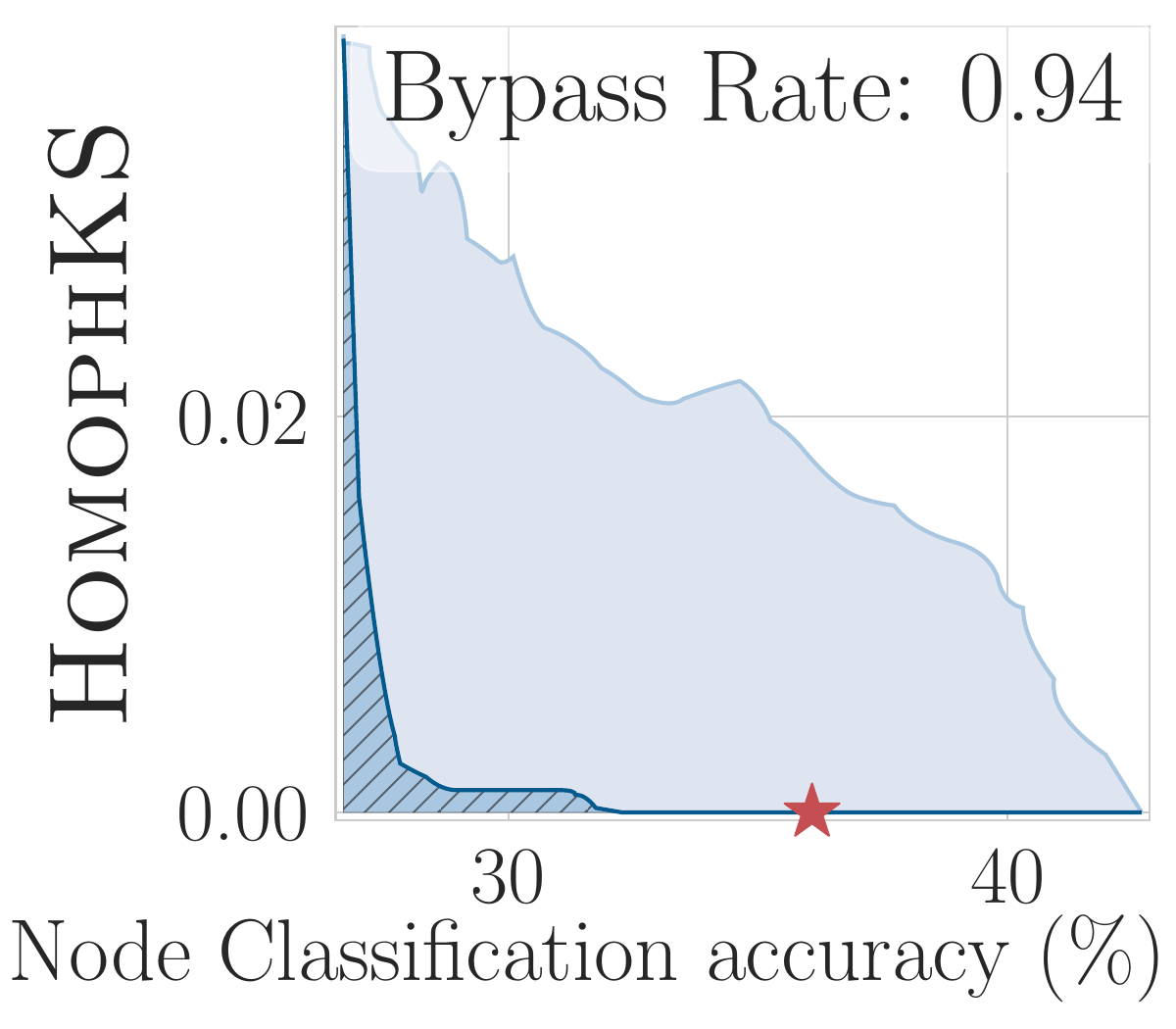} &
        \includegraphics[width=\linewidth]{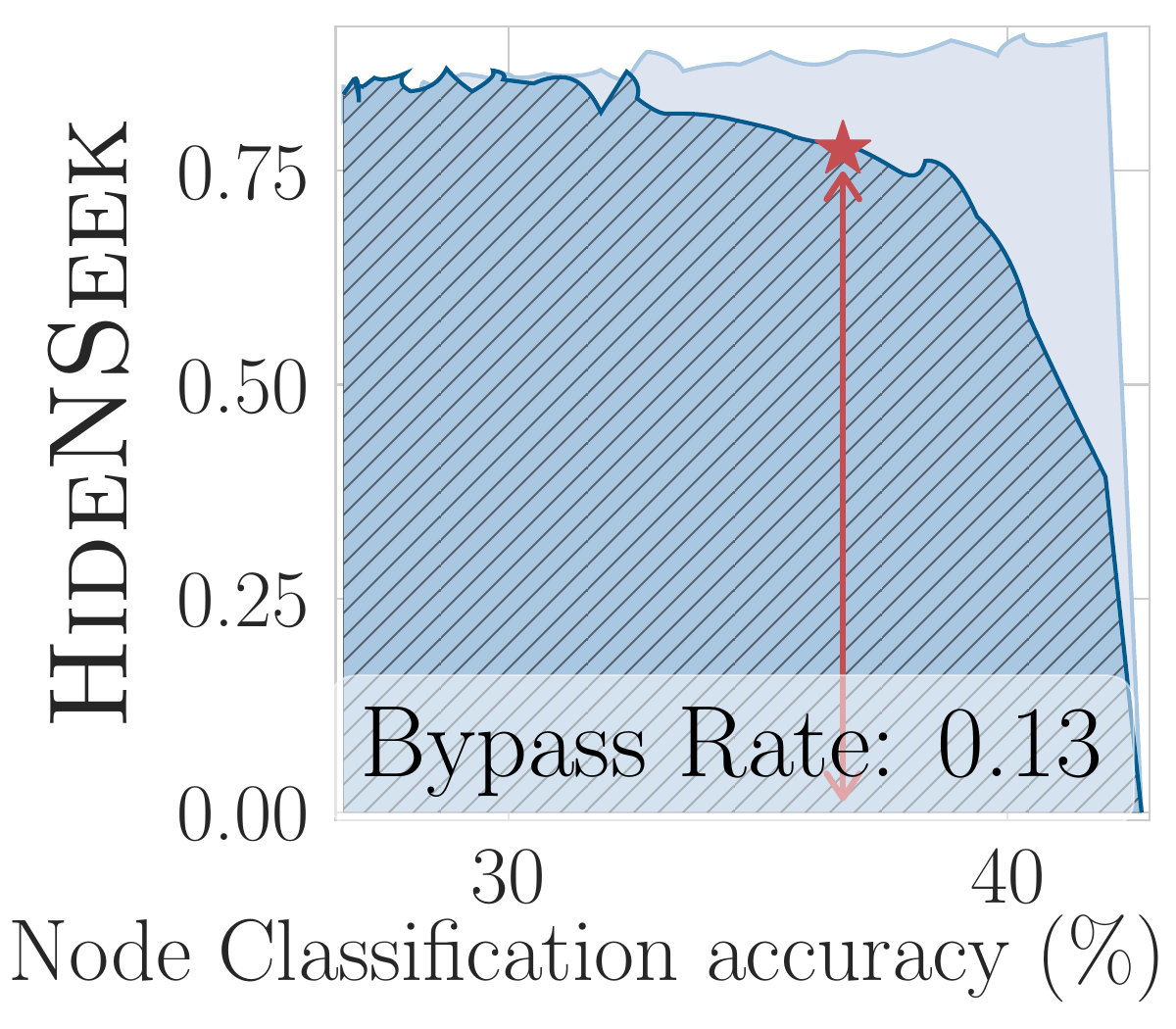} \\
        \hline
        \rotatebox[origin=c]{90}{\squirrel} & 
        \includegraphics[width=\linewidth]{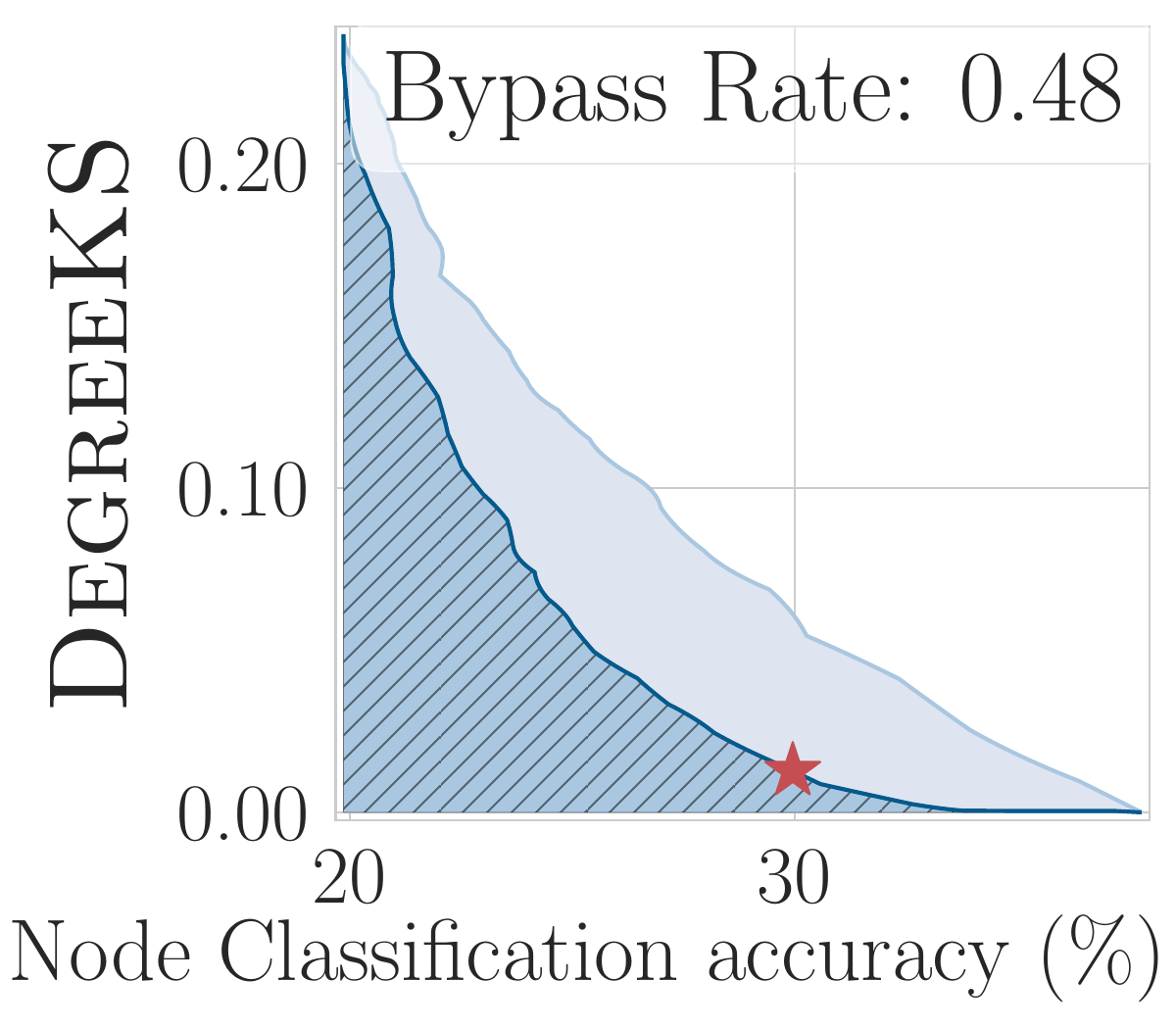} & 
        \includegraphics[width=\linewidth]{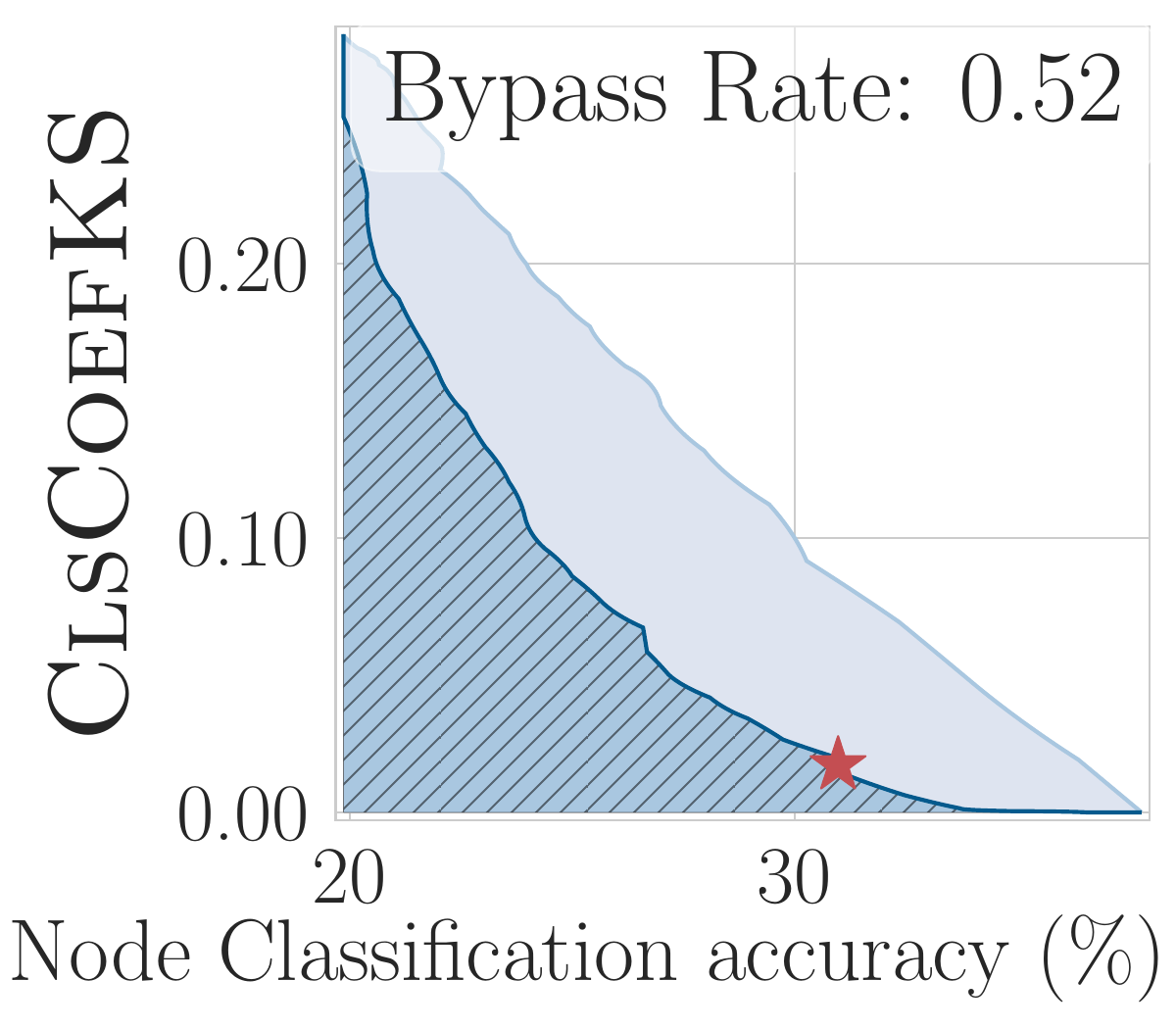} & 
        \includegraphics[width=\linewidth]{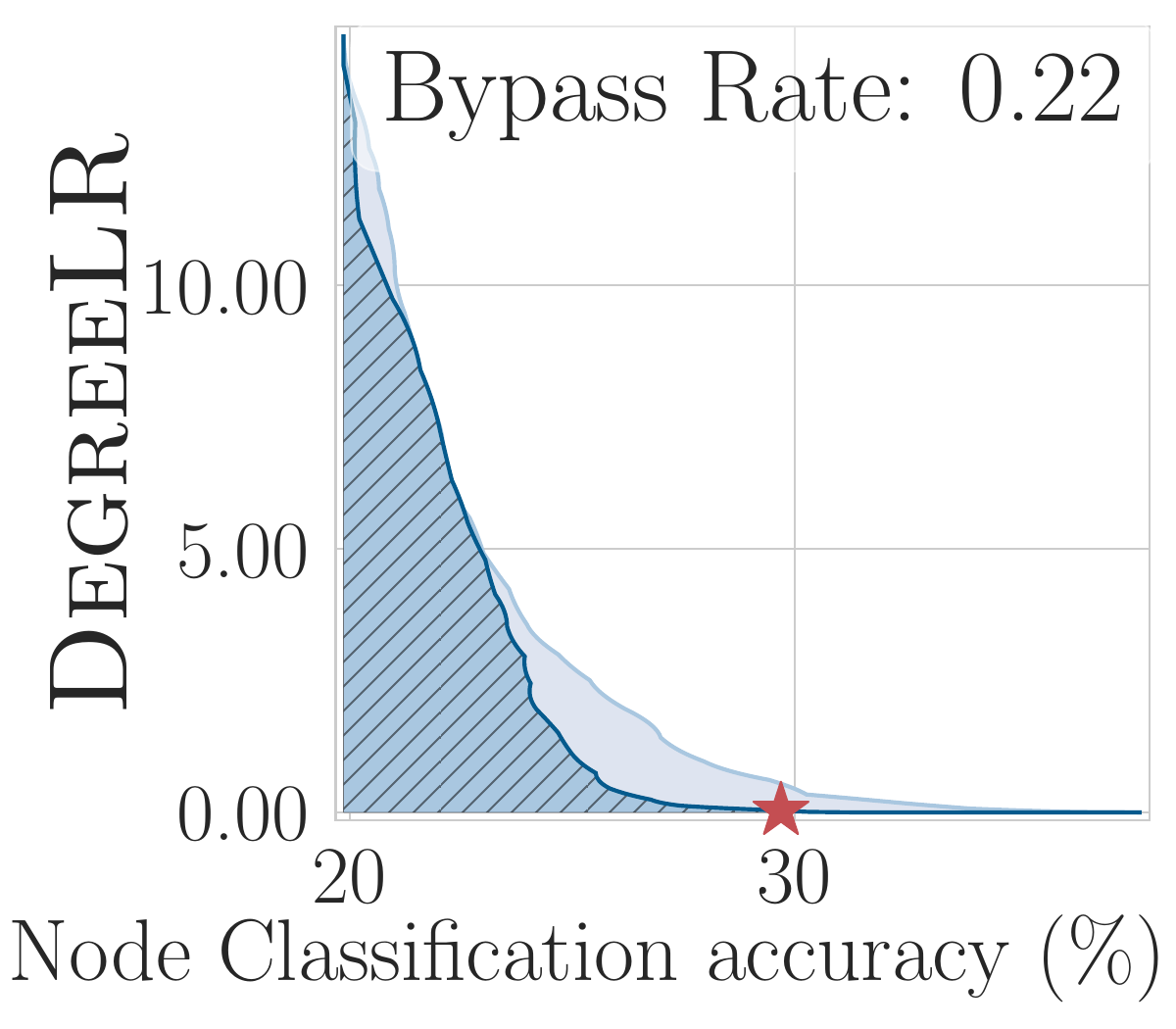} & 
        \includegraphics[width=\linewidth]{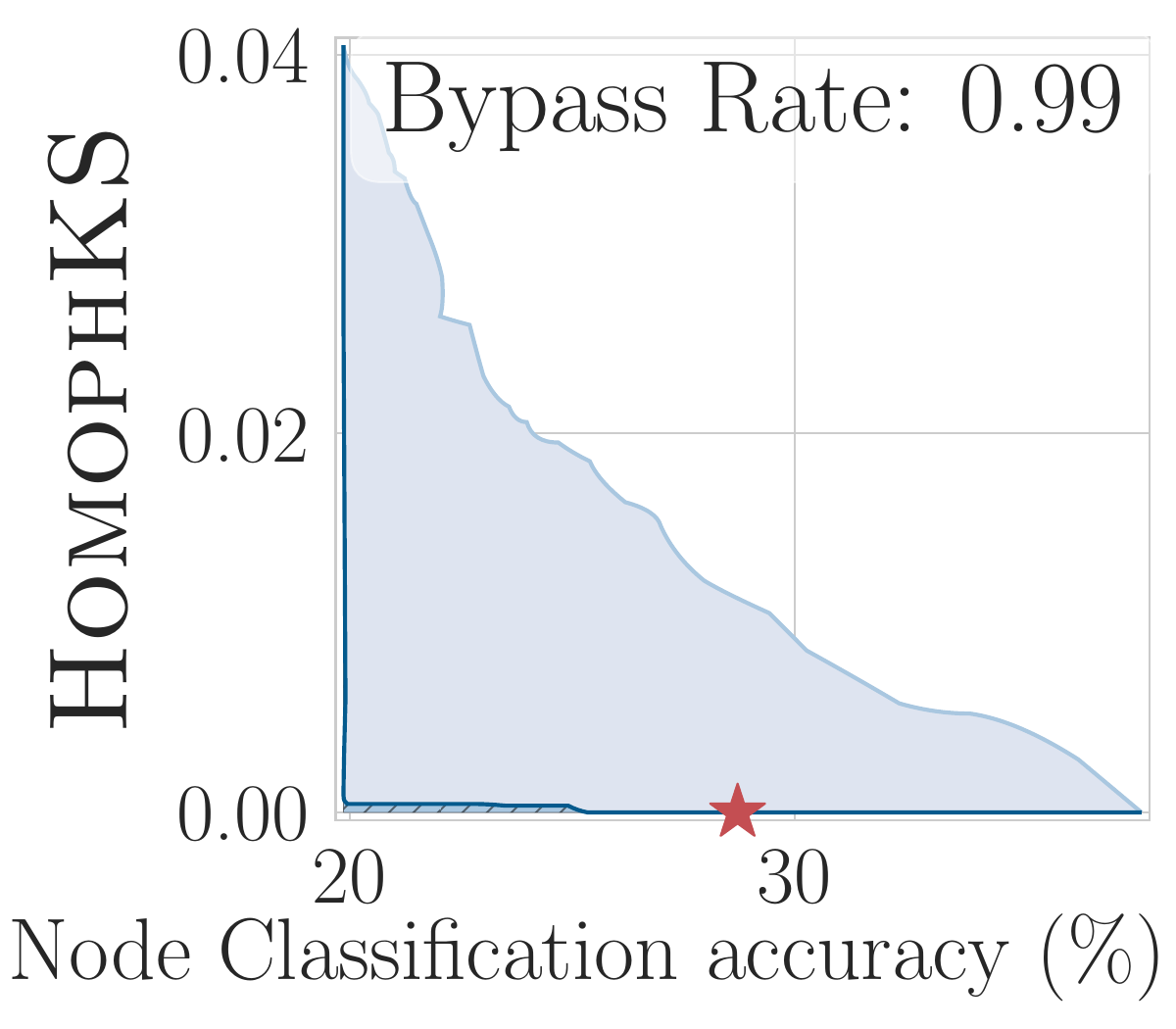} &
        \includegraphics[width=\linewidth]{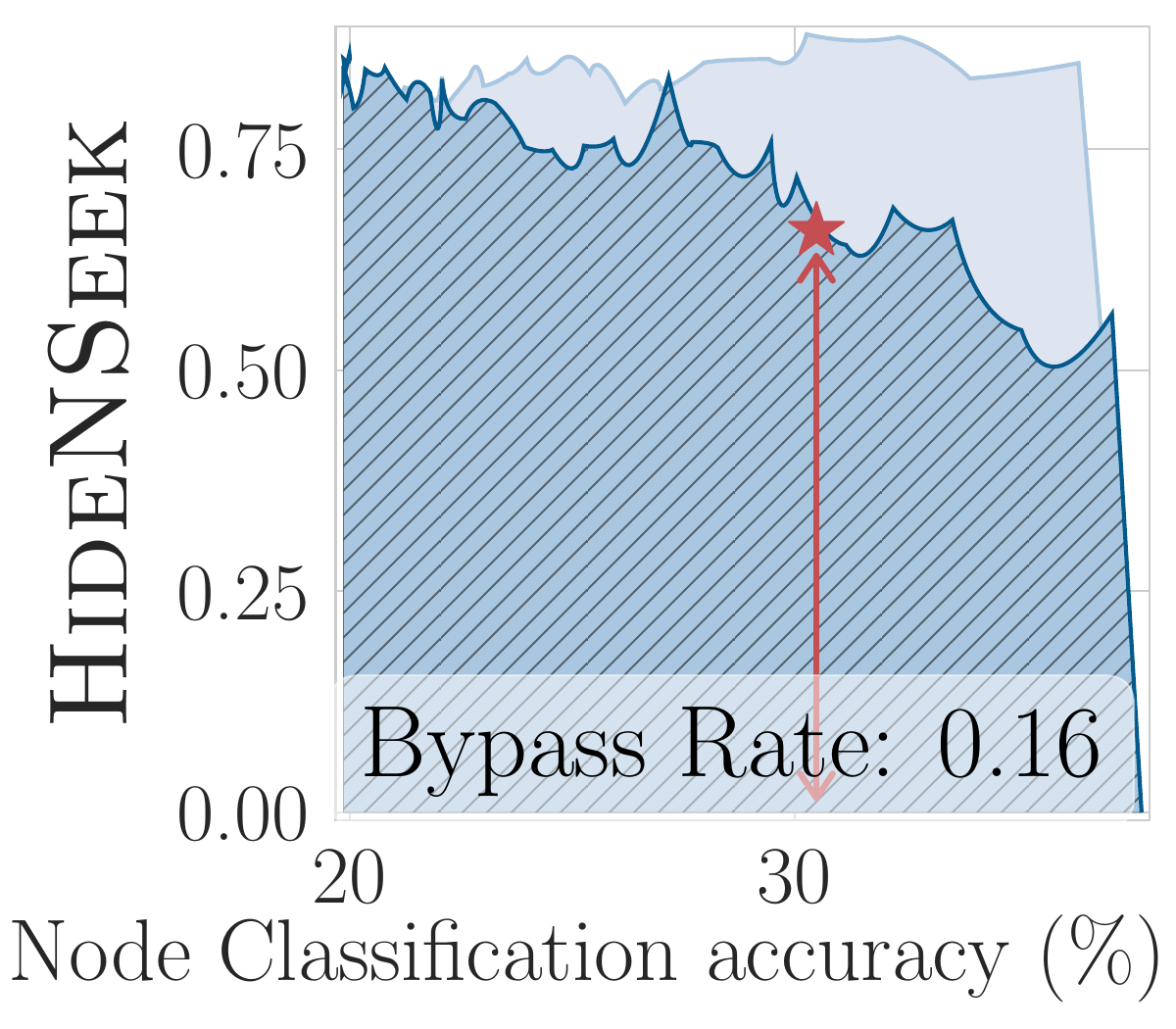} \\
        \hline
    \end{tabular}
\end{table*}
\endgroup

\subsection{Experiment Settings}\label{sec:exp:settings}

\smallsection{Hardware}
We run all experiments on a workstation with an Intel Xeon 4214 CPU, 512GB RAM, and RTX 8000 GPUs.

\smallsection{Datasets.}
We use six real-world datasets:
three citation networks (\cora, \coraml, and \citeseer \cite{sen2008collective}), 
one social network (\lastfmasia \cite{rozemberczki2020characteristic}), 
and two page-page networks from Wikipedia (\chameleon and \squirrel~\cite{rozemberczki2021multi}).
Basic statistics of the graphs are provided in Table~\ref{tab:dataset}. 
Detailed descriptions are as follows:
\begin{itemize}[leftmargin=*]
    \item \textbf{Cora, CoraML, and Citeseer} are citation networks in which nodes represent scientific publications, and edges indicate citation links between them.
    Each node feature is represented by a binary word vector, where each element indicates the absence or presence of a specific word from the dictionary.

    \item \textbf{LastFMAsia}  is a social network where nodes represent LastFM users from Asia, and edges denote their friendship relations.
    Node features are extracted based on the artists liked by the users.

    \item \textbf{Chameleon and Squirrel} are page-page networks on specific topics (chameleons and squirrels) representing relations between Wikipedia pages. 
    In this network, nodes represent web pages, and edges denote mutual links between them. 
    Each node feature is represented by a binary word vector indicating the absence/presence of an informative noun that appeared in the text of the Wikipedia article.
    In these datasets, we remove duplicated nodes, which lead to the train-test data leakage~\cite{platonov2022critical}.
\end{itemize}


\smallsection{Graph adversarial attack methods.}
To generate attacks, we employ five graph adversarial attack methods as follows: 
\begin{itemize}[leftmargin=*]
    \item \textbf{\random} inserts attack edges uniformly at random.
    \item \textbf{\dice}~\cite{waniek2018hiding} inserts attack edges chosen uniformly at random among those whose endpoints belong to different classes, and removes existing edges whose endpoint belongs to the same classes. 
    We set the number of inserted attack edges and the number of deleted existing edges to be the same.
    \item \textbf{\pgd}~\cite{xu2019topology} is a gradient-based attack method that utilizes the projected gradient descent. We use the hyperparameters provided in the source code released by the authors.
    \item \textbf{\structack}~\cite{hussain2021structack} selects nodes with low centrality and links pairs of nodes with low similarity.
    We use betweenness centrality for node centrality and the Katz index for node similarity. 
    \item \textbf{\metattack}~\cite{zugner2019adversarial}  addresses the graph adversarial attack problem as a min-max problem in poisoning attacks using meta-gradients through a surrogate model. 
    We use the hyperparameters in the source code released by the authors.
\end{itemize}
We set the attack rate $\gamma$ to $10\%$, \kijung{unless otherwise stated.}
All results are averaged over five trials.

\subsection{Q1. Justification of \model}\label{sec:exp:q1}
Recall that \measure uses the edge scores output by \model to compute attack noticeability. 
We shall show that \model is effective in detecting attack edges, supporting the reliability of \measure.

\smallsection{Competitors.}
We compare \model with various competitors that assign scores to edges:
(a) two \textit{node proximity-based} methods 
(\svd, \cosine), 
(b) three logistic regression-based methods that utilizes the \textit{properties used in existing noticeability measure}
(\degreelg, \lclg, and \homophilylg), and
(c) six \textit{GCN-based} methods 
(\gcn~\cite{kipf2016semi}, \gcnsvd~\cite{entezari2020all}, \rgcn~\cite{zhu2019robust}, \mediangcn~\cite{chen2021understanding2}, \gnnguard~\cite{zhang2020gnnguard}, and \metagc~\cite{jo2023robust}). 
See Online Appendix \textsc{E.2}~\cite{code} for more details.

\smallsection{Evaluation.}
For all the models, AUROC is used for performance evaluation, where a higher AUROC score indicates better classification of attack edges from real edges.
For each setting, we report the mean and standard deviation, computed over five trials.
For each attack method and each model, we also report the average rank (AR) over all datasets.
See Appendix~\ref{appendix:auroc} for more details.

\smallsection{Results.} 
In Table~\ref{tab:exp:q1}, we present the noticeability scores obtained from all twelve edge-scoring methods (\model and eleven baselines).
First of all, \model performs the best overall in distinguishing attack edges, achieving the best average rank (1.0 to 2.8) for each attack method, among all twelve methods.
Additional experiment results with different attack rates can be found in Online Appendix \textsc{F.2}~\cite{code}.

\begin{figure*}[t!]
\includegraphics[width=0.75\linewidth]{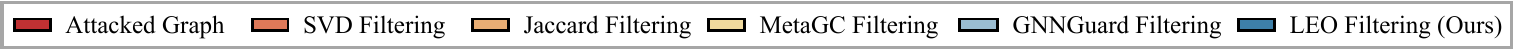} \\
\vspace{-4mm}
\textbf{\gcn}: \hfill \; \\
    \begin{subfigure}{0.195\linewidth}
        \centering
        \includegraphics[width=\textwidth]{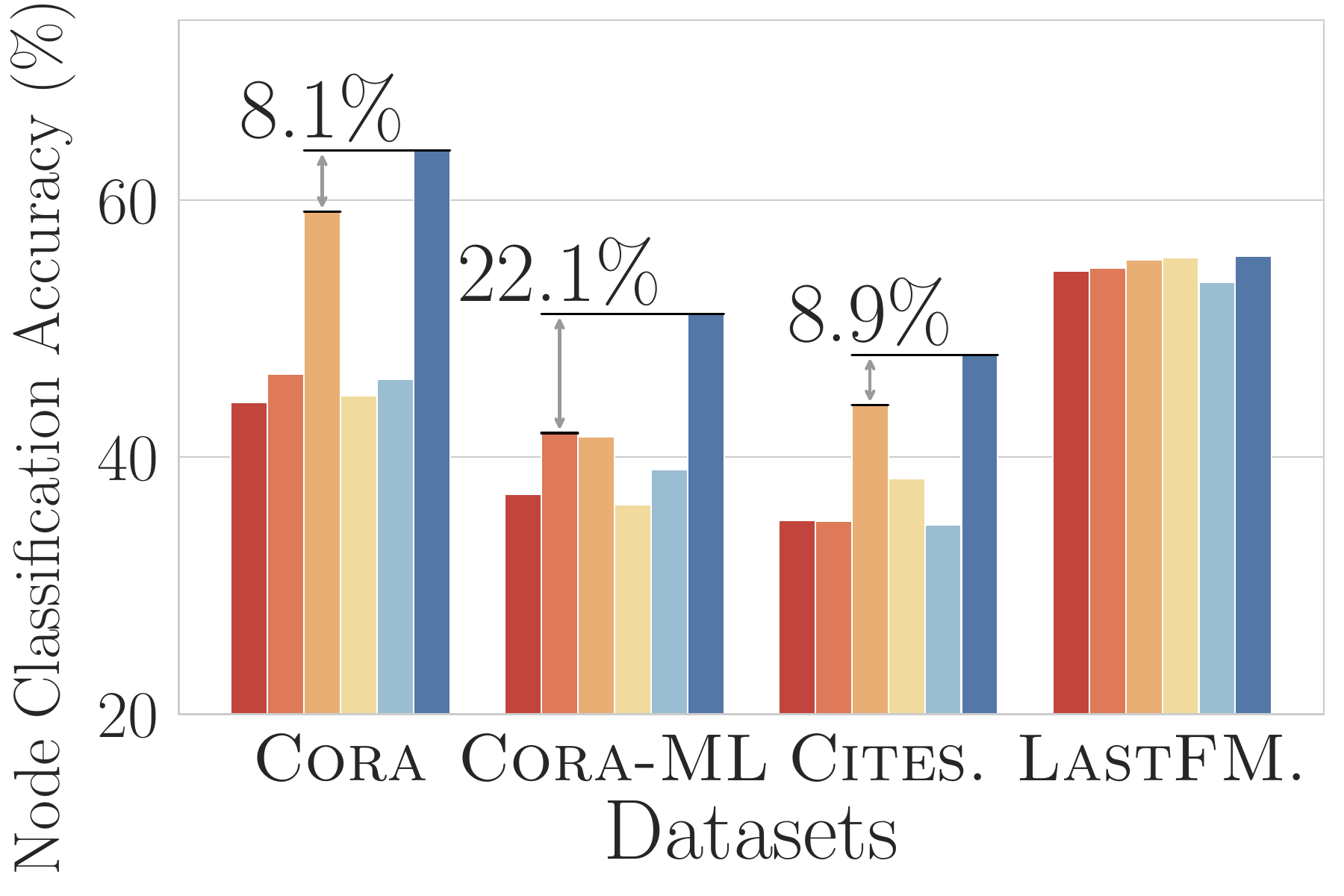}
        \Description[<short description>]{<long description>}
        \caption{\metattack}
        \label{fig:exp:q3_gcn_metattack}
    \end{subfigure}
    \hfill
    \begin{subfigure}{0.195\linewidth}
        \centering
        \includegraphics[width=\linewidth]{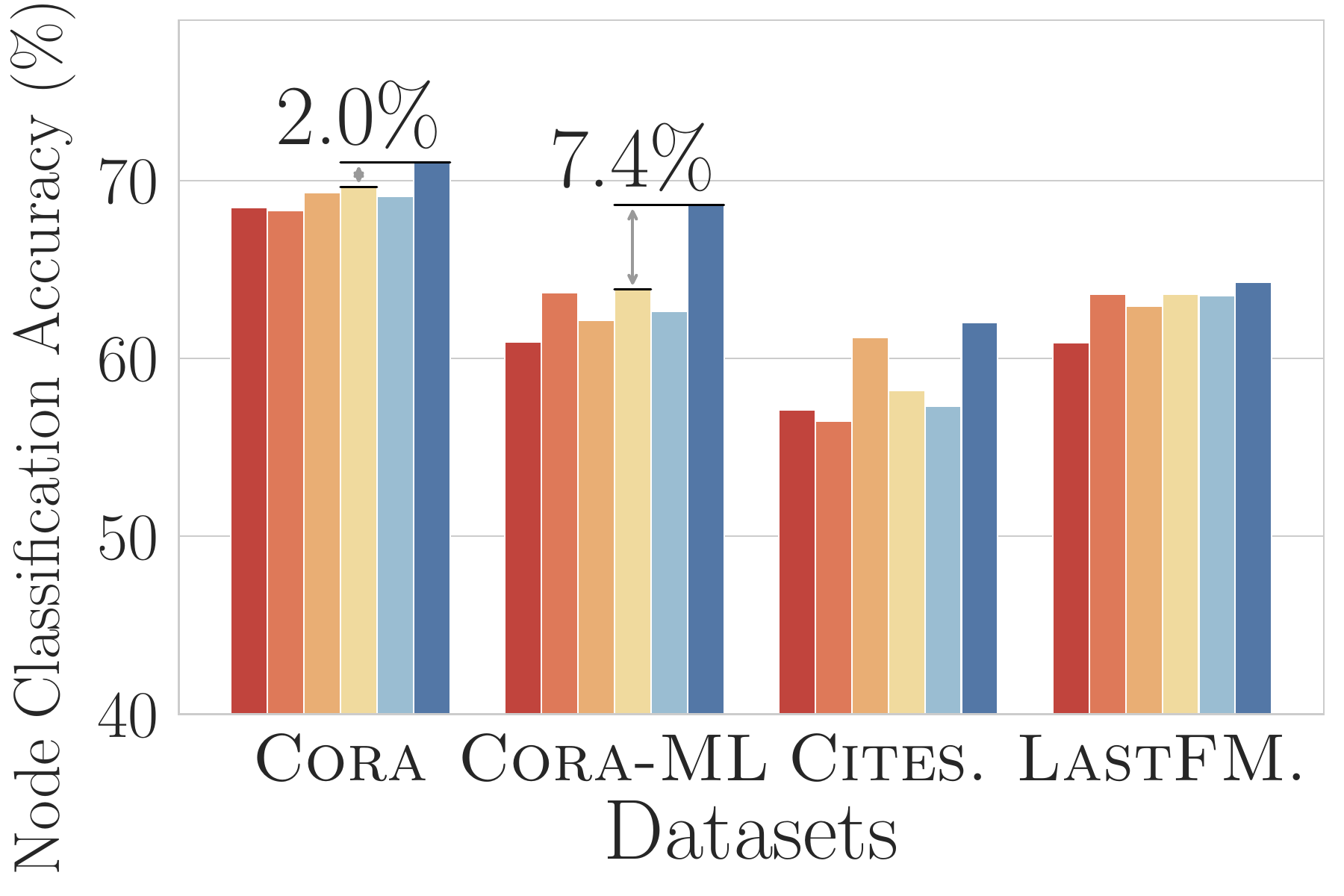}
        \Description[<short description>]{<long description>}
        \caption{\pgd}
        \label{fig:exp:q3_gcn_pgd}
    \end{subfigure} 
    \hfill
    \begin{subfigure}{0.195\linewidth}
        \centering
        \includegraphics[width=\linewidth]{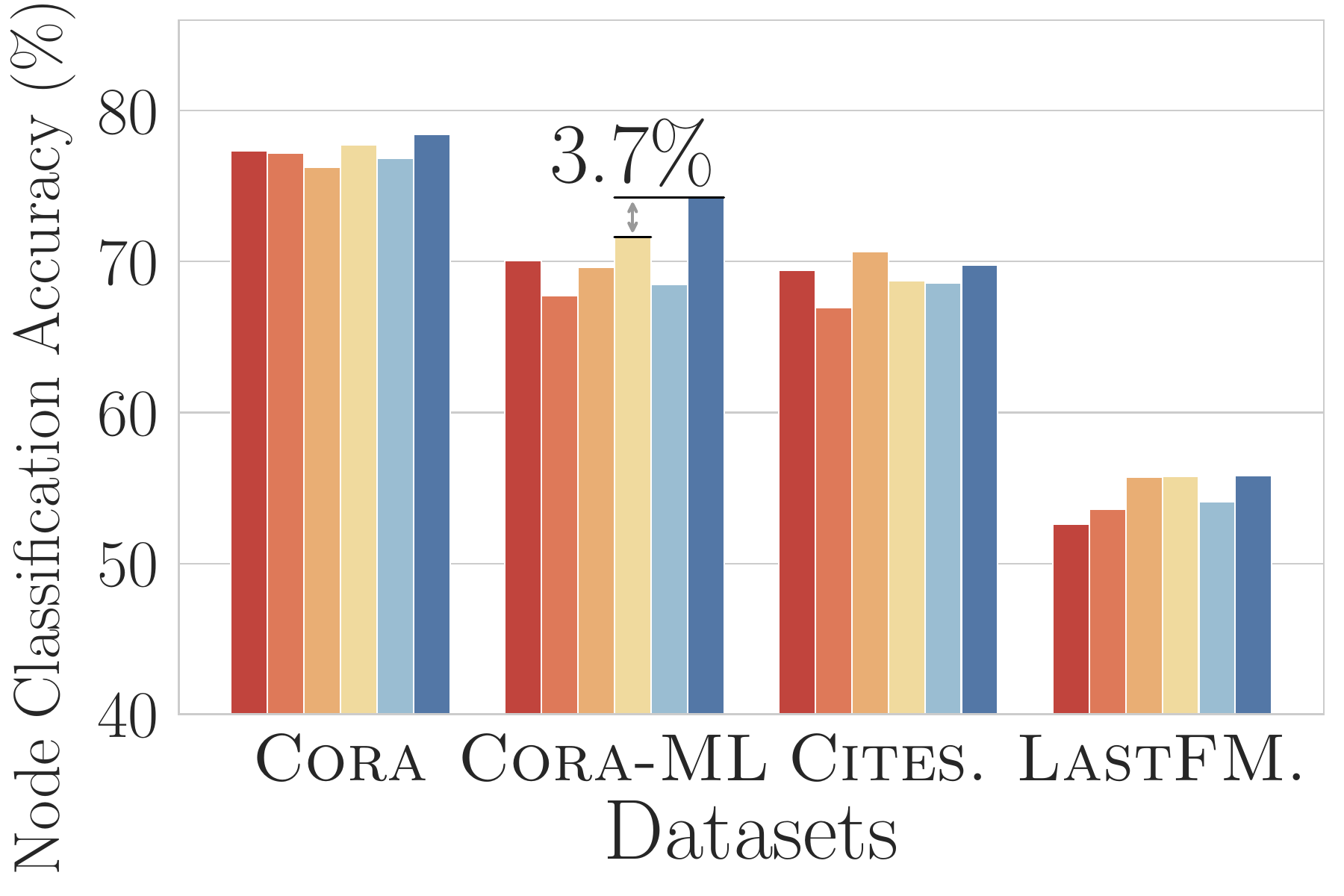}
        \Description[<short description>]{<long description>}
        \caption{\structack}
        \label{fig:exp:q3_gcn_structack}
    \end{subfigure} 
    \hfill
    \begin{subfigure}{0.195\linewidth}
        \centering
        \includegraphics[width=\linewidth]{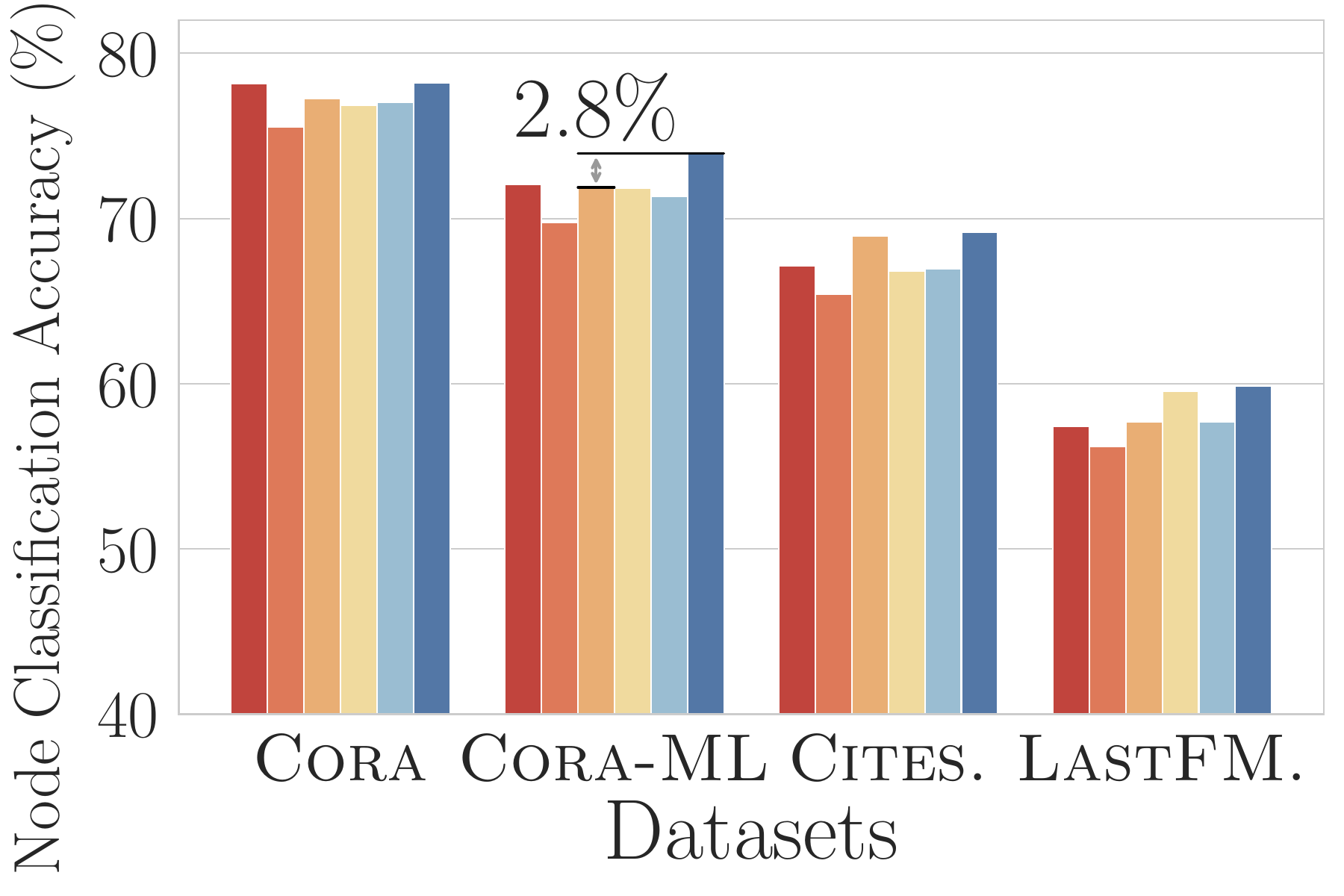}
        \Description[<short description>]{<long description>}
        \caption{\dice}
        \label{fig:exp:q3_gcn_dice}
    \end{subfigure} 
    \hfill
    \begin{subfigure}{0.195\linewidth}
        \centering
        \includegraphics[width=\linewidth]{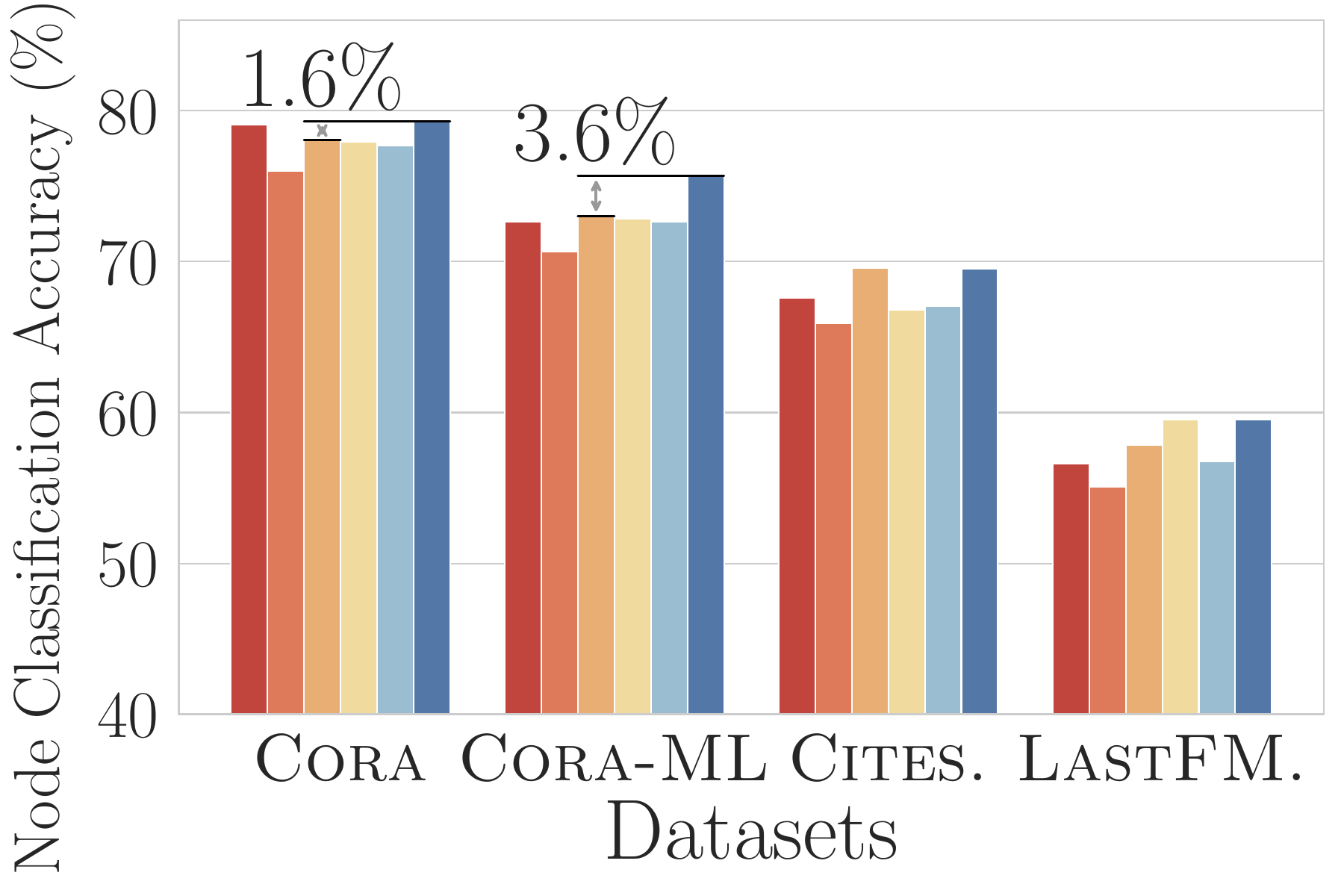}
        \Description[<short description>]{<long description>}
        \caption{\random}
        \label{fig:exp:q3_gcn_random}
    \end{subfigure} \\[-5pt]
    \textbf{\rgcn}: \hfill \; \\
    \begin{subfigure}{0.195\linewidth}
        \centering
        \includegraphics[width=\textwidth]{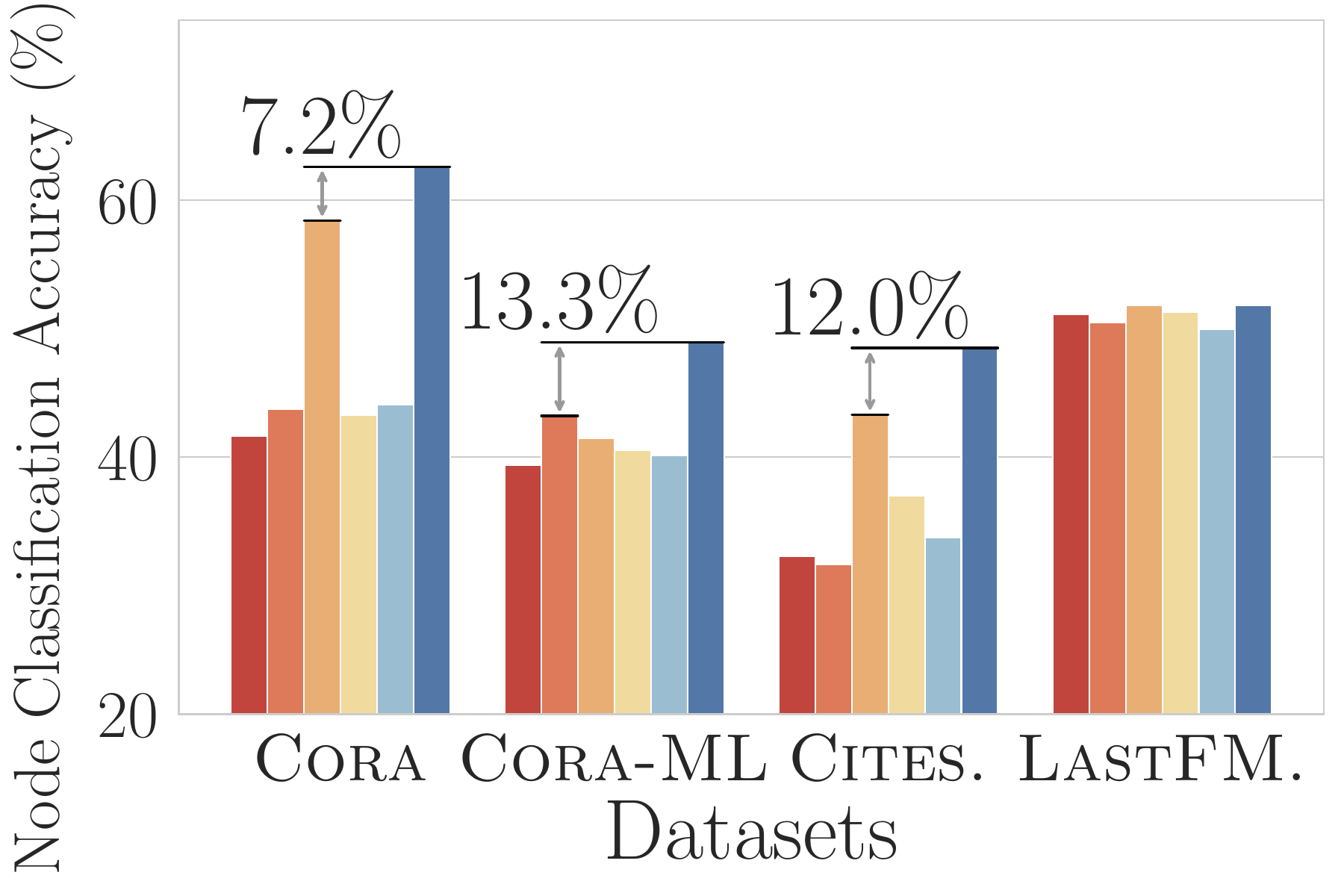}
        \Description[<short description>]{<long description>}
        \caption{\metattack}
        \label{fig:exp:q3_rgcn_metattack}
    \end{subfigure}
    \hfill
    \begin{subfigure}{0.195\linewidth}
        \centering
        \includegraphics[width=\linewidth]{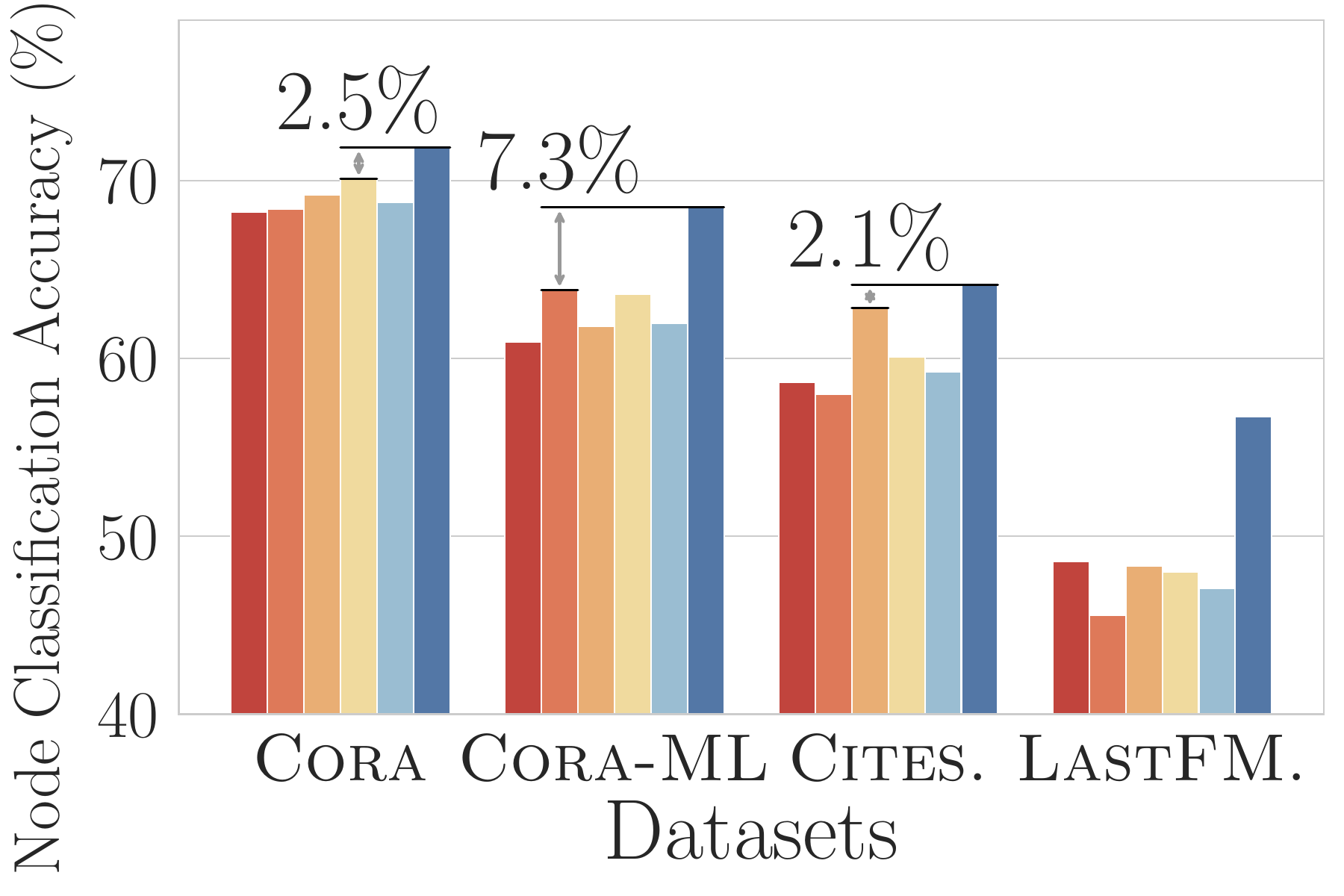}
        \caption{\pgd}
        \label{fig:exp:q3_rgcn_pgd}
    \end{subfigure} 
    \hfill
    \begin{subfigure}{0.195\linewidth}
        \centering
        \includegraphics[width=\linewidth]{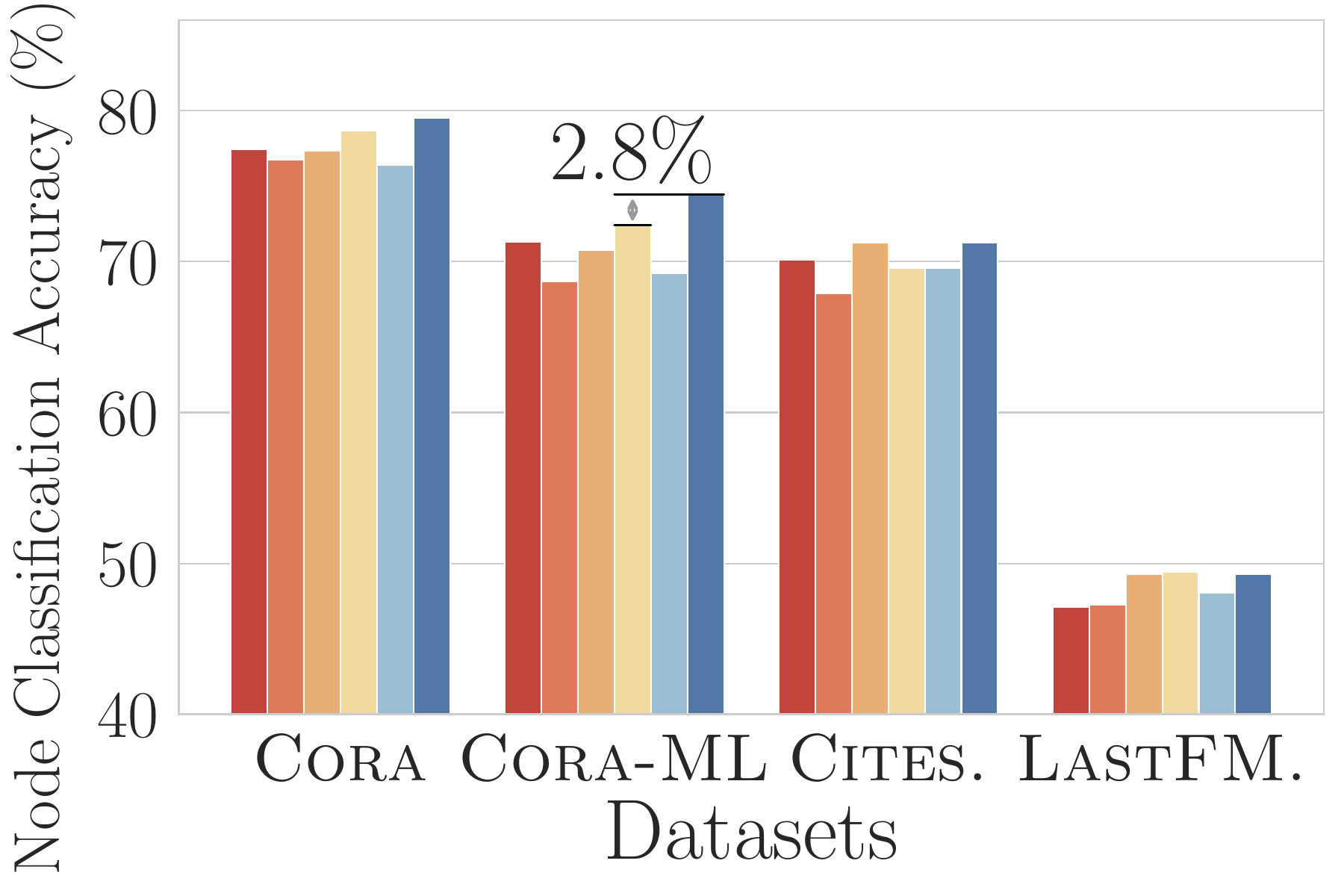}
        \Description[<short description>]{<long description>}
        \caption{\structack}
        \label{fig:exp:q3_rgcn_structack}
    \end{subfigure} 
    \hfill
    \begin{subfigure}{0.195\linewidth}
        \centering
        \includegraphics[width=\linewidth]{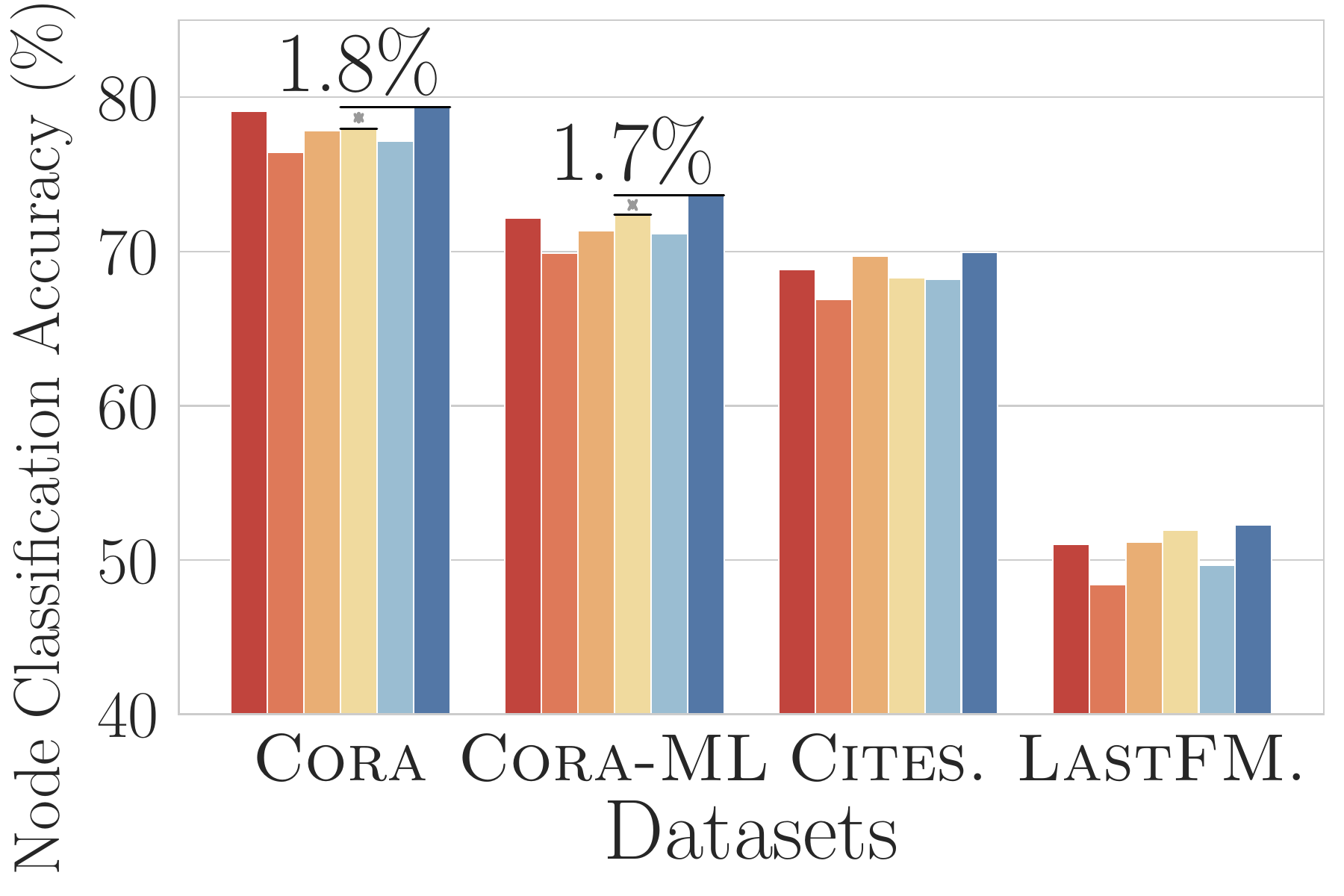}
        \Description[<short description>]{<long description>}
        \caption{\dice}
        \label{fig:exp:q3_rgcn_dice}
    \end{subfigure} 
    \hfill
    \begin{subfigure}{0.195\linewidth}
        \centering
        \includegraphics[width=\linewidth]{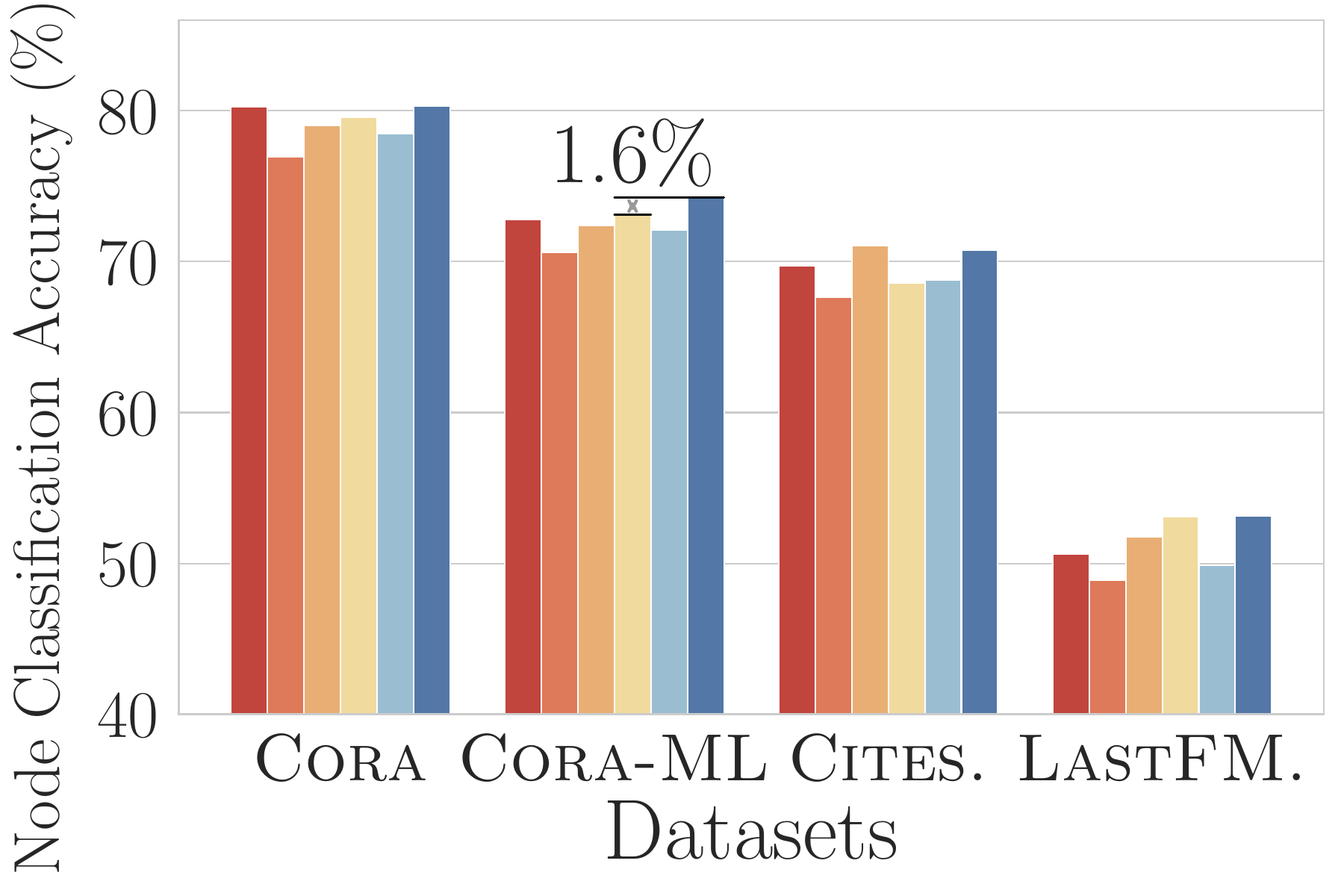}
        \Description[<short description>]{<long description>}
        \caption{\random}
        \label{fig:exp:q3_rgcn_random}
    \end{subfigure} \\[-5pt]
    \textbf{\mediangcn}: \hfill \; \\
    \begin{subfigure}{0.195\linewidth}
        \centering
        \includegraphics[width=\textwidth]{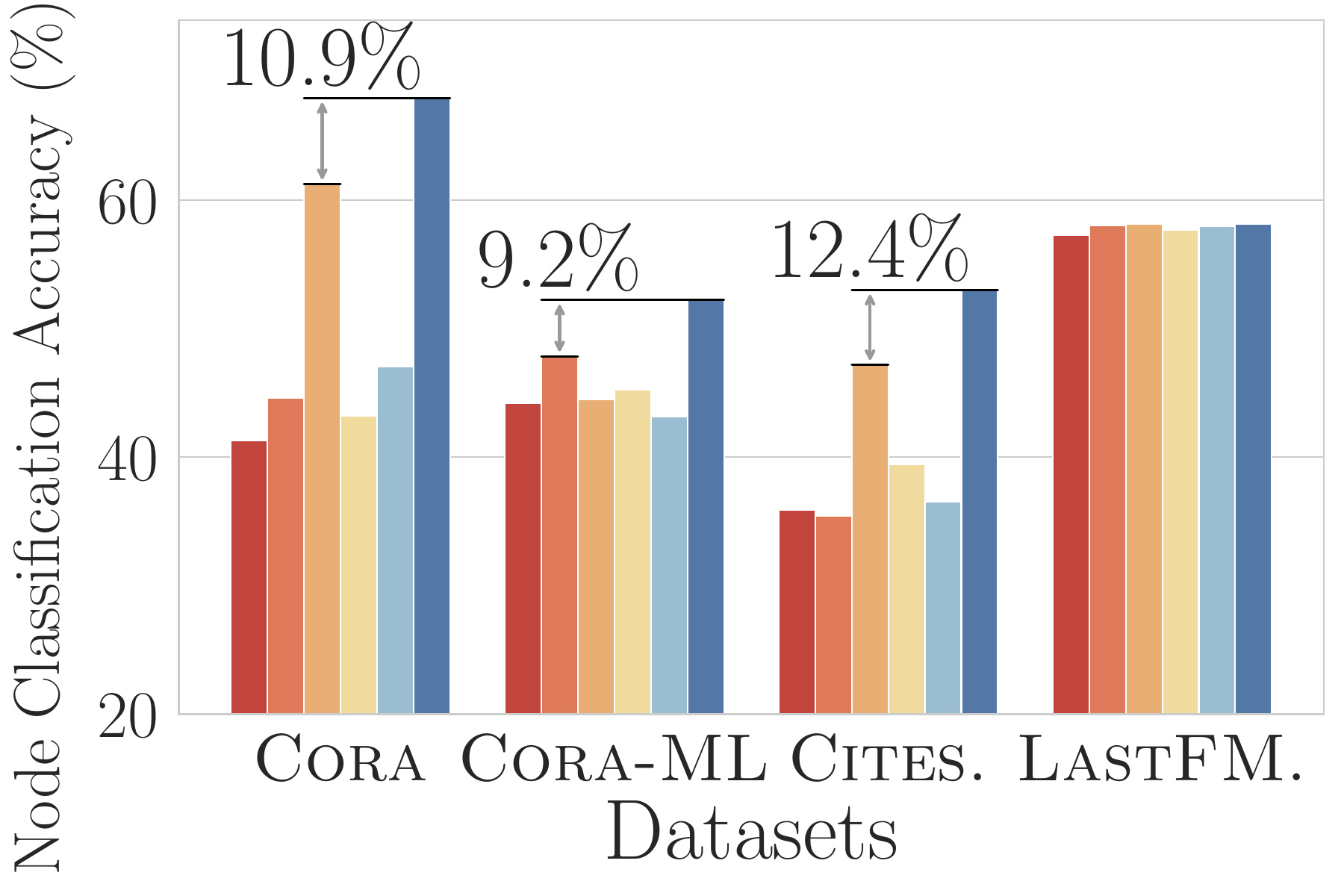}
        \Description[<short description>]{<long description>}
        \caption{\metattack}
        \label{fig:exp:q3_mediangcn_metattack}
    \end{subfigure}
    \hfill
    \begin{subfigure}{0.195\linewidth}
        \centering
        \includegraphics[width=\linewidth]{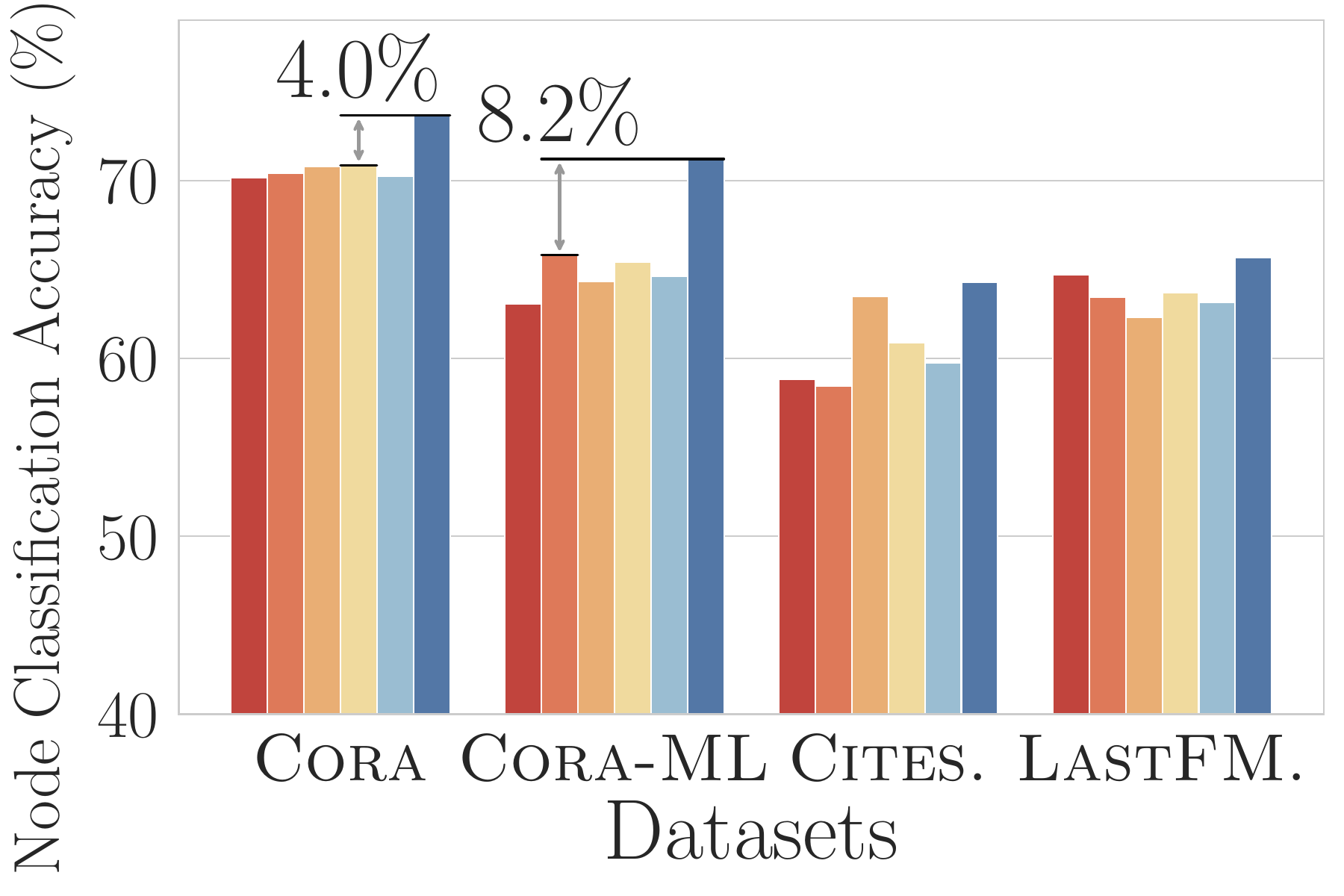}
        \Description[<short description>]{<long description>}
        \caption{\pgd}
        \label{fig:exp:q3_mediangcn_pgd}
    \end{subfigure} 
    \hfill
    \begin{subfigure}{0.195\linewidth}
        \centering
        \includegraphics[width=\linewidth]{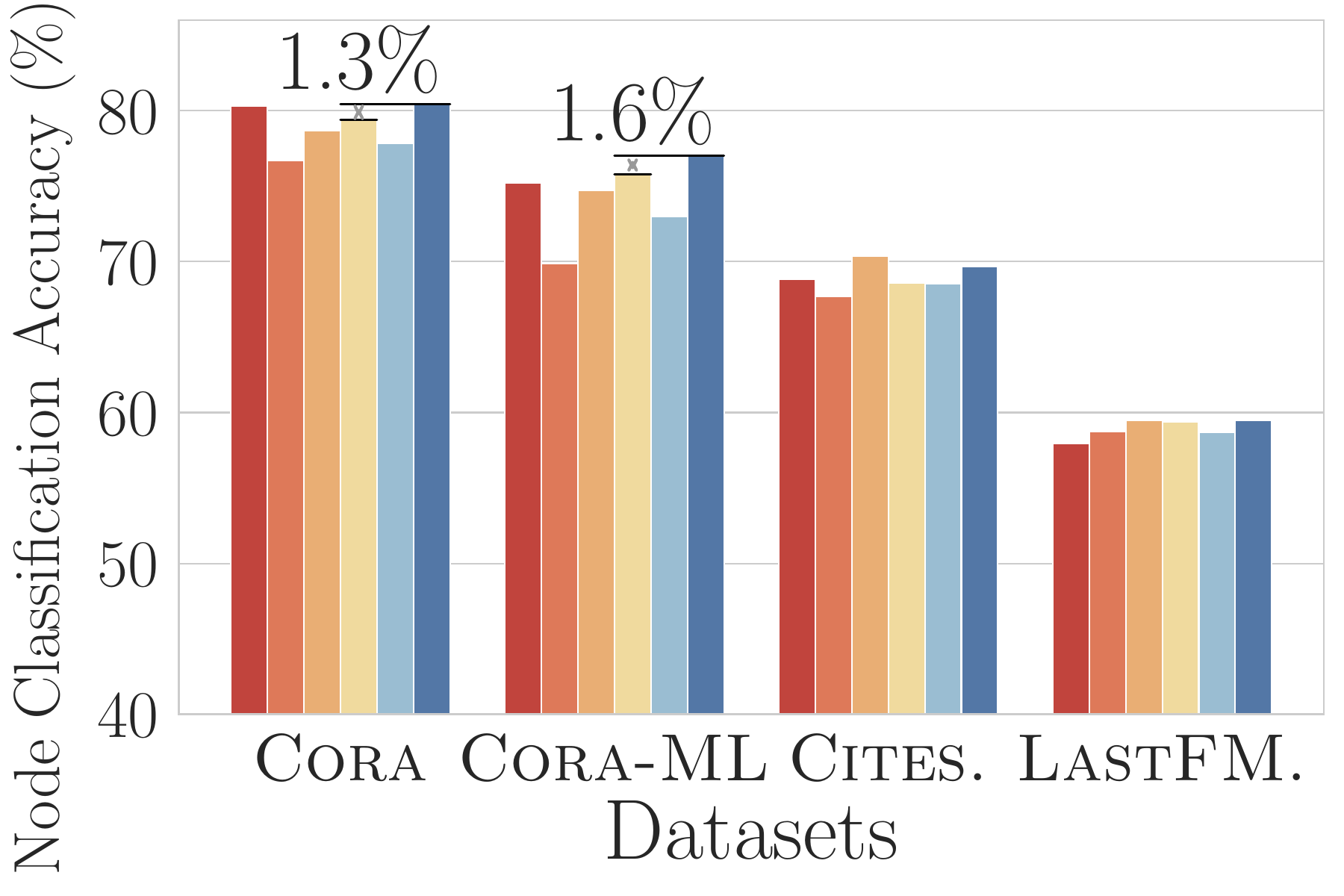}
        \Description[<short description>]{<long description>}
        \caption{\structack}
        \label{fig:exp:q3_mediangcn_structack}
    \end{subfigure} 
    \hfill
    \begin{subfigure}{0.195\linewidth}
        \centering
        \includegraphics[width=\linewidth]{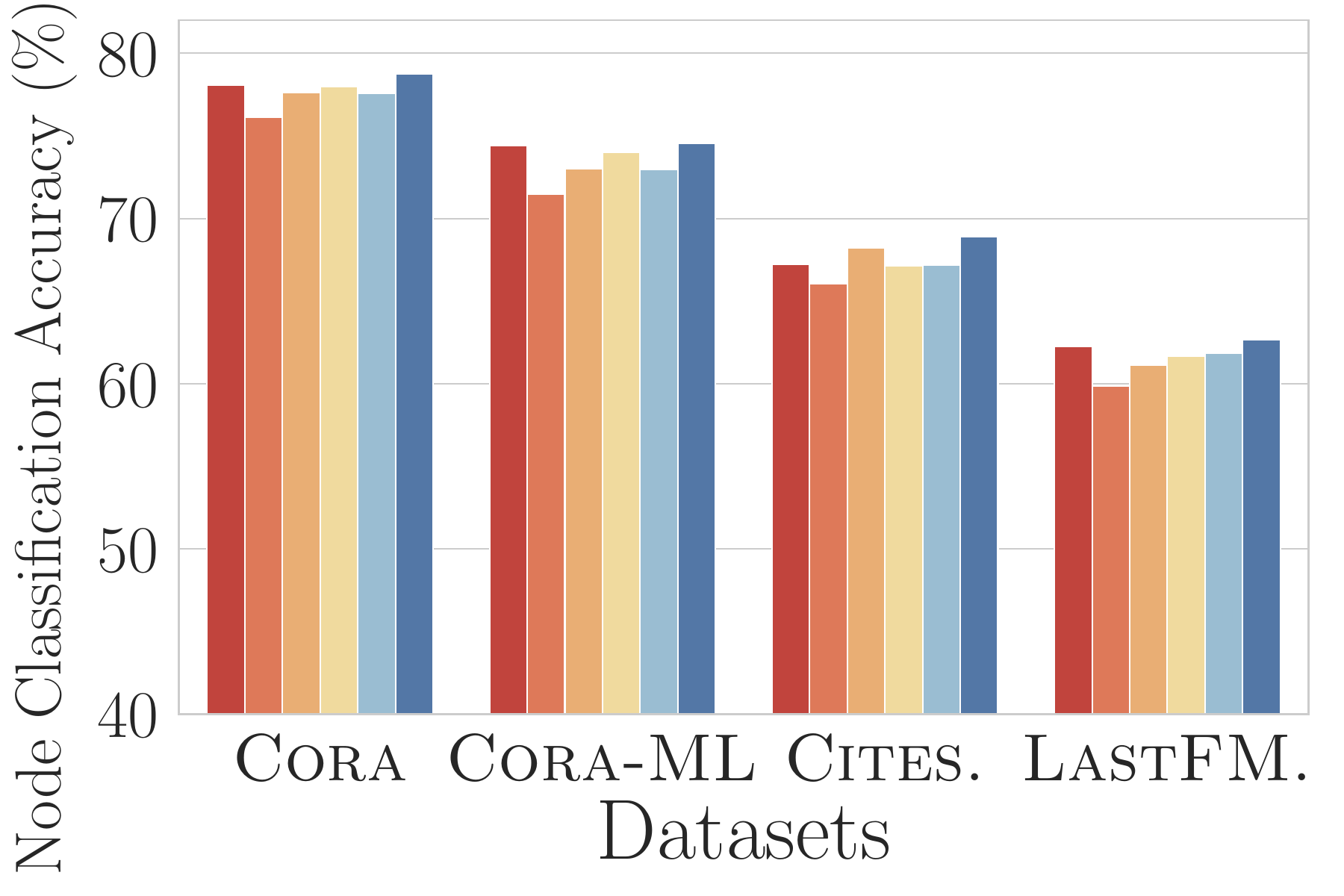}
        \Description[<short description>]{<long description>}
        \caption{\dice}
        \label{fig:exp:q3_mediangcn_dice}
    \end{subfigure} 
    \hfill
    \begin{subfigure}{0.195\linewidth}
        \centering
        \includegraphics[width=\linewidth]{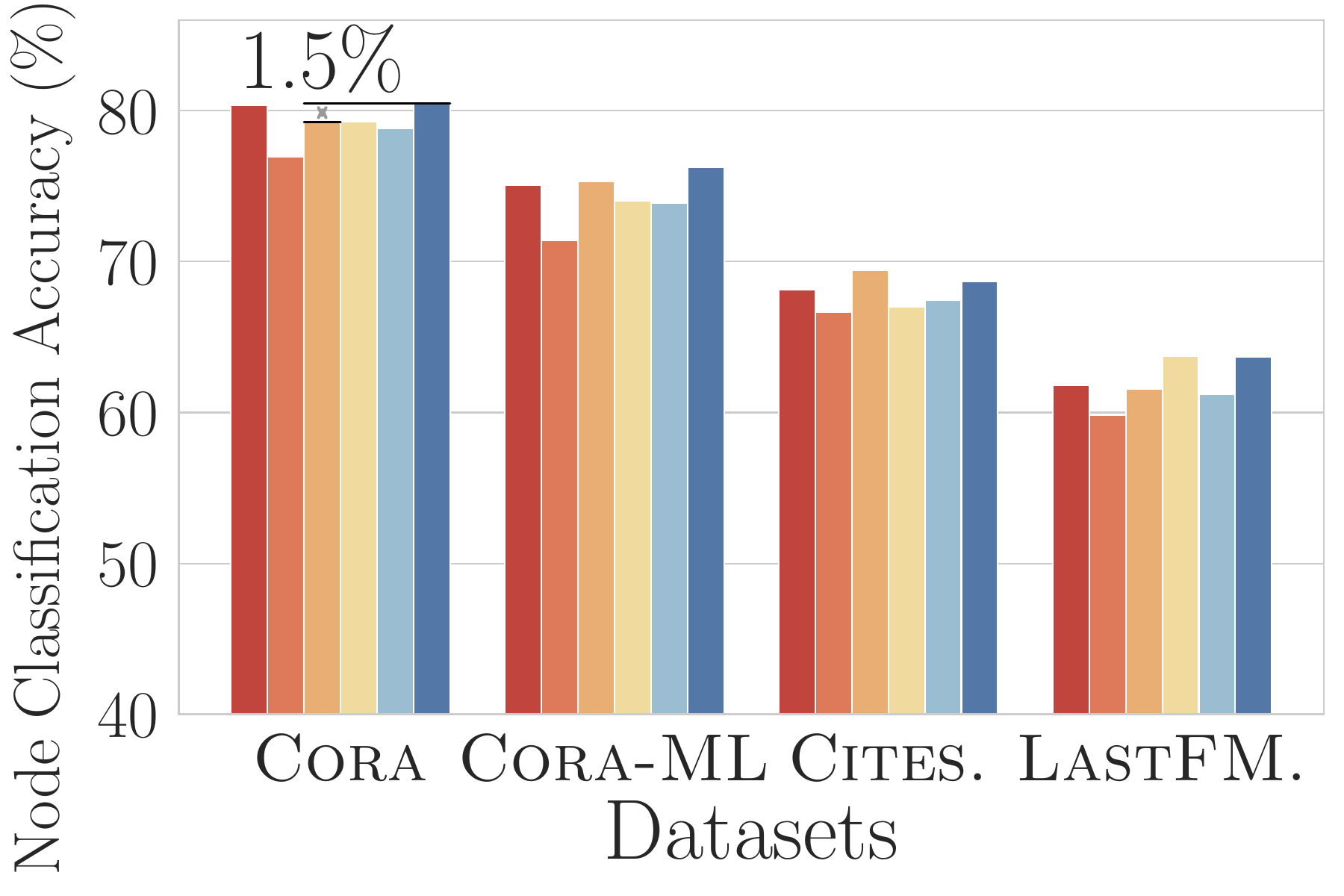}
        \Description[<short description>]{<long description>}
        \caption{\random}
        \label{fig:exp:q3_mediangcn_random}
    \end{subfigure} 
    \vspace{-1mm}
    \caption{
    \underline{\smash{Further application:}} \model consistently improves the node classification accuracy
    of \gcn ((a)-(e)), \rgcn ((f)-(j)), and \mediangcn ((k)-(o)). \kijung{Even for \mediangcn (a robust GNN), \model yields up to 12.4\% better accuracy than the best competitor.}
    }
    \label{fig:exp:q3}
\end{figure*}

\smallsection{Ablation studies.}
We further examine how much each module contributes to the performance of \model.
We compare 8 variants of \model, each consisting of a different combination of the three modules used in \model ($M_G$, $M_S$, and $M_P$).
In summary, the results indicate that each module of \model positively contributes to performance improvements.
The detailed results are in Online Appendix \textsc{F.1}~\cite{code}.

\subsection{Q2 \& Q3. Effectiveness of \measure}\label{sec:exp:q2}
We show that \measure is non-bypassable and sensitive (Properties~\ref{prop:nonbypassable} \& \ref{prop:sensitive}) to underscore its efficacy as a noticeability measure.

\smallsection{Experiment design.}
\kijung{For each dataset and each noticeability measure $U$, 
we consider two versions of $\pgd$: (1) \textbf{original}: we generate $\Delta = 0.1\Vert A\Vert_1$ attack edges using $\pgd$ and add them to the original (unattacked) graph one by one in  random order,\footnote{Alternatively, attack edges could be generated from scratch for every budget (i.e., the count of added attack edges), but this process was prohibitively time-consuming.} and (2) \textbf{adaptive}: the same attack edges are added one by one greedily to minimize the noticeability score w.r.t. $U$ (as described in Def.~\ref{defn:adaptive}).
In Table~\ref{tab:exp:q2_pgd}, 
for original (\textcolor{SkyBlue}{\textbf{sky blue curves}}) and adaptive (\textcolor{DarkBlue}{\textbf{dark blue curves}}) attacks,
we present the trade-off curves between the node classification accuracy and the noticeability scores (w.r.t. each measure) as attack edges are added one by one (from $0$ to $0.1\Vert A\Vert_{1}$). The node classification accuracy is evaluated using a GCN model trained on the original graph $G$ without attack.}

\smallsection{Competitors.}
We compare \measure with four measures: \textsc{DegreeKS}, \textsc{ClsCoefKS}, \textsc{DegreeLR}, and \textsc{HomophKS} (see Sec.~\ref{sec:prelim}).

\smallsection{Evaluation.}
To analyze measures in terms of \textit{non-bypassability} (Property~\ref{prop:nonbypassable}), we propose a metric called \textit{bypassable rate}. 
Roughly speaking, the bypassable rate is the expected reduction in noticeability with the same attack performance.
Specifically, it is defined as
\resizebox{2.5cm}{!}{$1 \minus {\dfrac{\text{dark-blue area  } \darkbluebox}{\text{sky-blue area  } \skybluebox}}$} (see Table~\ref{tab:exp:q2_pgd}).
To analyze the measures in terms of \textit{sensitivity} (Property~\ref{prop:sensitive}), we marked the noticeability measure of the adaptive attack graph at a low attack rate $\gamma = 2\%$ with a red star (\redstar{$\medstar$}).
A sensitive measure is supposed to give a considerable noticeability score even for such a low attack rate.

\smallsection{Results.}
In Table~\ref{tab:exp:q2_pgd}, we demonstrate the experimental results using \pgd (see Online Appendix~\textsc{F.3}~\cite{code} for additional results using different attack methods), where \measure consistently achieves lower bypassable rates (i.e., is less bypassable) compared to the existing measures.
Specifically, \measure was $\mathbf{85.19\%}$ {\bf less bypassable} than \textsc{DegreeKS}, which is the second best, in the \lastfmasia dataset.
In Appendix~\ref{appendix:crown_our}, we extend the analysis of bypassability (Sec.~\ref{sec:limit:bypassable}) and overlookingness (Sec.~\ref{sec:limit:overlooking}) to \measure. 
In Online Appendix \textsc{C.2}~\cite{code}, we provide an in-depth analysis regarding bypassability against various adaptive attack scenarios. 

Regarding \textit{sensitivity} (Property~\ref{prop:sensitive}), \measure can effectively detect attack edges even when the number is small.
From the red stars (\redstar{$\medstar$}) in Table~\ref{tab:exp:q2_pgd}, which corresponds to a 2\% attack rate (see the description of ``Evaluation'' above), 
we observe that \measure gives a considerable noticeability score even for such a low attack rate, while all the existing measures fail to do so (see also Sec.~\ref{sec:limit:overlooking}).

\section{Further Applications}\label{sec:exp:q3}
\vspace{-0.5mm}
In this section, we show additional applications of \model to enhance the performance of GNNs on downstream tasks, e.g., node classification.
Based on the edge scores output by \model, we can remove suspicious (i.e., attack-like) edges with low scores.
Compared to training GNNs on attacked graphs as they are, training on graphs after removing suspicious edges is supposed to perform better.

\smallsection{Experiment design.} 
We consider the node classification task which is widely considered to evaluate the performance of GNNs~\citep{maekawa2022beyond,xiao2022graph}.
We use three different GNN models, including robust GNNs:
\gcn~\cite{kipf2016semi}, \rgcn~\cite{zhu2019robust}, and \mediangcn~\cite{chen2021understanding2}.
Given an attacked graph $\hat{G}$, we train \model on $\hat{G}$, obtain the edge scores of all the edges in $\hat{G}$, and remove the suspicious edges with the lowest scores to obtain $\hat{G}_{\text{LEO}}$.
We compare the performance of the aforementioned GNN models when trained on $\hat{G}$ and when trained on $\hat{G}_{\text{LEO}}$.
For \model and all the competitors (to be introduced later),
we consistently use attack rate $\gamma = 10\%$ and remove the same number of suspicious edges.

\smallsection{Competitors.}
We compare \model with four competitors that be used to identify suspicious edges:
SVD Filtering~\cite{entezari2020all}, 
Jaccard Filtering~\cite{wu2019adversarial}, 
GNNGuard Filtering~\cite{zhang2020gnnguard}, and
MetaGC Filtering~\cite{jo2023robust}.
For their details, see Online Appendix \textsc{E.3}~\cite{code}.

\smallsection{Results.} 
As shown in Fig.~\ref{fig:exp:q3}, \model consistently improves the performance of not only vanilla \gcn but also robust GNNs (\rgcn and \mediangcn).
Specifically, \model enhances the node classification accuracy of \mediangcn (a robust GNN),
\kijung{leading to up to 12.4\% better accuracy} compared to the second-best method (see Fig.~\ref{fig:exp:q3}(f)).

\smallsection{Additional experiments.} 
In Online Appendix \textsc{F.3}~\cite{code}, we show 
(1) \model enhances GNNs also on \textit{heterophilic graphs}, and
(2) \model can be integrated into and enhance \textit{graph structure learning} methods.

\section{Conclusion}
\label{sec:conclusion}
In this work, we revisit the problem of measuring the unnoticeability of graph adversarial attacks.
We observe two critical limitations of the existing unnoticeability measures: 
they are bypassable (Fig.~\ref{fig:crown:bypass}) and overlooking (Fig.~\ref{fig:crown:overlooking}).
We propose a novel noticeability measure, \measure, that is less bypassable and less overlooking,
by adopting \model (\textit{\textbf{\myuline{l}}earnable} \textbf{\myuline{e}}dge sc\textbf{\myuline{o}}rer) and imbalance-aware aggregation, respectively.
Specifically, \measure is $0.38\times$ to $5.75\times$ less bypassable than the strongest baseline and returns considerable noticeability scores even with a small attack rate (Tab.~\ref{tab:exp:q2_pgd}).
In addition, \model outperforms eleven competitors in detecting attack edges in 23 out of 28 cases across five different attack methods (Tab.~\ref{tab:exp:q1}), and consistently boosts the performance of GNNs \kijung{(including those known for their robustness)} by removing attack-like edges  (Fig.~\ref{fig:exp:q3}).

\vspace{1mm}
\smallsection{Acknowledgements.}
This work was partly supported by the National Research Foundation of Korea (NRF) grant funded
by the Korea government (MSIT) (No. RS-2024-00406985) (50\%).
This work was partly supported by Institute of Information \& Communications Technology Planning \& Evaluation (IITP) grant funded by the Korea government (MSIT) (No. 2022-0-00871 / RS-2022-II220871, Development of AI Autonomy and Knowledge Enhancement for AI Agent Collaboration, 40\%) (No. RS-2019-II190075, Artificial Intelligence Graduate School Program (KAIST), 10\%).

\bibliographystyle{ACM-Reference-Format}
\balance
\bibliography{ref}


\begin{thebibliography}{48}


\ifx \showCODEN    \undefined \def \showCODEN     #1{\unskip}     \fi
\ifx \showISBNx    \undefined \def \showISBNx     #1{\unskip}     \fi
\ifx \showISBNxiii \undefined \def \showISBNxiii  #1{\unskip}     \fi
\ifx \showISSN     \undefined \def \showISSN      #1{\unskip}     \fi
\ifx \showLCCN     \undefined \def \showLCCN      #1{\unskip}     \fi
\ifx \shownote     \undefined \def \shownote      #1{#1}          \fi
\ifx \showarticletitle \undefined \def \showarticletitle #1{#1}   \fi
\ifx \showURL      \undefined \def \showURL       {\relax}        \fi
\providecommand\bibfield[2]{#2}
\providecommand\bibinfo[2]{#2}
\providecommand\natexlab[1]{#1}
\providecommand\showeprint[2][]{arXiv:#2}

\bibitem[Barbieri et~al\mbox{.}(2014)]%
        {barbieri2014follow}
\bibfield{author}{\bibinfo{person}{Nicola Barbieri}, \bibinfo{person}{Francesco Bonchi}, {and} \bibinfo{person}{Giuseppe Manco}.} \bibinfo{year}{2014}\natexlab{}.
\newblock \showarticletitle{Who to follow and why: link prediction with explanations}. In \bibinfo{booktitle}{\emph{ACM International Conference on Knowledge Discovery and Data Mining}}.
\newblock


\bibitem[Carlini and Wagner(2017)]%
        {carlini2017towards}
\bibfield{author}{\bibinfo{person}{Nicholas Carlini} {and} \bibinfo{person}{David Wagner}.} \bibinfo{year}{2017}\natexlab{}.
\newblock \showarticletitle{Towards evaluating the robustness of neural networks}. In \bibinfo{booktitle}{\emph{IEEE Symposium on Security and Privacy}}.
\newblock


\bibitem[Chen et~al\mbox{.}(2021a)]%
        {chen2021understanding2}
\bibfield{author}{\bibinfo{person}{Liang Chen}, \bibinfo{person}{Jintang Li}, \bibinfo{person}{Qibiao Peng}, \bibinfo{person}{Yang Liu}, \bibinfo{person}{Zibin Zheng}, {and} \bibinfo{person}{Carl Yang}.} \bibinfo{year}{2021}\natexlab{a}.
\newblock \showarticletitle{Understanding structural vulnerability in graph convolutional networks}. In \bibinfo{booktitle}{\emph{International Joint Conference on Artificial Intelligence}}.
\newblock


\bibitem[Chen et~al\mbox{.}(2021b)]%
        {chen2021understanding}
\bibfield{author}{\bibinfo{person}{Yongqiang Chen}, \bibinfo{person}{Han Yang}, \bibinfo{person}{Yonggang Zhang}, \bibinfo{person}{MA KAILI}, \bibinfo{person}{Tongliang Liu}, \bibinfo{person}{Bo Han}, {and} \bibinfo{person}{James Cheng}.} \bibinfo{year}{2021}\natexlab{b}.
\newblock \showarticletitle{Understanding and Improving Graph Injection Attack by Promoting Unnoticeability}. In \bibinfo{booktitle}{\emph{International Conference on Learning Representations}}.
\newblock


\bibitem[Dai et~al\mbox{.}(2018)]%
        {dai2018adversarial}
\bibfield{author}{\bibinfo{person}{Hanjun Dai}, \bibinfo{person}{Hui Li}, \bibinfo{person}{Tian Tian}, \bibinfo{person}{Xin Huang}, \bibinfo{person}{Lin Wang}, \bibinfo{person}{Jun Zhu}, {and} \bibinfo{person}{Le Song}.} \bibinfo{year}{2018}\natexlab{}.
\newblock \showarticletitle{Adversarial attack on graph structured data}. In \bibinfo{booktitle}{\emph{International Conference on Machine Learning}}.
\newblock


\bibitem[Entezari et~al\mbox{.}(2020)]%
        {entezari2020all}
\bibfield{author}{\bibinfo{person}{Negin Entezari}, \bibinfo{person}{Saba~A Al-Sayouri}, \bibinfo{person}{Amirali Darvishzadeh}, {and} \bibinfo{person}{Evangelos~E Papalexakis}.} \bibinfo{year}{2020}\natexlab{}.
\newblock \showarticletitle{All you need is low (rank) defending against adversarial attacks on graphs}. In \bibinfo{booktitle}{\emph{ACM International Conference on Web Search and Data Mining}}.
\newblock


\bibitem[Fatemi et~al\mbox{.}(2021)]%
        {fatemi2021slaps}
\bibfield{author}{\bibinfo{person}{Bahare Fatemi}, \bibinfo{person}{Layla El~Asri}, {and} \bibinfo{person}{Seyed~Mehran Kazemi}.} \bibinfo{year}{2021}\natexlab{}.
\newblock \showarticletitle{SLAPS: Self-supervision improves structure learning for graph neural networks}. In \bibinfo{booktitle}{\emph{Conference on Neural Information Processing Systems}}.
\newblock


\bibitem[Gasteiger et~al\mbox{.}(2019)]%
        {gasteiger2019predict}
\bibfield{author}{\bibinfo{person}{Johannes Gasteiger}, \bibinfo{person}{Aleksandar Bojchevski}, {and} \bibinfo{person}{Stephan G{\"u}nnemann}.} \bibinfo{year}{2019}\natexlab{}.
\newblock \showarticletitle{Predict then Propagate: Graph Neural Networks meet Personalized PageRank}. In \bibinfo{booktitle}{\emph{International Conference on Learning Representations}}.
\newblock


\bibitem[Goodfellow et~al\mbox{.}(2015)]%
        {goodfellow2014explaining}
\bibfield{author}{\bibinfo{person}{Ian~J Goodfellow}, \bibinfo{person}{Jonathon Shlens}, {and} \bibinfo{person}{Christian Szegedy}.} \bibinfo{year}{2015}\natexlab{}.
\newblock \showarticletitle{Explaining and harnessing adversarial examples}. In \bibinfo{booktitle}{\emph{International Conference on Learning Representations}}.
\newblock


\bibitem[Gosch et~al\mbox{.}(2023)]%
        {gosch2023revisiting}
\bibfield{author}{\bibinfo{person}{Lukas Gosch}, \bibinfo{person}{Daniel Sturm}, \bibinfo{person}{Simon Geisler}, {and} \bibinfo{person}{Stephan G{\"u}nnemann}.} \bibinfo{year}{2023}\natexlab{}.
\newblock \showarticletitle{Revisiting Robustness in Graph Machine Learning}. In \bibinfo{booktitle}{\emph{International Conference on Learning Representations}}.
\newblock


\bibitem[Hamilton et~al\mbox{.}(2017)]%
        {hamilton2017inductive}
\bibfield{author}{\bibinfo{person}{Will Hamilton}, \bibinfo{person}{Zhitao Ying}, {and} \bibinfo{person}{Jure Leskovec}.} \bibinfo{year}{2017}\natexlab{}.
\newblock \showarticletitle{Inductive representation learning on large graphs}. In \bibinfo{booktitle}{\emph{Conference on Neural Information Processing Systems}}.
\newblock


\bibitem[Hussain et~al\mbox{.}(2021)]%
        {hussain2021structack}
\bibfield{author}{\bibinfo{person}{Hussain Hussain}, \bibinfo{person}{Tomislav Duricic}, \bibinfo{person}{Elisabeth Lex}, \bibinfo{person}{Denis Helic}, \bibinfo{person}{Markus Strohmaier}, {and} \bibinfo{person}{Roman Kern}.} \bibinfo{year}{2021}\natexlab{}.
\newblock \showarticletitle{Structack: Structure-based adversarial attacks on graph neural networks}. In \bibinfo{booktitle}{\emph{ACM Conference on Hypertext and Social Media}}.
\newblock


\bibitem[Jin et~al\mbox{.}(2021)]%
        {jin2021adversarial}
\bibfield{author}{\bibinfo{person}{Wei Jin}, \bibinfo{person}{Yaxing Li}, \bibinfo{person}{Han Xu}, \bibinfo{person}{Yiqi Wang}, \bibinfo{person}{Shuiwang Ji}, \bibinfo{person}{Charu Aggarwal}, {and} \bibinfo{person}{Jiliang Tang}.} \bibinfo{year}{2021}\natexlab{}.
\newblock \showarticletitle{Adversarial attacks and defenses on graphs}.
\newblock \bibinfo{journal}{\emph{ACM SIGKDD Explorations Newsletter}} (\bibinfo{year}{2021}).
\newblock


\bibitem[Jo et~al\mbox{.}(2023)]%
        {jo2023robust}
\bibfield{author}{\bibinfo{person}{Hyeonsoo Jo}, \bibinfo{person}{Fanchen Bu}, {and} \bibinfo{person}{Kijung Shin}.} \bibinfo{year}{2023}\natexlab{}.
\newblock \showarticletitle{Robust Graph Clustering via Meta Weighting for Noisy Graphs}. In \bibinfo{booktitle}{\emph{ACM International Conference on Information and Knowledge Management}}.
\newblock


\bibitem[Jo et~al\mbox{.}(2025)]%
        {code}
\bibfield{author}{\bibinfo{person}{Hyeonsoo Jo}, \bibinfo{person}{Hyunjin Hwang}, \bibinfo{person}{Fanchen Bu}, \bibinfo{person}{Soo~Yong Lee}, \bibinfo{person}{Chanyoung Park}, {and} \bibinfo{person}{Kijung Shin}.} \bibinfo{year}{2025}\natexlab{}.
\newblock \bibinfo{title}{On Measuring Unnoticeability of Graph Adversarial Attacks: Observations, New Measure, and Applications: Online Appendix}.
\newblock \bibinfo{howpublished}{\url{https://github.com/HyeonsooJo/Resources-for-HideNSeek/blob/main/OnlineAppendix.pdf}}.
\newblock


\bibitem[King(1998)]%
        {king1998unifying}
\bibfield{author}{\bibinfo{person}{Gary King}.} \bibinfo{year}{1998}\natexlab{}.
\newblock \bibinfo{booktitle}{\emph{Unifying political methodology: The likelihood theory of statistical inference}}.
\newblock \bibinfo{publisher}{University of Michigan Press}.
\newblock


\bibitem[Kipf and Welling(2017)]%
        {kipf2016semi}
\bibfield{author}{\bibinfo{person}{Thomas~N Kipf} {and} \bibinfo{person}{Max Welling}.} \bibinfo{year}{2017}\natexlab{}.
\newblock \showarticletitle{Semi-supervised classification with graph convolutional networks}. In \bibinfo{booktitle}{\emph{International Conference on Learning Representations}}.
\newblock


\bibitem[Kurakin et~al\mbox{.}(2018)]%
        {kurakin2018adversarial}
\bibfield{author}{\bibinfo{person}{Alexey Kurakin}, \bibinfo{person}{Ian~J Goodfellow}, {and} \bibinfo{person}{Samy Bengio}.} \bibinfo{year}{2018}\natexlab{}.
\newblock \showarticletitle{Adversarial examples in the physical world}.
\newblock In \bibinfo{booktitle}{\emph{Artificial intelligence safety and security}}. \bibinfo{publisher}{Chapman and Hall/CRC}, \bibinfo{pages}{99--112}.
\newblock


\bibitem[Lee et~al\mbox{.}(2018)]%
        {lee2018graph}
\bibfield{author}{\bibinfo{person}{John~Boaz Lee}, \bibinfo{person}{Ryan Rossi}, {and} \bibinfo{person}{Xiangnan Kong}.} \bibinfo{year}{2018}\natexlab{}.
\newblock \showarticletitle{Graph classification using structural attention}. In \bibinfo{booktitle}{\emph{ACM International Conference on Knowledge Discovery and Data Mining}}.
\newblock


\bibitem[Ma et~al\mbox{.}(2020)]%
        {ma2020towards}
\bibfield{author}{\bibinfo{person}{Jiaqi Ma}, \bibinfo{person}{Shuangrui Ding}, {and} \bibinfo{person}{Qiaozhu Mei}.} \bibinfo{year}{2020}\natexlab{}.
\newblock \showarticletitle{Towards more practical adversarial attacks on graph neural networks}. In \bibinfo{booktitle}{\emph{Conference on Neural Information Processing Systems}}.
\newblock


\bibitem[Madry et~al\mbox{.}(2018)]%
        {madry2018towards}
\bibfield{author}{\bibinfo{person}{Aleksander Madry}, \bibinfo{person}{Aleksandar Makelov}, \bibinfo{person}{Ludwig Schmidt}, \bibinfo{person}{Dimitris Tsipras}, {and} \bibinfo{person}{Adrian Vladu}.} \bibinfo{year}{2018}\natexlab{}.
\newblock \showarticletitle{Towards Deep Learning Models Resistant to Adversarial Attacks}. In \bibinfo{booktitle}{\emph{International Conference on Learning Representations}}.
\newblock


\bibitem[Maekawa et~al\mbox{.}(2022)]%
        {maekawa2022beyond}
\bibfield{author}{\bibinfo{person}{Seiji Maekawa}, \bibinfo{person}{Koki Noda}, \bibinfo{person}{Yuya Sasaki}, {et~al\mbox{.}}} \bibinfo{year}{2022}\natexlab{}.
\newblock \showarticletitle{Beyond real-world benchmark datasets: An empirical study of node classification with GNNs}. In \bibinfo{booktitle}{\emph{Conference on Neural Information Processing Systems}}.
\newblock


\bibitem[Massey~Jr(1951)]%
        {massey1951kolmogorov}
\bibfield{author}{\bibinfo{person}{Frank~J Massey~Jr}.} \bibinfo{year}{1951}\natexlab{}.
\newblock \showarticletitle{The Kolmogorov-Smirnov test for goodness of fit}.
\newblock \bibinfo{journal}{\emph{Journal of the American statistical Association}} \bibinfo{volume}{46}, \bibinfo{number}{253} (\bibinfo{year}{1951}), \bibinfo{pages}{68--78}.
\newblock


\bibitem[Mujkanovic et~al\mbox{.}(2022)]%
        {mujkanovic2022defenses}
\bibfield{author}{\bibinfo{person}{Felix Mujkanovic}, \bibinfo{person}{Simon Geisler}, \bibinfo{person}{Stephan G{\"u}nnemann}, {and} \bibinfo{person}{Aleksandar Bojchevski}.} \bibinfo{year}{2022}\natexlab{}.
\newblock \showarticletitle{Are Defenses for Graph Neural Networks Robust?}. In \bibinfo{booktitle}{\emph{Conference on Neural Information Processing Systems}}.
\newblock


\bibitem[Ou et~al\mbox{.}(2016)]%
        {ou2016asymmetric}
\bibfield{author}{\bibinfo{person}{Mingdong Ou}, \bibinfo{person}{Peng Cui}, \bibinfo{person}{Jian Pei}, \bibinfo{person}{Ziwei Zhang}, {and} \bibinfo{person}{Wenwu Zhu}.} \bibinfo{year}{2016}\natexlab{}.
\newblock \showarticletitle{Asymmetric transitivity preserving graph embedding}. In \bibinfo{booktitle}{\emph{ACM International Conference on Knowledge Discovery and Data Mining}}.
\newblock


\bibitem[Platonov et~al\mbox{.}(2022)]%
        {platonov2022critical}
\bibfield{author}{\bibinfo{person}{Oleg Platonov}, \bibinfo{person}{Denis Kuznedelev}, \bibinfo{person}{Michael Diskin}, \bibinfo{person}{Artem Babenko}, {and} \bibinfo{person}{Liudmila Prokhorenkova}.} \bibinfo{year}{2022}\natexlab{}.
\newblock \showarticletitle{A critical look at the evaluation of GNNs under heterophily: Are we really making progress?}. In \bibinfo{booktitle}{\emph{International Conference on Learning Representations}}.
\newblock


\bibitem[Rozemberczki et~al\mbox{.}(2021)]%
        {rozemberczki2021multi}
\bibfield{author}{\bibinfo{person}{Benedek Rozemberczki}, \bibinfo{person}{Carl Allen}, {and} \bibinfo{person}{Rik Sarkar}.} \bibinfo{year}{2021}\natexlab{}.
\newblock \showarticletitle{Multi-scale attributed node embedding}.
\newblock \bibinfo{journal}{\emph{Journal of Complex Networks}} \bibinfo{volume}{9}, \bibinfo{number}{2} (\bibinfo{year}{2021}), \bibinfo{pages}{cnab014}.
\newblock


\bibitem[Rozemberczki and Sarkar(2020)]%
        {rozemberczki2020characteristic}
\bibfield{author}{\bibinfo{person}{Benedek Rozemberczki} {and} \bibinfo{person}{Rik Sarkar}.} \bibinfo{year}{2020}\natexlab{}.
\newblock \showarticletitle{Characteristic functions on graphs: Birds of a feather, from statistical descriptors to parametric models}. In \bibinfo{booktitle}{\emph{ACM International Conference on Information and Knowledge Management}}.
\newblock


\bibitem[Sen et~al\mbox{.}(2008)]%
        {sen2008collective}
\bibfield{author}{\bibinfo{person}{Prithviraj Sen}, \bibinfo{person}{Galileo Namata}, \bibinfo{person}{Mustafa Bilgic}, \bibinfo{person}{Lise Getoor}, \bibinfo{person}{Brian Galligher}, {and} \bibinfo{person}{Tina Eliassi-Rad}.} \bibinfo{year}{2008}\natexlab{}.
\newblock \showarticletitle{Collective classification in network data}.
\newblock \bibinfo{journal}{\emph{AI magazine}} \bibinfo{volume}{29}, \bibinfo{number}{3} (\bibinfo{year}{2008}), \bibinfo{pages}{93--93}.
\newblock


\bibitem[Smirnov(1939)]%
        {smirnov1939estimate}
\bibfield{author}{\bibinfo{person}{Nikolai~V Smirnov}.} \bibinfo{year}{1939}\natexlab{}.
\newblock \showarticletitle{Estimate of deviation between empirical distribution functions in two independent samples}.
\newblock \bibinfo{journal}{\emph{Bulletin Moscow University}} \bibinfo{volume}{2}, \bibinfo{number}{2} (\bibinfo{year}{1939}), \bibinfo{pages}{3--16}.
\newblock


\bibitem[Sun et~al\mbox{.}(2022)]%
        {sun2022adversarial}
\bibfield{author}{\bibinfo{person}{Lichao Sun}, \bibinfo{person}{Yingtong Dou}, \bibinfo{person}{Carl Yang}, \bibinfo{person}{Kai Zhang}, \bibinfo{person}{Ji Wang}, \bibinfo{person}{S~Yu Philip}, \bibinfo{person}{Lifang He}, {and} \bibinfo{person}{Bo Li}.} \bibinfo{year}{2022}\natexlab{}.
\newblock \showarticletitle{Adversarial attack and defense on graph data: A survey}.
\newblock \bibinfo{journal}{\emph{IEEE Transactions on Knowledge and Data Engineering}} \bibinfo{volume}{35}, \bibinfo{number}{8} (\bibinfo{year}{2022}), \bibinfo{pages}{7693--7711}.
\newblock


\bibitem[Sun et~al\mbox{.}(2020)]%
        {sun2020adversarial}
\bibfield{author}{\bibinfo{person}{Yiwei Sun}, \bibinfo{person}{Suhang Wang}, \bibinfo{person}{Xianfeng Tang}, \bibinfo{person}{Tsung-Yu Hsieh}, {and} \bibinfo{person}{Vasant Honavar}.} \bibinfo{year}{2020}\natexlab{}.
\newblock \showarticletitle{Adversarial attacks on graph neural networks via node injections: A hierarchical reinforcement learning approach}. In \bibinfo{booktitle}{\emph{The Web Conference}}.
\newblock


\bibitem[Takahashi(2019)]%
        {takahashi2019indirect}
\bibfield{author}{\bibinfo{person}{Tsubasa Takahashi}.} \bibinfo{year}{2019}\natexlab{}.
\newblock \showarticletitle{Indirect adversarial attacks via poisoning neighbors for graph convolutional networks}. In \bibinfo{booktitle}{\emph{IEEE International Conference on Big Data}}.
\newblock


\bibitem[Tao et~al\mbox{.}(2021)]%
        {tao2021single}
\bibfield{author}{\bibinfo{person}{Shuchang Tao}, \bibinfo{person}{Qi Cao}, \bibinfo{person}{Huawei Shen}, \bibinfo{person}{Junjie Huang}, \bibinfo{person}{Yunfan Wu}, {and} \bibinfo{person}{Xueqi Cheng}.} \bibinfo{year}{2021}\natexlab{}.
\newblock \showarticletitle{Single node injection attack against graph neural networks}. In \bibinfo{booktitle}{\emph{ACM International Conference on Information and Knowledge Management}}.
\newblock


\bibitem[Veli{\v{c}}kovi{\'c} et~al\mbox{.}(2018)]%
        {velivckovic2018graph}
\bibfield{author}{\bibinfo{person}{Petar Veli{\v{c}}kovi{\'c}}, \bibinfo{person}{Guillem Cucurull}, \bibinfo{person}{Arantxa Casanova}, \bibinfo{person}{Adriana Romero}, \bibinfo{person}{Pietro Li{\`o}}, {and} \bibinfo{person}{Yoshua Bengio}.} \bibinfo{year}{2018}\natexlab{}.
\newblock \showarticletitle{Graph Attention Networks}. In \bibinfo{booktitle}{\emph{International Conference on Learning Representations}}.
\newblock


\bibitem[Waniek et~al\mbox{.}(2018)]%
        {waniek2018hiding}
\bibfield{author}{\bibinfo{person}{Marcin Waniek}, \bibinfo{person}{Tomasz~P Michalak}, \bibinfo{person}{Michael~J Wooldridge}, {and} \bibinfo{person}{Talal Rahwan}.} \bibinfo{year}{2018}\natexlab{}.
\newblock \showarticletitle{Hiding individuals and communities in a social network}.
\newblock \bibinfo{journal}{\emph{Nature Human Behaviour}} \bibinfo{volume}{2}, \bibinfo{number}{2} (\bibinfo{year}{2018}), \bibinfo{pages}{139--147}.
\newblock


\bibitem[Wu et~al\mbox{.}(2019)]%
        {wu2019adversarial}
\bibfield{author}{\bibinfo{person}{Huijun Wu}, \bibinfo{person}{Chen Wang}, \bibinfo{person}{Yuriy Tyshetskiy}, \bibinfo{person}{Andrew Docherty}, \bibinfo{person}{Kai Lu}, {and} \bibinfo{person}{Liming Zhu}.} \bibinfo{year}{2019}\natexlab{}.
\newblock \showarticletitle{Adversarial examples for graph data: deep insights into attack and defense}. In \bibinfo{booktitle}{\emph{International Joint Conference on Artificial Intelligence}}.
\newblock


\bibitem[Wu et~al\mbox{.}(2020)]%
        {wu2020comprehensive}
\bibfield{author}{\bibinfo{person}{Zonghan Wu}, \bibinfo{person}{Shirui Pan}, \bibinfo{person}{Fengwen Chen}, \bibinfo{person}{Guodong Long}, \bibinfo{person}{Chengqi Zhang}, {and} \bibinfo{person}{S~Yu Philip}.} \bibinfo{year}{2020}\natexlab{}.
\newblock \showarticletitle{A comprehensive survey on graph neural networks}.
\newblock \bibinfo{journal}{\emph{IEEE Transactions on Neural Networks and Learning Systems}} \bibinfo{volume}{32}, \bibinfo{number}{1} (\bibinfo{year}{2020}), \bibinfo{pages}{4--24}.
\newblock


\bibitem[Xiao et~al\mbox{.}(2022)]%
        {xiao2022graph}
\bibfield{author}{\bibinfo{person}{Shunxin Xiao}, \bibinfo{person}{Shiping Wang}, \bibinfo{person}{Yuanfei Dai}, {and} \bibinfo{person}{Wenzhong Guo}.} \bibinfo{year}{2022}\natexlab{}.
\newblock \showarticletitle{Graph neural networks in node classification: survey and evaluation}.
\newblock \bibinfo{journal}{\emph{Machine Vision and Applications}} \bibinfo{volume}{33}, \bibinfo{number}{1} (\bibinfo{year}{2022}), \bibinfo{pages}{4}.
\newblock


\bibitem[Xu et~al\mbox{.}(2019)]%
        {xu2019topology}
\bibfield{author}{\bibinfo{person}{Kaidi Xu}, \bibinfo{person}{Hongge Chen}, \bibinfo{person}{Sijia Liu}, \bibinfo{person}{Pin-Yu Chen}, \bibinfo{person}{Tsui-Wei Weng}, \bibinfo{person}{Mingyi Hong}, {and} \bibinfo{person}{Xue Lin}.} \bibinfo{year}{2019}\natexlab{}.
\newblock \showarticletitle{Topology Attack and Defense for Graph Neural Networks: An Optimization Perspective}. In \bibinfo{booktitle}{\emph{International Joint Conference on Artificial Intelligence}}.
\newblock


\bibitem[Xu et~al\mbox{.}(2024)]%
        {xu2024attacks}
\bibfield{author}{\bibinfo{person}{Ying Xu}, \bibinfo{person}{Michael Lanier}, \bibinfo{person}{Anindya Sarkar}, {and} \bibinfo{person}{Yevgeniy Vorobeychik}.} \bibinfo{year}{2024}\natexlab{}.
\newblock \showarticletitle{Attacks on Node Attributes in Graph Neural Networks}. In \bibinfo{booktitle}{\emph{Artificial Intelligence for Cyber Security Workshop, AAAI}}.
\newblock


\bibitem[Zhang and Chen(2018)]%
        {zhang2018link}
\bibfield{author}{\bibinfo{person}{Muhan Zhang} {and} \bibinfo{person}{Yixin Chen}.} \bibinfo{year}{2018}\natexlab{}.
\newblock \showarticletitle{Link prediction based on graph neural networks}. In \bibinfo{booktitle}{\emph{Conference on Neural Information Processing Systems}}.
\newblock


\bibitem[Zhang et~al\mbox{.}(2018)]%
        {zhang2018end}
\bibfield{author}{\bibinfo{person}{Muhan Zhang}, \bibinfo{person}{Zhicheng Cui}, \bibinfo{person}{Marion Neumann}, {and} \bibinfo{person}{Yixin Chen}.} \bibinfo{year}{2018}\natexlab{}.
\newblock \showarticletitle{An end-to-end deep learning architecture for graph classification}. In \bibinfo{booktitle}{\emph{AAAI Conference on Artificial Intelligence}}.
\newblock


\bibitem[Zhang and Zitnik(2020)]%
        {zhang2020gnnguard}
\bibfield{author}{\bibinfo{person}{Xiang Zhang} {and} \bibinfo{person}{Marinka Zitnik}.} \bibinfo{year}{2020}\natexlab{}.
\newblock \showarticletitle{Gnnguard: Defending graph neural networks against adversarial attacks}. In \bibinfo{booktitle}{\emph{Conference on Neural Information Processing Systems}}.
\newblock


\bibitem[Zhiyao et~al\mbox{.}(2023)]%
        {zhiyao2024opengsl}
\bibfield{author}{\bibinfo{person}{Zhou Zhiyao}, \bibinfo{person}{Sheng Zhou}, \bibinfo{person}{Bochao Mao}, \bibinfo{person}{Xuanyi Zhou}, \bibinfo{person}{Jiawei Chen}, \bibinfo{person}{Qiaoyu Tan}, \bibinfo{person}{Daochen Zha}, \bibinfo{person}{Yan Feng}, \bibinfo{person}{Chun Chen}, {and} \bibinfo{person}{Can Wang}.} \bibinfo{year}{2023}\natexlab{}.
\newblock \showarticletitle{Opengsl: A comprehensive benchmark for graph structure learning}. In \bibinfo{booktitle}{\emph{Conference on Neural Information Processing Systems}}.
\newblock


\bibitem[Zhu et~al\mbox{.}(2019)]%
        {zhu2019robust}
\bibfield{author}{\bibinfo{person}{Dingyuan Zhu}, \bibinfo{person}{Ziwei Zhang}, \bibinfo{person}{Peng Cui}, {and} \bibinfo{person}{Wenwu Zhu}.} \bibinfo{year}{2019}\natexlab{}.
\newblock \showarticletitle{Robust graph convolutional networks against adversarial attacks}. In \bibinfo{booktitle}{\emph{ACM International Conference on Knowledge Discovery and Data Mining}}.
\newblock


\bibitem[Z{\"u}gner et~al\mbox{.}(2018)]%
        {zugner2018adversarial}
\bibfield{author}{\bibinfo{person}{Daniel Z{\"u}gner}, \bibinfo{person}{Amir Akbarnejad}, {and} \bibinfo{person}{Stephan G{\"u}nnemann}.} \bibinfo{year}{2018}\natexlab{}.
\newblock \showarticletitle{Adversarial attacks on neural networks for graph data}. In \bibinfo{booktitle}{\emph{ACM International Conference on Knowledge Discovery and Data Mining}}.
\newblock


\bibitem[Z{\"u}gner and G{\"u}nnemann(2019)]%
        {zugner2019adversarial}
\bibfield{author}{\bibinfo{person}{Daniel Z{\"u}gner} {and} \bibinfo{person}{Stephan G{\"u}nnemann}.} \bibinfo{year}{2019}\natexlab{}.
\newblock \showarticletitle{Adversarial Attacks on Graph Neural Networks via Meta Learning}. In \bibinfo{booktitle}{\emph{International Conference on Learning Representations}}.
\newblock


\end{thebibliography}





\clearpage
\appendix
\section{\measure on Node-Feature Attacks}\label{appendix:feature}
\resubmit{
In the main paper, we focus on topological attacks and propose a novel measure \measure (see Sec.~\ref{sec:method}).
In this section, we shall show that \measure can be easily extended to a noticeability measure on node-feature attacks.
}

\smallsection{Node-feature attack methods.}
\resubmit{
There are several methods designed to perform attacks on the node features~\cite{zugner2018adversarial, zugner2019adversarial, ma2020towards, xu2024attacks, takahashi2019indirect}.
Although these methods significantly degrade GNN performance, most of these methods either do not consider the noticeability of attacked node features, or apply a simple L1 or L2 norm constraint on the difference between the original features and the attacked features.
To our best knowledge, the only exception is a co-occurrence-based feature score~\citep{zugner2018adversarial}, which we will use as a competitor below (see Sec.~\ref{appendix:feature:exp}).
Moreover, despite the fact that many real-world datasets~\cite{sen2008collective, rozemberczki2021multi} use binary node features (e.g., bag-of-words), some methods~\cite{ma2020towards, xu2024attacks, takahashi2019indirect} perturb the features with continuous values, which makes the attacks blatant and easily distinguishable.
}

\subsection{LFO: \underline{L}earnable \underline{F}eature Sc\underline{o}rer}\label{appendix:feature:method}
\resubmit{
Like LEO (learnable edge scorer) designed for topological attacks (see Sec.~\ref{sec:method:measure}), to measure the noticeability of attacked node features, a learnable feature scorer (LFO) can be employed in \measure to identify suspicious features.
Below, we present how we design the LFO and how it is trained in a self-supervised manner without knowing the original unattacked features.}

\smallsection{Structure.}
\resubmit{
Recall that $X \in {\mathbb{R}}^{n \times k}$ denotes node features, where each node $v \in [n]$ has a $k$-dimensional node feature $X_v \in \mathbb{R}^k$.
The input of \lfo is an attacked graph $\hat{G}$, and the output is a feature score $f((v,i);\hat{X})$ for each node $v \in [n]$ and each feature (index) $i \in [k]$, which indicates how likely each $X_{vi} \in \mathbb{R}$ is an unattacked node feature (element).
For the results here, we use an autoencoder as the backbone of \lfo.
Note that any model that can output feature scores can be used as the backbone.
The autoencoder is designed to take the attacked node features $\hat{X}$ as input and output a matrix $\tilde{X}$ of the same size, where each element $\tilde{X}_{vi}$ represents the score for the corresponding feature $X_{vi}$.}

\smallsection{Training.}
\resubmit{
Since \lfo is trained on (potentially) attacked features, i.e., \lfo does not assume to know whether the given feature is attacked or not, we determine the feature labels for training \lfo in a self-supervised manner.
As in Sec.~\ref{sec:method:training}, given attacked node features $\hat{X}$, the positive samples are $T_p=\{(v,i)|\hat{X}_{vi}\neq0\}$, and the negative samples $T_n$ are randomly sampled from $\{(v,i)|\hat{X}_{vi}=0\}$.
The number of negative samples is a hyperparameter.}
\resubmit{
Furthermore, to train \lfo, a similar loss function is used as the one used to train \model (see Sec.~\ref{sec:method:training}).
That is, the loss function for training \lfo is based on the cross-entropy loss and adaptive filtering.}

\subsection{Experiments}\label{appendix:feature:exp}
\smallsection{Datasets.}
\resubmit{
To generate attacked node features, we employ two node feature attack methods, \textsc{RWCS}~\cite{ma2020towards} and \metattack~\cite{zugner2019adversarial}, on \cora.
We obtain attacked features with different attack rates (4\%, 8\%, 12\%, and 16\%).
All results are averaged over five trials.
}

\smallsection{Competitor.}
\resubmit{
We compare \lfo with a co-occurrence-based feature score~\cite{zugner2018adversarial}, denoted by \cooccur.
In \cooccur, a probabilistic random walker is employed on the co-occurrence graph $C=(V^{(X)}, E^{(X)})$ of the node features $X \in \mathbb{R}^{n \times k}$, where the node set $V^{(X)} = [k]$ represents the set of features and the edge set $E^{(X)} \subseteq(V^{(X)}\times V^{(X)})$ denotes which features have co-occurred, i.e., $(i, j) \in E^{(X)}$ if and only if there exists a node $v$ such that $X_{vi} > 0$ and $X_{vj} > 0$.
With \cooccur, for a node $v \in [n]$ and a feature $i \in V^{(X)} = [k]$, the node-feature pair $(v, i)$ is unnoticeable if and only if the probability of the random walker on $C' = (V^{(X)}, E^{(X')})$ reaching $i$ after one step from the features originally present for the node $v$ is significantly large, where $X'_{vi} = 0$ and $X'_{v'i'} = X_{v'i'}$ for $(v',i') \neq (v, i)$ (i.e., $X'$ is obtained by removing the feature $X_{vi}$ from $X$).
Formally, for a node $v \in [n]$ and an attacked feature $i \in V^{(X)} = [k]$, the node-feature pair $(v, i)$ is unnoticeable if and only if the probability $p(i|S_{v})$ is larger than a threshold $\sigma$, where $S_{v} = \{j \in V^{(X)} | X'_{vj} > 0\} = \{j \in V^{(X)} | X_{vj} > 0, j \neq i\}$ is the set of features originally present for node $v$ (except for $i$ itself),
where the probability $p(i|S_{v})= \frac{1}{|S_v|} \sum_{j\in S_{v}} \frac{E^{(X)}_{ij}}{d_{j}}$ and $d_{j}$ is the degree of feature $j$ in the co-occurrence graph $C$.
The threshold is set as $\sigma=\frac{1}{2} \cdot \sum_{j \in S_{v}} \frac{1}{d_{j}}$, which is half of the maximum achievable probability.
Based on the idea of \cooccur, we define a noticeability score as follows.
Given a node $v \in [n]$ and an attacked feature $i \in [k]$, we set the feature score $f(v,i)$ as $\frac{1 - p(i|S_{v}))}{2\sigma} \in [0, 1]$.
A higher $f(v,i)$ value indicates that the attacked feature $i$ of node $v$ is more noticeable.
}

\smallsection{Evaluation.}
\resubmit{
For both \lfo and \cooccur, AUROC is used for performance evaluation, where a higher AUROC score indicates better performance in distinguishing attacked features and unattacked features. 
We report the mean and standard deviation computed over five trials. 
}

\smallsection{Results.}
\resubmit{
In Table~\ref{tab:appendix:feature}, we present the noticeability scores obtained from two feature scoring methods (\lfo and \cooccur). 
\lfo outperforms \cooccur for each attack method and for each attack rate.
}

\begin{table}[ht]
    
    \centering
    \caption{\label{tab:appendix:feature}
    \underline{Performance of \lfo}: \lfo consistently outperforms \cooccur, better classifying  attacked and real features.}
    \begin{subtable}[b]{0.5\textwidth}
    \centering
    \textbf{(1) \textsc{RWCS}} \\ 
    \resizebox{0.97\textwidth}{!}{
    \begin{tabular}{c||c|c|c|c}
        \hline
        \multirow{2}{*}{\textbf{Method}} & \multicolumn{4}{c}{\textbf{Attack Rate}} \\
        \cline{2-5}
        & \textbf{4\%} & \textbf{8\%} & \textbf{12\%} & \textbf{16\%} \\
        \hline
        \hline
        \cooccur & 0.522$\pm$0.011 & 0.574$\pm$0.012 & 0.583$\pm$0.007 & 0.626$\pm$0.022 \\
        \lfo (proposed) & \textbf{0.769$\pm$0.041} & \textbf{0.895$\pm$0.024} & \textbf{0.935$\pm$0.016} & \textbf{0.940$\pm$0.014} \\
        \hline
    \end{tabular}}
    \end{subtable}
    \newline
    \vspace*{0.5em}
    \newline
    \begin{subtable}[b]{0.5\textwidth}
    \centering
    \textbf{(2) \metattack}\\
    \resizebox{0.97\textwidth}{!}{
    \begin{tabular}{c||c|c|c|c}
        \hline
        \multirow{2}{*}{\textbf{Method}} & \multicolumn{4}{c}{\textbf{Attack Rate}} \\
        \cline{2-5}
        & 4\% & 8\% & 12\% & 16\% \\
        \hline
        \hline
        \cooccur & 0.715$\pm$0.007 & 0.736$\pm$0.001 & 0.751$\pm$0.008 & 0.765$\pm$0.009 \\
        \lfo (proposed) & \textbf{0.944$\pm$0.009} & \textbf{0.939$\pm$0.003} & \textbf{0.941$\pm$003} & \textbf{0.943$\pm$0.003} \\
        \hline
    \end{tabular}}
    \end{subtable}
\end{table}

\section{Algorithmic Details of \measure}
\label{appendix:measure}\label{appendix:auroc}

The key algorithmic step in \measure is the computation of AUROC scores.
Below, we shall describe the details. 

Given an original graph $G$ and an attacked graph $\hat{G}$,
let $E = \{(u,v) | A_{uv} = 1\}$ denote the edge set in the original graph, 
let $\hat{E} = \{(u,v) | \hat{A}_{uv} = 1\}$ denote the edge set in the attacked graph,
and let $E_0 = E \cup \hat{E}$ denote the union of edge sets in the original graph and the attacked graph.
For scores on $E_0$, i.e., $f(e;\hat{G})$ for $e \in E_0$,
we compute the AUROC score $AUROC(E_0, E, f)$ w.r.t. $f = f(e;\hat{G})$'s as follows:
\begin{enumerate}
    \item we sort the pairs in $E_0$ w.r.t. $f_e$'s in descending order: $(e_1, e_2, \ldots, e_{|E_0|})$ with $f_{e_1} \geq f_{e_2} \geq \cdots \geq f_{e_{|E_0|}}$
    \item \textbf{initialization:} $E_s = \emptyset$
    \item for $i = 1, 2, \ldots, |E_0|$, we first add $e_i$ into $E_s$, i.e., $E_s \gets E_s \cup \{e_i\}$, and compute the true-positive rate $\text{TP}_i = \frac{|E_s \cap E|}{|E_s|}$ and the false-positive rate $\text{FP}_i = \frac{|E_s \cap \hat{E}|}{|E_s|}$; we then plot a point $(\text{FP}_i, \text{TP}_i)$
    \item we connect all the points $(\text{FP}_i, \text{TP}_i)$'s to obtain a curve, and compute the AUROC of the curve of the plotted points.
\end{enumerate}

See Algorithm~\ref{algorithm:hidenseek} for the pseudocode of the whole process of calculating \measure.

\begin{algorithm}[h!]
	\caption{Procedure of computing \measure}
    \label{algorithm:hidenseek}
    \begin{flushleft}
        \textbf{INPUT:} An original (unattacked) graph $G=(A,X)$, an attacked graph $\hat{G}=(\hat{A},X)$, and a trained edge scorer (\model) $f$.\\
        \textbf{OUTPUT:} noticeability $N_{\hat{G}}$
    \end{flushleft}
	\begin{algorithmic}[1]
        \State $E = \{(u,v) | A_{uv} = 1\}$
        \State $\hat{E} = \{(u,v) | \hat{A}_{uv} = 1\}$
        \State $E_0 = E \cup \hat{E}$
        \State $N_{\hat{G}} = AUROC(E_0, E, f)$ \Comment{See Appendix~\ref{appendix:auroc}}\\
        \Return $N_{\hat{G}}$
	\end{algorithmic} 
\end{algorithm}

\section{Additional Analysis on \measure}
\subsection{Analysis on the Bypassability and Overlookingness of \measure}
\label{appendix:crown_our}
\begin{figure}[h]
    \begin{flushright}
    \includegraphics[width=0.48\linewidth]{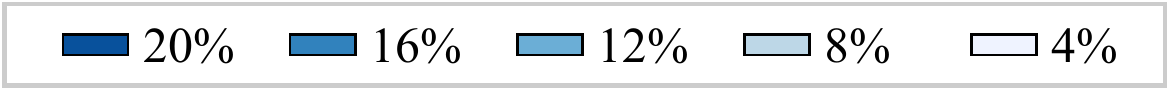} 
    \end{flushright}
    
    \begin{subfigure}{0.49\linewidth}
        \centering
        \includegraphics[width=\linewidth]{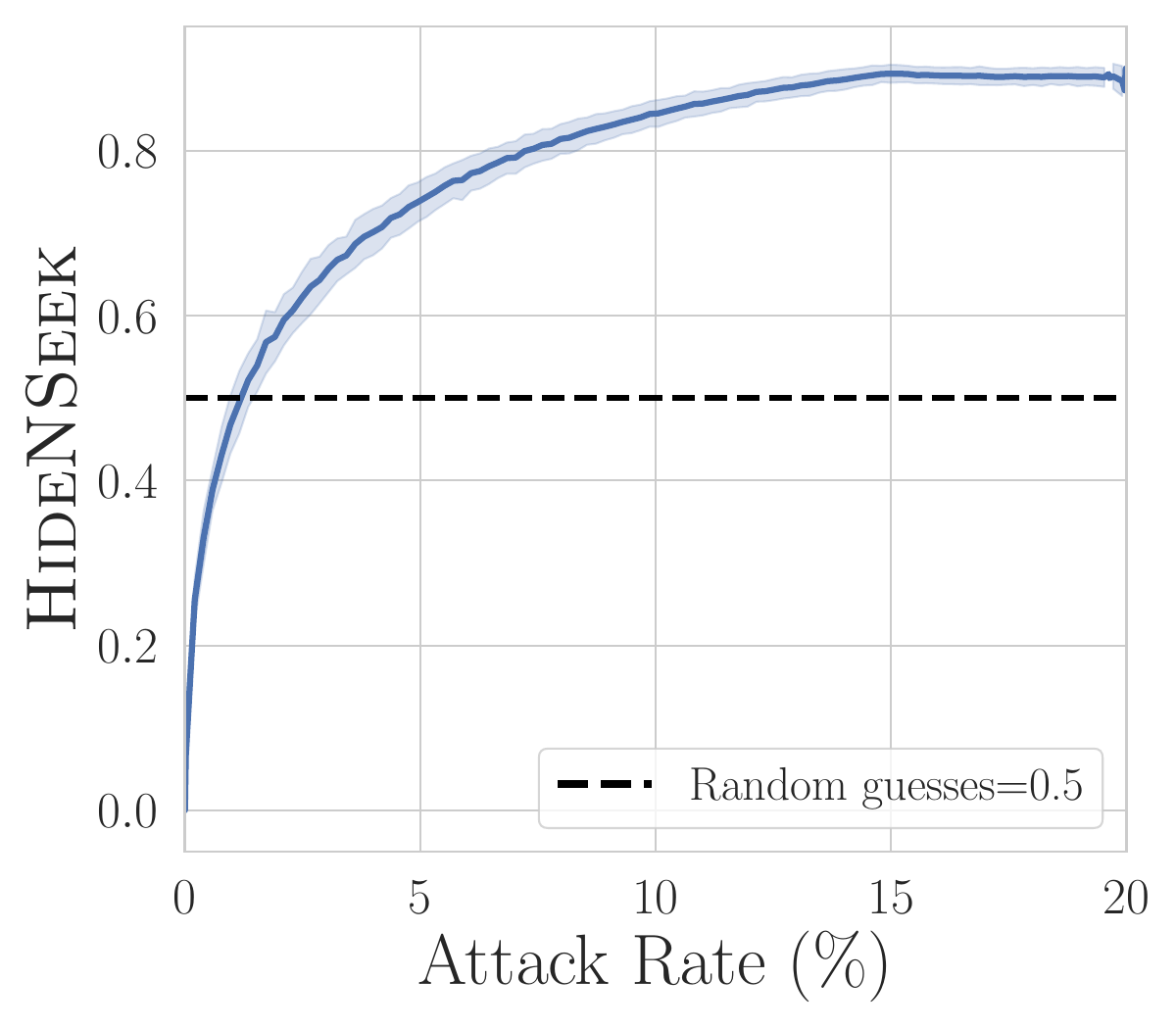}
        \caption{\measure is sensitive}
        \label{fig:crown:our_overlooking}
    \end{subfigure} 
    \begin{subfigure}{0.49\linewidth}
        \centering
        \includegraphics[width=\linewidth]{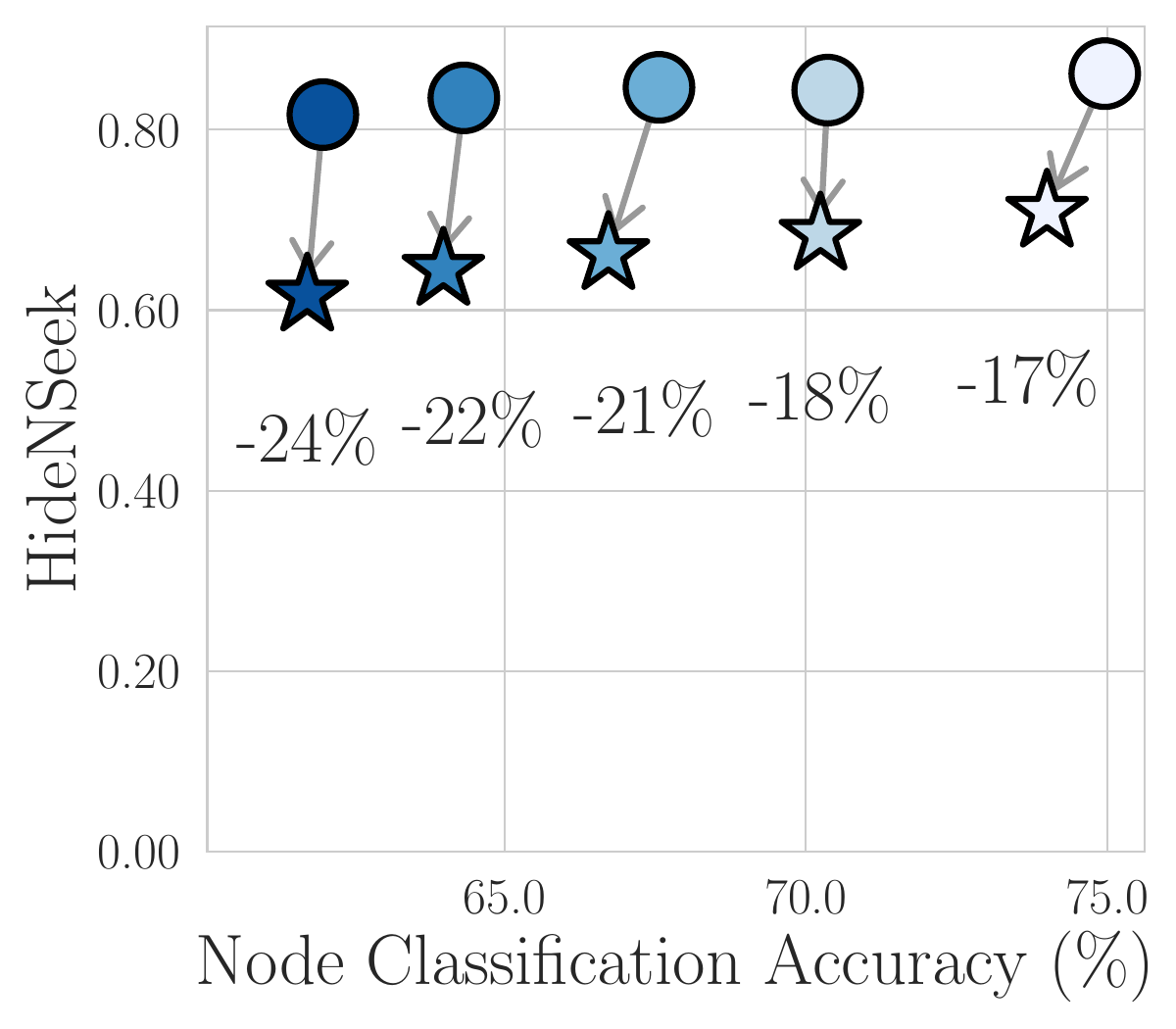}
        \caption{\measure is less bypassable}
        \label{fig:crown:our_bypass}
    \end{subfigure} 
    \caption{
    (left) \measure is sensitive: \measure notices attacks immediately, even for adaptive attacks with a low attack rate, while the existing measures fail to do so (c.f. Fig.~\ref{fig:crown:overlooking});
    (right) \measure is less bypassable: when the adaptive attack is adopted, the noticeability of \measure reduces less, compared to the existing measures (c.f. Fig.~\ref{fig:crown:bypass}).
    }
    \label{fig:crown:our}
\end{figure}

\resubmit{We provide the analysis on \measure regarding the properties of bypassability and overlookingness (supplementing Sec.~\ref{sec:limit:bypassable} and Sec.~\ref{sec:limit:overlooking}).
Fig.~\ref{fig:crown:our}(a) shows that \measure is very sensitive, noticing the change immediately and increasing the attack rate rapidly, while the existing measures fail to do so (c.f. Fig.~\ref{fig:crown:overlooking}).
Fig.~\ref{fig:crown:our}(b) shows that \measure is less bypassable than the existing measures; recall that the existing noticeability measures reduce at least 64\%, and some even reduce near 100\% (c.f. Fig.~\ref{fig:crown:bypass}).
Below, we shall provide more analysis regarding the bypassability of \measure.
}

\section{Complexity Analysis}
\label{appendix:proofcomplexity}
We provide the time and space complexity of each noticeability measure shown in Table~\ref{tab:method:complexity}.

Consider an original graph $G$ and an attacked graph $\hat{G}$, each with their corresponding adjacency matrix.
Let $n$ represent the number of nodes in both $G$ and $\hat{G}$, and let $m$ and $\hat{m}$ represent the number of edges in $G$ and $\hat{G}$, respectively.
Let $k$ denote the dimension of the raw node features.
Additionally, we assume that the attacked graph $\hat{G}$ is also stored in the form of an adjacency list, which contains the list of the neighbors of each node.
For the analysis, we assume that \model consists of $l$ layers of GCN, where the hidden dimension of each layer is $O(k)$.

\subsection{Time Complexity}
\label{appendix:proofcomplexity:time}
The process of calculating \measure includes inferring scores for $\hat{m}$ 
edges in $\hat{G}$ using \model and calculate AUROC with label consisting of $m$ real edges in 
$G$ labeled as 1 and $\hat{m}-m$ attack edges existing in $\hat{G}$ labeled as 0.
The time complexity of inferring scores for $\hat{m}$ edges in $\hat{G}$ using \model is dominated by the forward pass of the GCN~\cite{wu2020comprehensive}  ($O(lk(\hat{m}+nk))$) and the bilinear operation ($O(k^2\hat{m})$).
\resubmit{
Specifically, 
the time complexity of training \model is equivalent to multiplying the inference complexity by the number of training iterations, which is a constant, so the time complexity remains $O(k((l+k)\hat{m}+lkn))$.
}
The time complexity of calculating AUROC is dominated by sorting scores of $\hat{m}$ edges, which is $O(\hat{m}log\hat{m})$.
Therefore, the total time complexity is $O(\hat{m}log\hat{m} + k((l+k)\hat{m}+lkn))$.

\begin{table}[ht]
    \centering
    \caption{\label{tab:method:complexity}
    Time and space complexities of computing noticeability measures.}
    \scalebox{0.8}{
    \begin{tabular}{c|c|c}
    \toprule
    Measure & Time Complexity & Space Complexity \\
    \midrule
    \textsc{DegreeKS}~\citep{hussain2021structack} & $O(n\log n + m + \hat{m})$ & $O(n + m + \hat{m})$ \\ 
    \textsc{ClsCoefKS}~\citep{hussain2021structack} & $O(n\log n + m^{1.48} + \hat{m}^{1.48})$ & $O(n + m^{1.05} + \hat{m}^{1.05})$ \\
    \textsc{DegreeLR}~\citep{zugner2018adversarial} & $O(n +  m + \hat{m})$ & $O(n + m + \hat{m})$ \\
    \textsc{HomophKS}~\citep{chen2021understanding} & $O(n\log n + k(n + m + \hat{m}))$ & $O(nk + m + \hat{m})$\\
    \measure & $O(\hat{m}log\hat{m} + k((l+k)\hat{m}+lkn))$ & $O(lk(n + k) + m + \hat{m})$\\
    \bottomrule
    \end{tabular}}
\end{table}

\subsection{Space Complexity}
\label{appendix:proofcomplexity:space}
\resubmit{
The space complexity of \measure depends on the training parameters of \model.
Specifically, the Graph Structure Learning (GSL) module within \model is the most dominant factor, as most GSL methods typically require high space complexity~\citep{zhiyao2024opengsl}. 
To alleviate this, we use a GSL method that constructs a $k$-nearest neighbor (kNN) graph based on node embeddings, as described in~\cite{fatemi2021slaps}. 
The space complexity of this GSL method is $O(lk (n + k))$ (where $l$ is the number of layers), which is comparable to the space complexity of the GCN.
For the other parts, the space complexity involves storing $G$, $\hat{G}$, and node features, which is 
$O(n + m + \hat{m})$.
Therefore, the total space complexity is 
$O(lk(n + k) + m + \hat{m})$.
}
\end{document}